\documentclass[
12pt, 
english, 
onehalfspacing, 
headsepline,
openany,
]{MastersDoctoralThesis} 
\usepackage{booktabs}
\usepackage[utf8]{inputenc} 
\usepackage[T1]{fontenc} 
\usepackage{amsmath}
\usepackage{xcolor}
\usepackage{colortbl,booktabs}
\usepackage{amsmath}
\usepackage{caption}
\usepackage{subfigure}
\usepackage{tabularx}
\usepackage{graphicx}
\usepackage{picinpar}
\usepackage{amsmath}
\usepackage{url}
\usepackage{flushend}
\usepackage{colortbl}
\usepackage{soul}
\usepackage{multirow}
\usepackage{pifont}
\usepackage{color}
\usepackage{alltt}
\usepackage{mathtools}
\usepackage[hidelinks]{hyperref}
\usepackage{enumerate}
\usepackage{breakurl}
\usepackage{epstopdf}
\usepackage{pbox}
\usepackage{crop}
\usepackage{graphicx} 
\usepackage{float} 
\usepackage{subfigure} 
\usepackage{CJK}
\usepackage{mathpazo}
\usepackage{rotating}
\usepackage{bbm}
\usepackage{algorithmic}
\usepackage[ruled,vlined]{algorithm2e}
\usepackage[normalem]{ulem}
\usepackage{adjustbox}

\usepackage[backend=bibtex,natbib=true]{biblatex} 

\usepackage[autostyle=true]{csquotes} 

\thesistitle{Medical Image Analysis using Deep Relational Learning} 
\supervisor{Dr. Huiyu \textsc{Zhou} } 

\examiner{} 
\degree{Master of Philosophy} 
\author{Zhihua \textsc{Liu}} 
\addresses{} 

\subject{Computer Science} 
\university{\href{https://le.ac.uk/}{University of Leicester}} 
\department{\href{https://le.ac.uk/informatics}{School of Informatics}} 
\group{\href{https://sites.google.com/site/huiyujoe/}{Biomedical Image Processing Lab}} 
\faculty{\href{http://faculty.university.com}{}} 

\AtBeginDocument{
\hypersetup{pdftitle=\ttitle} 
\hypersetup{pdfauthor=\authorname} 
\hypersetup{pdfkeywords=\keywordnames} 
}

\begin{document}
\linespread{1.25}
\frontmatter 

\pagestyle{plain} 


\begin{titlepage}
\begin{center}


{\huge \bfseries \ttitle\par}\vspace{2.5cm} 
 
 
\vfill

\LARGE {Thesis submitted for the degree of \\ \degreename\\at the University of Leicester}\\[1cm] 
\textit{by}\\[1cm]
{Zhihua Liu}\\ \groupname\\\deptname\\{University of Leicester}\\[2cm] 
 
\vfill

{\LARGE \today}\\[4cm] 

\vfill
\end{center}
\end{titlepage}


\begin{declaration}
\addchaptertocentry{\authorshipname} 
\noindent I, \authorname, declare that this thesis titled, \enquote{\ttitle} and the work presented in it are my own. I confirm that:

\begin{itemize} 
\item This work was done wholly or mainly while in candidature for a research degree at this University.
\item Where any part of this thesis has previously been submitted for a degree or any other qualification at this University or any other institution, this has been clearly stated.
\item Where I have consulted the published work of others, this is always clearly attributed.
\item Where I have quoted from the work of others, the source is always given. With the exception of such quotations, this thesis is entirely my own work.
\item I have acknowledged all main sources of help.
\item Where the thesis is based on work done by myself jointly with others, I have made clear exactly what was done by others and what I have contributed myself.\\
\end{itemize}
 
\noindent Signed: Zhihua Liu\\
\rule[0.5em]{25em}{0.5pt} 
 
\noindent Date: 23/Oct/2020\\
\rule[0.5em]{25em}{0.5pt} 
\end{declaration}







\begin{abstract}
\addchaptertocentry{\abstractname} 
Benefited from deep learning techniques, remarkable progress has been made within the medical image analysis area in recent years. However, it is very challenging to fully utilize the relational information (the relationship between tissues or organs or images) within the deep neural network architecture. Thus in this thesis, we propose two novel solutions to this problem called implicit and explicit deep relational learning. We generalize these two paradigms of deep relational learning into different solutions and evaluate them on various medical image analysis tasks.

Automated segmentation of brain glioma in 3D magnetic resonance imaging plays an active role in glioma diagnosis, progression monitoring and surgery planning. In this work, we propose a novel Context-Aware Network that effectively models implicit relation information between features to perform accurate 3D glioma segmentation. We evaluate our proposed method on publicly accessible brain tumor segmentation datasets BRATS2017 and BRATS2018 against several state-of-the-art approaches using different segmentation metrics. The experimental results show that the proposed algorithm has better or competitive performance, compared to the standard approaches.

Subsequently, we propose a new hierarchical homography estimation network to achieve accurate medical image mosaicing by learning the explicit spatial relationship between adjacent frames. We use the UCL Fetoscopy Placenta dataset to conduct experiments and our hierarchical homography estimation network outperforms the other state-of-the-art mosaicing methods while generating robust and meaningful mosaicing results on unseen frames.
\end{abstract}


\begin{acknowledgements}
\addchaptertocentry{\acknowledgementname} 
First of all, I want to thank my father Mr. Yuhua Du, my mother Mrs. Zengxia Xiao and my fiancee Miss. Xinyu Wang. Thank them for their support to me and the family. I cannot repay the support from my family, and it is also my biggest motivation to continue scientific research.

Secondly, I want to thank my first supervisor, Prof. Huiyu Zhou. Prof. Huiyu Zhou has a rigorous attitude towards science. He also dedicates himself to his work. He is both a good teacher and a good friend. I am grateful to him for his continuous criticism and teaching, which have benefited me for life. At the same time, I would also like to thank my second supervisor, Prof. Yudong Zhang, for his professional guidance on my academic work.

Also, I would like to thank all colleagues of Biomedical Image Processing Lab (BIPL), especially Mr. Zheheng Jiang, Mr. Lei Tong, Mr. Long Chen, Mr. Feixiang Zhou, Mr. Jialin Lyu, Mr. Honghui Du. Thank all of you for your continuous support and help.

Finally, I would also like to thank all the collaborators, including, Dr. Qianni Zhang from Queen Mary Univeristy of London, Dr. Yinhai Wang from AstraZeneca R$\&$D Cambridge, Dr. Caifeng Shan from Philips Research Edinhoven, Prof. Ling Li from University of Kent, Prof. Xiangrong Zhang from Xidian University. Prof. Bin Yang from University of Leicester, Prof. Stephen McKenna from University of Dundee, Prof. Jianguo Zhang from Southern University of Science and Technology, Dr. Tianjun Huang from Southern University of Science and Technology. Dr. Sophia Bano from University College London. It is my honor to collaborate with these world-class researchers. Thank all collaborators for their trust, support and communication, which broadened my horizons and enabled me to move forward in the direction of machine learning and medical image processing. At the same time, I would also like to thank all the staff at the School of Informatics at the University of Leicester. Their professionalism has provided logistical support during my research work.

This thesis was written during the COVID epidemic in 2020. I hope you can stay safe and stay healthy if you are reading this dissertation.

\end{acknowledgements}


\tableofcontents 

\listoffigures 

\listoftables 

\mainmatter 

\pagestyle{thesis} 



\chapter{Introduction} 

\label{Chapter1} 


\newcommand{\keyword}[1]{\textbf{#1}}
\newcommand{\tabhead}[1]{\textbf{#1}}
\newcommand{\code}[1]{\texttt{#1}}
\newcommand{\file}[1]{\texttt{\bfseries#1}}
\newcommand{\option}[1]{\texttt{\itshape#1}}

\section{Research Background}
As an important means to assist doctors in diagnosis and treatment, automated medical imaging analysis provides doctors with rich and accurate diagnostic information. At the same time, medical image interpretation is also promoting the understanding of human physiological structure and drug discovery. With the rapid development and progress of computation and digital medical imaging, digital medical image analysis has gradually become one of the most important and valuable research fields. Automated medical image analysis is an interdisciplinary research field that integrates imaging technology, numerical calculation and modeling, digital image processing and artificial intelligence. In the early application of medical imaging technology, such as X-ray, doctors obtain and interpret the physiological simulation images of patients with years of learning and accumulated experience, and then give diagnosis conclusions and treatment suggestions. However, it is inefficient to rely on the experience and knowledge of doctors. During the period of large-scale investigation of specific diseases, the disadvantages of purely artificial analysis methods are exposed. Nowadays, medical imaging technology and its application have developed the diagnosis capability from a limited range of diseases with specific symptoms, such as breast cancer, to a wide range of diseases and application scenarios.

With the development of medical imaging technology and hardware equipment, the research and analysis of high-dimensional medical images, such as CT and MRI, are becoming the mainstream. The acquisition of high-dimensional information also makes automated medical image analysis more extensive. It can be extended to the imaging of multiple organ physiology, anatomical morphology, process function and then promotes the development of related fields such as psychology, sociology and sports.

In dealing with massive and complex medical image information, pure manual interpretation and understanding are greatly restricted, and the potential value of massive information cannot be fully exploited and utilized. Therefore, researchers are more focused on the development of automatic and accurate computer-aided algorithms to help clinicians and researchers process massive information and complex tasks more efficiently and effectively. Referring to the task division in traditional computer vision, there are three basic visual tasks for medical image analysis and application:

\begin{enumerate}
\item Image registration: the purpose of early medical image registration is to jointly display images with different information, such as structure map and function information map, in a unified coordinate system. With the development of using three-dimensional imaging to study complex organs, such as the brain, there has been a study of time-series image registration and standard atlas template registration. At the same time, in order to eliminate the motion artifact in the process of image generation, image registration is also the preprocessing work of most image analysis approaches. The basic task of medical image registration is to find the corresponding relationship of objects in different images by various methods. After registration and transformation, the involved images can be connected in the same space.

\item Medical image classification: for one or more input images, an automatic algorithm can accurately classify whether the input image contains disease or not. Afterward, the classification algorithm can classify the input image into different grades, such as mild or severe disease, benign or malignant tumors. Accurate classification information can help researchers and doctors understand the disease and then give treatment suggestions.

\item Medical image segmentation: segmentation in medical images is one of the basic, important, and widely studied subfields in medical image analysis. The aim is to segment the region of interest (ROI), such as pathological tissues and organs, from complex image background by automated or semi-automated methods. Accurate, robust and fast image segmentation provides important prior knowledge for downstream tasks such as quantitative analysis, 3D reconstruction, medical robot, and provides an important foundation for image-guided surgery and radiotherapy planning.
\end{enumerate}

In recent ten years, the application and development of deep learning technology in various fields, such as computer vision, natural language processing, speech recognition, has been growing explosively. As a representation learning method, deep learning uses multiple processing layers consisting of complex structures or multiple nonlinear transformations to abstract features at a high level. Deep learning networks can be understood as the extension of the traditional neural networks, which extract representative features using nonlinear functions such as convolution. Deep learning techniques replace the traditional method of extracting hand-crafted features so that the system can learn and produce reasonable features. This powerful feature extraction ability makes the deep learning technology, especially the deep neural network, receive increasing attention in medical image analysis, and become the mainstream method in medical image analysis. Therefore, how to effectively design deep learning architecture, integrate different levels of image information, and further improve the performance of the automated algorithm in medical image analysis, is a broad research and application prospect.
\section{Research Objective}
In the past two years, we have focused on developing more effective and generalized deep learning algorithms for medical image tasks. We focused on two tasks here: brain glioma segmentation and fetoscopic photography image mosaicing task. Although brain glioma segmentation and medical image mosaicing belong to two different types of visual tasks, their medical image understanding and machine learning perspectives have great similarities and similarities. First of all, the two types of tasks share similar processing pipelines, that is, building a visual model based on feature learning and extraction. Second, relational information plays an important role in both tasks. In brain glioma segmentation tasks, more accurate sub-region segmentation can be performed by learning the correlation information between different tissue regions. In the medical image mosaicing task, the relational information is reflected in the spatial position relationship between adjacent frames. Accurate relational information can help reduce drift errors. Finally, efficient and accurate automated quantification technology for these two types of tasks has high clinical application value, which can further help doctors to more accurately locate and analyze the disease.

Therefore, we first propose a more accurate segmentation algorithm for specific diseases such as glioma. Then, we propose a more effective network to estimate the relative homography, i.e. the registration between adjacent images, which is expected to achieve accurate image mosaicing applied on a set of medical images with limited field-of-view (FoV).

The development of medical image analysis systems can be roughly divided into three stages. The early-stage algorithms are mainly based on modeling specific geometric features, such as edges and circles. These systems are simple and direct, but the performance is unsatisfactory. In the 1990s, data-driven supervised learning systems have become the mainstream. The most widely used models include decision tree, random field and support vector machine. Traditional machine learning systems effectively use the information provided from data, and then learn and update the model parameters, which greatly improve the performance of the model in various tasks of medical image analysis. One of the shortcomings of the traditional machine learning systems is that the final performance of the model is highly dependent on manual feature selection and combination, namely feature engineering. Effective feature engineering refers to the manual extraction of various pre-defined features from data and the optimal combination based on a researcher's experience and task characteristics. Compared to deep learning, traditional work is labor-intensive and time-consuming. In the past decade, deep learning systems based on the deep neural network has promoted the performance of medical image analysis to a new level. Especially in the application of medical image segmentation and classification, new works have been proposed. However, most of the deep neural network models do not combine the relational information between organs and tissues. Besides, the generalization performance of recent works is poor and cannot be effectively transferred from one task domain to another. To address these challenges, the objectives of our research project are summarized as follows:

(1) The main goal of this research project is to build novel deep learning tools to effectively address various problems and tasks in medical image analysis.

(2) The learning systems for medical image analysis proposed in the previous works do not effectively utilize the relational information between organs and tissues in medical images. Therefore, our second goal is to propose a novel deep learning paradigm to better model the relationship between features of different organs or tissues. We then apply it to various medical image analysis tasks, such as glioma segmentation and medical image mosaicing.

(3) Our final goal is to evaluate the effectiveness of our proposed method by using multiple datasets on different medical image analysis tasks.

\section{Contributions}
\label{C1:Contributions}
In this thesis, we present our research project in two parts. In the first part, we propose a new automated glioma segmentation system which is presented in Chapter \ref{Chapter3}

(1) We propose a novel Hybrid Context Aware Feature Extractor (HCA-FE). HCA-FE is built with a 3D feature interaction graph neural network and a 3D encoder-decoder convolutional neural network. Different from previous works that usually extract features in the convolutional space, HCA-FE learns hybrid context guided features in both a convolutional space and a feature interaction graph space (the relationship between neighboring feature nodes is utilized and continuously updated). To our knowledge, this is the first practice on brain glioma segmentation, which incorporates adaptive contextual information with graph convolution updates.

(2) We further propose a novel Context Guided Attentive Conditional Random Field (CG-ACRF) strategy for feature fusion. CG-ACRF based fusion module can attentively aggregate features from the feature interaction graph and convolutional space. Moreover, we formulate the mean-field approximation of the inference in the proposed CG-ACRF as a convolution operation, enabling the CG-ACRF to be embedded within any deep neural network seamlessly to achieve end-to-end training.

(3) We conduct extensive evaluations and demonstrate that our proposed method outperforms several state-of-the-art technologies using difference measure metrics on Multimodal Brain Tumor Image Segmentation Challenge (BraTS) datasets, i.e. BraTS2017 and BraTS2018.

In the second phase of our research project, we mainly contribute to the improvement of medical image mosaicing task. At this stage, our contributions are summarized as follows:

(1) We propose a new deep hierarchical homography estimation network to automatically and hierarchically estimate the 8 degree-of-freedom homography, upon which multi-scale (local level between adjacent frames and non-local level between long-range frames) homography are jointly learned and optimized in a data-driven manner.

(2) Inspired by recent works on color homography \cite{finlayson2017color}, we propose a new data generation method called Partially Image Generation (PIG). PIG only perturb the color, rotation, and translation movement between adjacent frames. The generated frames by PIG can be used for evaluating model performance during network training.

(3) We conduct extensive experiments on five different video clips from UCL Fetoscopy Placenta dataset. The results show that our method achieves state-of-the-art performance and generalizes well on different clips under different numbers of frames and different acquisition methods.

\section{Thesis Outline}
In this chapter, we provide an overview of the whole thesis. In Chapter \ref{Chapter1}, we briefly introduce the research background and introduction and show the expected objectives and contributions of the research project.

In Chapter \ref{Chapter2}, we review the related works of learning system based medical image analysis. These works can be roughly divided into two categories according to the design principles: one is manual feature extraction learning, the other is the deep neural network based automatic feature extraction learning. We systematically analyze the representative works of these two kinds of systems and point out their advantages and disadvantages and improvement direction.

In Chapter \ref{Chapter3}, we introduce our first contribution to glioma segmentation and propose Context Aware Network (CANet) for 3D brain glioma segmentation. In this chapter, we introduce our new segmentation system in detail and discuss the experimental results in depth.

In Chapter \ref{Chapter4}, we propose a novel medical image mosaicing method. We show the new hierarchical neural homography estimation network and partially image generation in detail. We evaluate the effectiveness of our mosaicing method on different video clips.

In Chapter \ref{Chapter5}, we summarize the research work in this thesis and discuss the limitation of research at this stage. We present the plan and the direction of future work.

\section{Publication List}

The content of Chapter \ref{Chapter3} appears in:
\\

Liu, Zhihua, Lei Tong, Long Chen, Feixiang Zhou, Zheheng Jiang, Qianni Zhang, Yinhai Wang, Caifeng Shan, Ling Li, and Huiyu Zhou. "CANet: Context Aware Network for 3D Brain Tumor Segmentation." arXiv preprint arXiv:2007.07788 (2020). Submitted to IEEE Transactions on Medical Imaging.
\\

Liu, Zhihua, Long Chen, Lei Tong, Feixiang Zhou, Zheheng Jiang, Qianni Zhang, Caifeng Shan et al. "Deep Learning Based Brain Tumor Segmentation: A Survey." arXiv preprint arXiv:2007.09479 (2020). Submitted to Elsevier Journal on Computerized Medical Imaging and Graphics.
\\

\noindent
\textbf{Non-thesis research:} I have also contributed to the following publications:
\\

Chen, Long, Zhihua Liu, Lei Tong, Zheheng Jiang, Shengke Wang, Junyu Dong, and Huiyu Zhou. "Underwater object detection using Invert Multi-Class Adaboost with deep learning", Proc. of International Joint Conference on Neural Networks (IJCNN), Glasgow, UK, 19-24 July, 2020.
\\

Jiang Zheheng, Zhihua Liu, Long Chen, Lei Tong, Xiangrong Zhang, Xiangyuan Lan, Danny Crookes, Ming-Hsuan Yang and Huiyu Zhou. “Detection and Tracking of Multiple Mice Using Part Proposal Networks”. arXiv preprint arXiv:1906.02831 (2019). Submitted to IEEE Transactions on Neural Networks and Learning Systems.
\\

Tong Lei, Zhihua Liu, Zheheng Jiang, Feixiang Zhou, Long Chen, Jialin Lyu, Xiangrong Zhang, Qianni Zhang, Sadka Abdul, Yinhai Wang, Ling Li, Huiyu Zhou. "Cost-sensitive Boosting Pruning Trees for depression detection on Twitter". arXiv preprint arXiv:1906.00398 (2019). Submitted to IEEE Transactions on Affective Computing.
\\

Chen Long, Zheheng Jiang, Lei Tong, Zhihua Liu, Aite Zhao, Qianni Zhang, Junyu Dong, and Huiyu Zhou. "Detection perceptual underwater image enhancement with deep learning and physical priors". arXiv preprint arXiv:2008.09697 (2020). IEEE Transactions on Circuits and Systems for Video Technology.


\chapter{Literature Review} 

\label{Chapter2} 
\section{Computational Aided Diagnosis with Medical Image Analysis}
\begin{figure*}[t]
\includegraphics[width=1\textwidth]{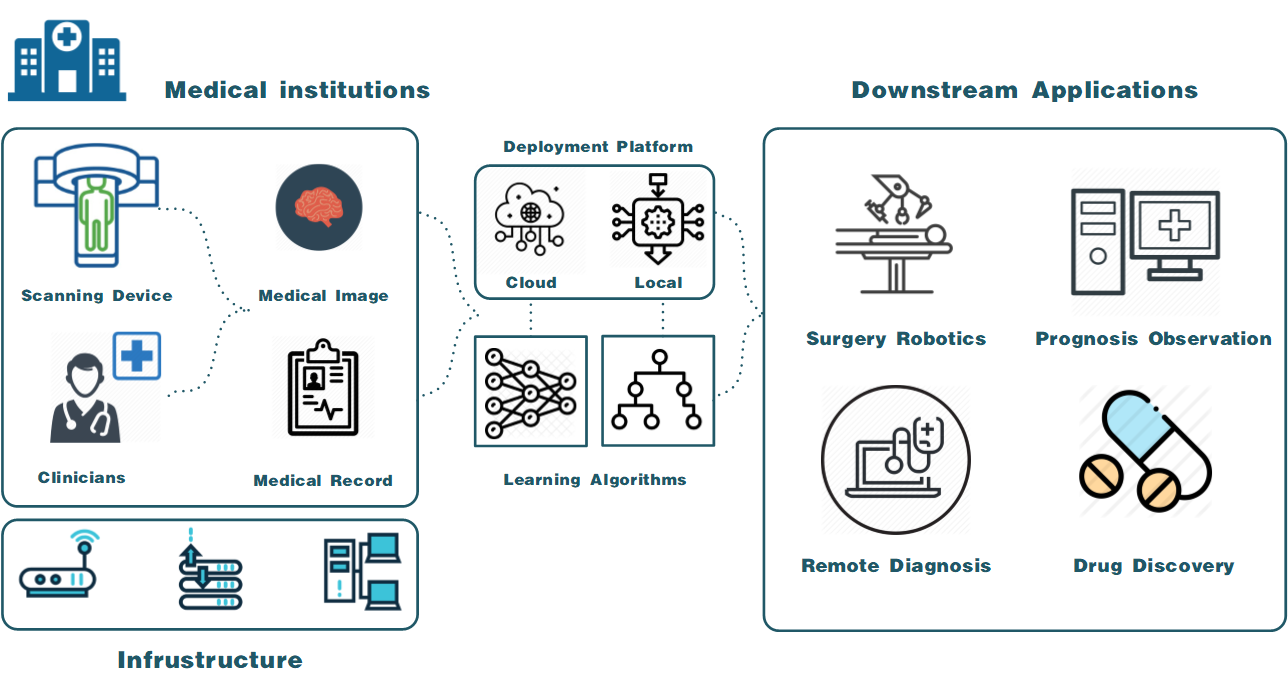}
\caption{A general framework of modern medical image analysis based CAD system}
\label{fig:CAD_System}
\end{figure*}

Computer-Aided Diagnosis (CAD) refers to the use of advanced computer software and hardware to process and analyse medical images, discover and detect lesions and their characteristics, and use the results as a second opinion for physicians' diagnosis reference. The purpose is to help physicians improve the accuracy, efficiency and reproducibility of diagnosis. In a narrow view, CAD mainly refers to a system that can be used for clinical application. The main purpose is to help physicians and doctors find diseases and assist them in judging the degree of diseases, that is, benign or malignant. With the rapid development of modern computing and imaging technology, today's CAD is no longer limited to a simple application system. CAD nowadays has been developed into a complex area involving medicine, computational intelligence, image analysis, data storage and mining. The application scenario of CAD has also been expanded from the early detection and diagnosis of certain diseases to a wider range of fields, such as epidemic screening, genetic diagnosis, and drug discovery. The systematic view of a CAD is shown in Fig. \ref{fig:CAD_System}.

The main working steps of CAD can be roughly divided into three steps:
\begin{enumerate}
\item Image acquisition, that is, to obtain digital images through specific equipment. Physicians usually expect to obtain images with a high signal-to-noise ratio (SNR), high resolution, and high contrast. High SNR refers to the intensity ratio of signal to noise contained in the image. When SNR is high, the interference of noise will be small, the signal transmission quality will be high, and the value of information obtained by imaging will be rich. The resolution has a great influence on the CAD system's performance. High-resolution images will display lesion details. High contrast makes the image better show the details of the unclear sections and further helps the CAD system to make decisions.

\item Feature extraction and quantification, that is, extract and quantify the features in the image through specific algorithms. The extracted features can be referred as pathological manifestations with actual diagnostic values, such as lesion size, density, shape. The extracted features can also be referred as the special coding used as the input of the learning system, such as the Fisher Vector.

\item Training and testing, that is, input the image representation obtained in the second part into a mathematical or statistical algorithm to fit and classify the images. Commonly used traditional machine learning systems include decision trees \cite{azar2013decision} and support vector machines \cite{rojas2017optimal}. In recent years, deep learning based CAD has gradually integrated the second and third steps jointly to achieve better diagnostic performance \cite{khachnaoui2018review}.
\end{enumerate}

\begin{sidewaystable}
\centering
\footnotesize
\begin{tabular}{|c|l|l|l|l|l|}
\hline
No.                & \multicolumn{1}{c|}{Survey Title}                                                                                                                                             & \multicolumn{1}{c|}{Ref.} & \multicolumn{1}{c|}{Year} & \multicolumn{1}{c|}{Venue}                                                                              & \multicolumn{1}{c|}{Content}                                                                                                                            \\ \hline
\multirow{4}{*}{1} & \multirow{4}{*}{\begin{tabular}[c]{@{}l@{}}Computer-Aided Diagnosis in \\ Chest Radiography: A Survey\end{tabular}}                                                           & \multirow{4}{*}{\cite{van2001computer}}         & \multirow{4}{*}{2001}     & \multirow{4}{*}{\begin{tabular}[c]{@{}l@{}}IEEE Trans on \\ Medical Imaging\end{tabular}}               & \multirow{4}{*}{\begin{tabular}[c]{@{}l@{}}A review of CAD in Chest \\ Radiography.\end{tabular}}                                                       \\
                   &                                                                                                                                                                               &                           &                           &                                                                                                         &                                                                                                                                                         \\
                   &                                                                                                                                                                               &                           &                           &                                                                                                         &                                                                                                                                                         \\
                   &                                                                                                                                                                               &                           &                           &                                                                                                         &                                                                                                                                                         \\ \hline
\multirow{4}{*}{2} & \multirow{4}{*}{\begin{tabular}[c]{@{}l@{}}Improve Computer-Aided Diagnosis \\ With Machine Learning Techniques \\ Using Undiagnosed Samples\end{tabular}}                    & \multirow{4}{*}{\cite{li2007improve}}         & \multirow{4}{*}{2007}     & \multirow{4}{*}{\begin{tabular}[c]{@{}l@{}}IEEE Trans on Systems, \\ Man, and Cybernetics\end{tabular}} & \multirow{4}{*}{\begin{tabular}[c]{@{}l@{}}A semi-supervised learning method \\ for improving CAD accuracy in \\ breast cancer detection.\end{tabular}} \\
                   &                                                                                                                                                                               &                           &                           &                                                                                                         &                                                                                                                                                         \\
                   &                                                                                                                                                                               &                           &                           &                                                                                                         &                                                                                                                                                         \\
                   &                                                                                                                                                                               &                           &                           &                                                                                                         &                                                                                                                                                         \\ \hline
\multirow{4}{*}{3} & \multirow{4}{*}{\begin{tabular}[c]{@{}l@{}}Computer-Aided Diagnosis in Medical \\ Imaging:Historical Review, Current \\ Status and Future Potential\end{tabular}}             & \multirow{4}{*}{\cite{doi2007computer}}         & \multirow{4}{*}{2007}     & \multirow{4}{*}{\begin{tabular}[c]{@{}l@{}}Computerized Medical\\ Imaging and Graphics\end{tabular}}    & \multirow{4}{*}{A review of PACS based CAD systems.}                                                                                                    \\
                   &                                                                                                                                                                               &                           &                           &                                                                                                         &                                                                                                                                                         \\
                   &                                                                                                                                                                               &                           &                           &                                                                                                         &                                                                                                                                                         \\
                   &                                                                                                                                                                               &                           &                           &                                                                                                         &                                                                                                                                                         \\ \hline
\multirow{4}{*}{4} & \multirow{4}{*}{\begin{tabular}[c]{@{}l@{}}Computer-Aided Diagnosis System \\ Based on Fuzzy Logic for Breast \\ Cancer Categorization\end{tabular}}                          & \multirow{4}{*}{\cite{miranda2015computer}}         & \multirow{4}{*}{2014}     & \multirow{4}{*}{\begin{tabular}[c]{@{}l@{}}Computers in Biology\\ and Medicine\end{tabular}}            & \multirow{4}{*}{\begin{tabular}[c]{@{}l@{}}A review of fuzzy logic based CAD \\ in breast cancer categorization.\end{tabular}}                          \\
                   &                                                                                                                                                                               &                           &                           &                                                                                                         &                                                                                                                                                         \\
                   &                                                                                                                                                                               &                           &                           &                                                                                                         &                                                                                                                                                         \\
                   &                                                                                                                                                                               &                           &                           &                                                                                                         &                                                                                                                                                         \\ \hline
\multirow{4}{*}{5} & \multirow{4}{*}{\begin{tabular}[c]{@{}l@{}}Image Based Computer Aided Diagnosis \\ System for Cancer Detection\end{tabular}}                                                  & \multirow{4}{*}{\cite{lee2015image}}         & \multirow{4}{*}{2014}     & \multirow{4}{*}{\begin{tabular}[c]{@{}l@{}}Expert Systems with\\ Applications\end{tabular}}             & \multirow{4}{*}{\begin{tabular}[c]{@{}l@{}}Systematic review of imaging \\ based cancer detection system.\end{tabular}}                                 \\
                   &                                                                                                                                                                               &                           &                           &                                                                                                         &                                                                                                                                                         \\
                   &                                                                                                                                                                               &                           &                           &                                                                                                         &                                                                                                                                                         \\
                   &                                                                                                                                                                               &                           &                           &                                                                                                         &                                                                                                                                                         \\ \hline
\multirow{4}{*}{6} & \multirow{4}{*}{\begin{tabular}[c]{@{}l@{}}Clinical Evaluation of A Computer-Aided \\ Diagnosis System for Determining \\ Cancer Aggressiveness in Prostate MRI\end{tabular}} & \multirow{4}{*}{\cite{litjens2015clinical}}         & \multirow{4}{*}{2015}     & \multirow{4}{*}{European Radiology}                                                                     & \multirow{4}{*}{\begin{tabular}[c]{@{}l@{}}Investigat the added value of \\ CAD on cancer diagnostic accuracy.\end{tabular}}                            \\
                   &                                                                                                                                                                               &                           &                           &                                                                                                         &                                                                                                                                                         \\
                   &                                                                                                                                                                               &                           &                           &                                                                                                         &                                                                                                                                                         \\
                   &                                                                                                                                                                               &                           &                           &                                                                                                         &                                                                                                                                                         \\ \hline
\multirow{4}{*}{7} & \multirow{4}{*}{\begin{tabular}[c]{@{}l@{}}Clinically Applicable Deep Learning \\ for Diagnosis and Referral in \\ Retinal Disease\end{tabular}}                              & \multirow{4}{*}{\cite{de2018clinically}}         & \multirow{4}{*}{2018}     & \multirow{4}{*}{Nature Medicine}                                                                        & \multirow{4}{*}{\begin{tabular}[c]{@{}l@{}}A deep learning based CAD on \\ retinal disease segmentation.\end{tabular}}                                  \\
                   &                                                                                                                                                                               &                           &                           &                                                                                                         &                                                                                                                                                         \\
                   &                                                                                                                                                                               &                           &                           &                                                                                                         &                                                                                                                                                         \\
                   &                                                                                                                                                                               &                           &                           &                                                                                                         &                                                                                                                                                         \\ \hline
\multirow{4}{*}{8} & \multirow{4}{*}{\begin{tabular}[c]{@{}l@{}}Artificial Intelligence and Computer-Aided \\ Diagnosis in Colonoscopy: Current Evidence \\ and Future Directions.\end{tabular}}   & \multirow{4}{*}{\cite{ahmad2019artificial}}         & \multirow{4}{*}{2019}     & \multirow{4}{*}{\begin{tabular}[c]{@{}l@{}}The Lancet Gastroenterology\\ \& Hepatology\end{tabular}}    & \multirow{4}{*}{\begin{tabular}[c]{@{}l@{}}A review of modern AI based \\ CAD in colonoscopy.\end{tabular}}                                             \\
                   &                                                                                                                                                                               &                           &                           &                                                                                                         &                                                                                                                                                         \\
                   &                                                                                                                                                                               &                           &                           &                                                                                                         &                                                                                                                                                         \\
                   &                                                                                                                                                                               &                           &                           &                                                                                                         &                                                                                                                                                         \\ \hline
\end{tabular}
\makeatletter\def\@captype{table}\makeatother\caption{Summary of CAD related review articles.}
\label{Tab: CAD_Review}
\end{sidewaystable}

For different physiological systems or diseases, the deployment and application of CAD could be different. For this reason, in Table \ref{Tab: CAD_Review}, we have listed some selected CAD survey articles, which relate to different physiological systems or diseases in recent years. Table \ref{Tab: CAD_Review} aims to help readers to understand the needs, current situation and development trends of different CAD systems in different fields.

As the core component of CAD, learning algorithm based medical image analysis has been developing rapidly. Based on the paradigm of feature extraction, we divide the learning algorithm based medical imaging analysis into two categories: medical image analysis with hand-crafted features and medical image analysis with deep learning. The rest two subsections of this chapter will discuss them respectively.

\section{Medical Image Analysis with Hand-Crafted Features}
\begin{figure*}[t]
\includegraphics[width=1\textwidth]{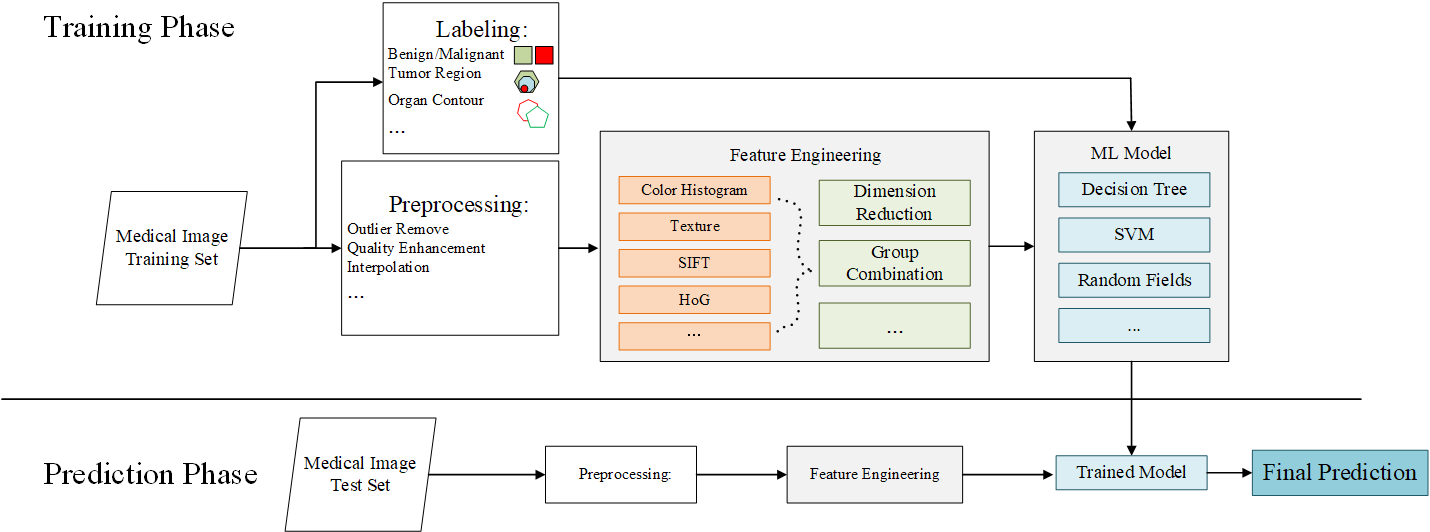}
\caption{The general pipeline of traditional machine learning based medical image analysis.}
\label{fig:ml_workflow}
\end{figure*}

In the early stage of medical image analysis based on traditional machine learning, the most critical is the feature extraction and optimization, which is also called feature engineering. The quality of feature extraction and optimization directly affects the performance of the downstream task-oriented models. A general workflow of traditional machine learning based medical image analysis is shown in Fig. \ref{fig:ml_workflow}. 

Compared with natural images captured by digital cameras, medical images are very different in visual perception, and their features are also very specific:

\begin{itemize}
\item Local similarity. In medical images, the components in a small region usually have similar appearance and structure. Sometimes there is little difference between images of diseased and healthy tissues.
\item	Low brightness and contrast. Due to the limitations of imaging methods and equipment properties, medical images generally have low brightness and contrast, and color changes are not very obvious. This will cause an insufficient number of features or unrepresentative features.
\item The characteristics are complex. Most medical images come from different imaging equipment, different patients, and different shooting times and environments. During the image generation, it is difficult to avoid adverse effects such as changes in illumination during shooting, individual physiological differences of patients, the influence of scanning positions and angles, and noise from equipment. Therefore, medical images often show great complexity and variability. Some methods that are applicable in one image may not be carried out in other images.
\end{itemize}

To solve the aforementioned problems, early feature engineering for medical image analysis tends to satisfy the following conditions. 
\begin{enumerate}
\item Features can make an accurate description of images. From images with similar textures and structures, features should be distinguishable and highly repetitive.
\item Feature extraction algorithm is required to have high stability and adaptability and have a certain degree of robustness to various unfavorable factors and complex environments that may appear.
\item In some applications, selected medical image analysis tools have strict requirements on processing time \cite{netsch2001towards}. Therefore the image feature extraction algorithm should be easy to implement, and the processing speed must be fast, and the computational complexity should be low.
\end{enumerate}

Features can be generally divided into global and local features. Global features describe the overall properties of the whole image or a large region of interest. Common global features include color or grey distribution and texture structure. Local features only describe some representative information such as special points, lines in a small patch of an image, or a small region of interest. The selection and combination of features is a key point of feature engineering. For different types of images and application requirements, selection and combination of features can be different.

\textbf{Global features.}  Global feature is the overall description of the whole image or a large region of interest. It is usually extracted from a certain mathematical method to count the color or texture information of all pixels.

The color feature is the simplest and most intuitive description of an image. It is usually generated from a set of statistical functions. The color feature is used to reflect the color information of the image. color histogram is the most common feature. The color distribution function is obtained by calculating the color value of all pixels in the color space and counting the frequency of each color value in the image. Color histogram is simple, fast to calculate, and will not be affected by image rotation and scale changes. It has been applied in the field of image retrieval in an early stage \cite{han2002fuzzy}. In recent years, many related studies have shown that color features can still be used as an effective supplement, combined with other advanced features, and play an important role in various application fields such as image matching \cite{li2002tongue}, medical image segmentation \cite{wei1997real}, and disease classification \cite{arjunan2009image}. However, color histogram has disadvantages. It only counts all the color values within the picture. Thus it is difficult to effectively reflect the information of the image itself, such as edge and texture. At the same time, color histogram is sensitive to noise and has low robustness.

The texture feature represents the structure that repeatedly appears in the image. Compared with color features, the calculation of texture features is not limited to the color value of a single pixel. Instead, it counts complex information such as intensity distribution, neighborhood relations, etc. The smallest texture structure is called a primitive. The texture feature is the composition of the texture primitive in the image and the frequency of its repetition. Texture feature is the most widely used global feature in image analysis. Commonly used texture feature extraction methods include statistics-based methods \cite{tesavr2008medical}, geometry-based methods \cite{patil2011geometrical}, model-based methods \cite{sertel2009histopathological}, and signal processing-based methods \cite{nanni2010local}. The texture reflects the semantic information contained in the image to a certain extent, but the texture also has its shortcomings. First of all, the texture is very sensitive to image resolution, and the calculated texture of the same object at different resolutions could be quite different. Second, when reflection or noise interferes with the image, the calculated texture will contain wrong information and can mislead the model.

\textbf{Local features.} The core idea of local features is that the image is divisible. It assumes that an image is a collection of many regions with different characteristics. Among them, the more prominent region (that is, the closest part to the observer) is called the foreground, and the remaining area is called the background. Generally, the foreground information reflects the main subject of the image, that is, the main semantic information contained in the image. In the research based on global features, people have found that the undifferentiated description of the entire image could not allow the algorithm to understand the semantic information of the image. Therefore, researchers have begun to search for an alternative description and proposed several local feature representations.

The SIFT operator proposed by Lowe \cite{lowe1999object} is a milestone in the field of local feature extraction. SIFT uses the Gaussian differential pyramid to approximate the extreme points of the Gaussian Laplacian space, which solves the problem of the latter’s calculation difficulties and realizes the extreme value in the multi-scale space. The feature points can be detected with invariant scales, and then obtains rotation invariance by determining the main direction of feature sampling points. On this basis, various local feature extraction methods have been proposed. The PCA-SIFT \cite{ke2004pca} operator uses the principal component analysis method to reduce the dimension of the SIFT descriptors to increase the calculation speed. GLOH (gradient location and orientation histogram) operator \cite{mikolajczyk2005performance} adjustment the shape of the sampling window in the calculation of the feature descriptor in the SIFT algorithm is used, and a radial circle is used to replace the original grid. The SUFR (Speeded-up robust features)\cite{bay2008speeded} operator is the most successful improvement method to SIFT, which uses the Harris space for extreme value detection, and then calculates the Haar wavelet feature acquisition descriptor in the neighborhood of feature points, which greatly reduces the complexity and computing time of feature extraction. It is worth noting that the SIFT operator can usually achieve good results in natural images. When directly applied to medical images, it may be due to problems such as uniform intensities, lack of obvious edges, etc. For example, some researchers use pre-set sampling points to omit the process of interest point detection \cite{detone2018superpoint}. These methods require prior knowledge to determine the size of the sampling window, usually without multi-scale transformation and extreme value detection, which are suitable for medical image feature extraction. However, when the edges of objects in the image are invisible, SIFT often fails to extract accurate feature points and results in poor performance.

In summary, the early learning systems used for medical image analysis were mostly based on manual feature engineering. The extraction and quantity method of manually defined features are important factors that determine the performance of the learning system. The traditional learning system has the advantages of simple structure and convenient implementation, but it also has disadvantages such as feature engineering cannot be jointly trained with downstream models, feature representation is simple, and higher-dimensional information representation cannot be learned. These shortcomings make the performance of the feature engineering based learning system is unsatisfactory in various medical image analysis tasks.

\section{Medical Image Analysis with Deep Learning}

In recent years, the application of traditional learning methods to medical image analysis has mainly faced two major problems. One is that the medical image data to be processed has a higher dimensionality and requires a model with stronger learning and adaptability. The second is that medical image big data is more fragmented, and the data structure is more complex, often requiring the integration of different information. When facing these demands, traditional artificial feature engineering is particularly weak. The main disadvantages are as follows:

\begin{itemize}
\item Manually selected features contain very limited content. Manually selected features are often limited to visible features, such as grayscale, color, and edges. Or inspired by the visual model, it is limited to relatively simple implicit expressions, such as HOG \cite{dalal2005histograms} and SIFT \cite{lowe1999object}. These feature extraction algorithms have the advantages of fast speed, low resource consumption, and high interpretability. However, with the increase in the complexity of tasks, researchers began to focus on how to effectively extract features with more expressive information, such as semantic information and attention information.

\item The types of manually selected features are limited. The complexity of feature engineering is proportional to the number of features. If too many types of features are selected, the corresponding computing will be time-consuming. If the selected feature types are too few, the feature space dimension is too low, and the data cannot be completely and effectively described.

\item Manually selected feature combination and optimization rely heavily on expert experience. Different features have different descriptions of data. Effective feature combination and optimization can fully explore and utilize the relationship between features, thus improving the performance of the model. However, this process relies heavily on the developer's experience and knowledge.

\item Feature engineering and downstream models cannot be updated jointly. In most traditional learning systems, the tuning of feature engineering is performed separately from the tuning of the downstream model. This makes it difficult to map the performance error of the model back to feature engineering and is also one of the reasons why most traditional learning systems have relatively mediocre performance.
\end{itemize}

\begin{figure*}[t]
\centering
\includegraphics[width=0.5\textwidth]{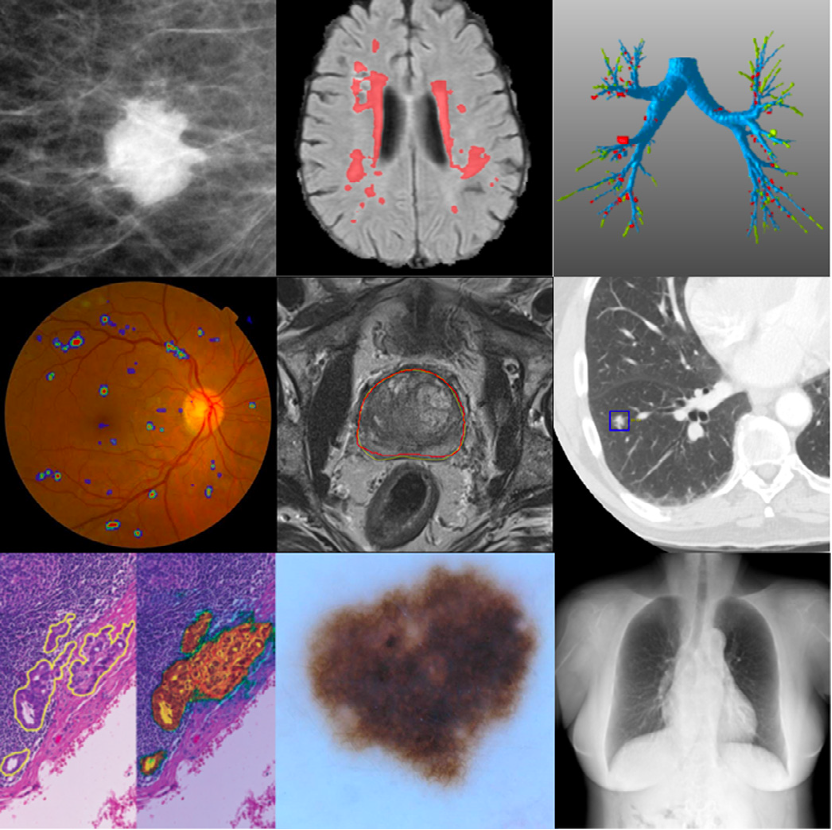}
\caption{Selection of some medical imaging analysis applications with state-of-the-art performance generated by deep learning frameworks. From top-left to bottom-right: Mammographic mass classification ( Kooi et al. \cite{kooi2017large} ). Brain white matter hyperintensity segmentation (Ghafoorian et al. \cite{ghafoorian2016non}). Air tree segmentation with leak localization ( Charbonnier et al. \cite{charbonnier2017improving} ).  Retinopathy classification (Kaggle Diabetic Retinopathy challenge 2015, van Grinsven et al. \cite{van2016fast}). Prostate segmentation (top rank in PROMISE12 challenge \cite{litjens2014evaluation}). Nodule classification (top ranking in LUNA16 challenge \cite{setio2017validation}). Breast cancer detection (top ranking and human expert performance in CAMELYON16 \cite{bejnordi2017diagnostic}). Skin lesion classification ( Esteva et al. \cite{esteva2017dermatologist}). Bone suppression (Yang et al. \cite{yang2017cascade}). Image courtesy of \citep{litjens2017survey}}
\label{fig:selected_problems}
\end{figure*}

Therefore, how to automatically learn high-level feature information from data, and to optimize the model and feature engineering together, has become the focus of attention of researchers in recent years, and has also become a focus of attention in the industry.

Deep learning is a new branch field and developing rapidly from traditional machine learning systems. It aims to automatically learn high-level discriminative features of various levels from data by simulating the human neural network perception. Since Hinton proposed a multi-layer restricted Boltzmann machine based on a probabilistic graph model in 2006 \cite{hinton2012practical}, deep learning has become a dominant tool in various fields including computer vision. In recent years, deep learning has achieved significant success in image recognition, speech recognition, natural language processing and other fields. Deep learning has triggered a wave of data mining and analysis in broader fields. In the field of medical image analysis, deep learning has also gained the attention of academia and industry.

\begin{figure*}[t]
\includegraphics[width=1\textwidth]{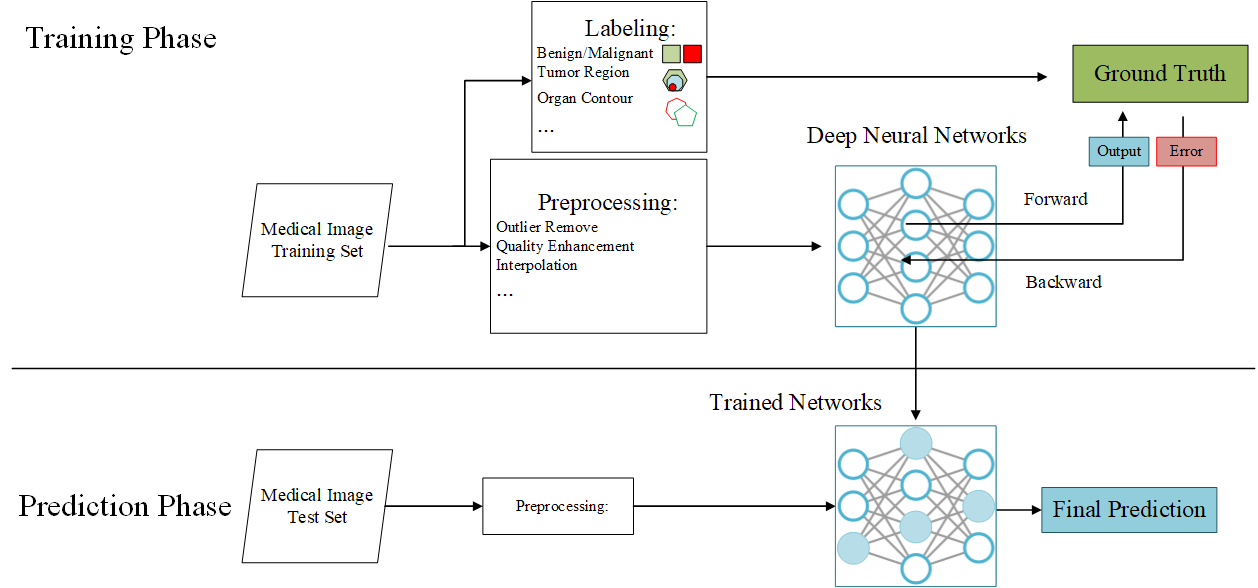}
\caption{The general pipeline of deep learning based medical image analysis.}
\label{fig:dl_workflow}
\end{figure*}

Deep learning was originally developed from artificial neural networks. In the 1980s, the BP algorithm for artificial neural networks was proposed \cite{rumelhart1986learning}, which started the upsurge of machine learning based on statistical learning. However, in the subsequent training process, it was found that the BP algorithm has some defects such as slow convergence speed and easy to fall into a local minimum. In the 1990s, shallow machine learning models such as boosting and SVM were proposed. These models have been successfully achieved in theory and application, making shallow machine learning popular for a long time. By 2006, the introduction of the Deep Belief Network (DBN) opened a new chapter in modern deep learning research \cite{hinton2006fast}. In 2012, Hinton used the CNN model to win the ImageNet challenge with an accuracy rate of more than 10 percent higher than the runner-up. This made a breakthrough in the field of computer vision. Since then, with the emergence of models such as recurrent neural networks (RNN) for sequence data modeling, deep residual networks for image processing, the improvement of GPU computing power, deep learning has been achieved great success in various fields. A general view of the deep learning system workflow is shown in Fig. \ref{fig:dl_workflow}.

Deep learning can learn high dimensional discriminative features by building a multiple hidden layer learning model with massive training data to improve the accuracy of classification or prediction. Compared with traditional machine learning, deep learning has the following advantages:

\begin{enumerate}
\item Automated feature learning. Deep learning methods can automatically learn the high dimensional discriminative feature representations from massive data according to different applications, and can better express the internal information of the data.

\item High generalization and transferability. A deep learning model structure is usually with 5 or more hidden layers, including more nonlinear transformations, which greatly enhances the ability to fit complex functions. The deep network can be applied to different tasks. The trained model can be reused through strategies such as transfer learning.

\end{enumerate}

Deep learning systems have been widely applied in medical image analysis on different tasks. Traditional machine learning based medical image analysis method is mostly based on multi-feature fusion, singular value decomposition and wavelet transform methods. Compared with traditional machine learning based medical image analysis, deep learning can model nonlinear relationships in medical images with higher feature extraction efficiency. In recent years, many research works have been proposed to apply deep learning on different medical image analysis tasks and these works have provided an important baseline for further clinical research (Fig. \ref{fig:selected_problems}). There are two basic medical image analysis tasks: medical image classification and segmentation. Medical image classification is carried out to determine whether a sample is sick or how severe it is, while medical image segmentation is the localization of a certain lesion and other parts of a medical image. At present, deep learning systems are widely used in the above two fields. At the same time, deep learning has also been widely used in other medical image analysis areas such as medical image registration. Due to limited space, we only review related works on medical image classification and segmentation. However, we select some representative state-of-the-art methods in various medical image analysis tasks and summarize them in Table \ref{Tab: DLMIA_Review}. The rest of this section mainly introduces the research progress of deep learning in medical image analysis in two aspects: disease classification and medical image segmentation.

\textbf{Medical Image Classification} is one of the earliest applications of deep learning in the field of medical image analysis. It refers to taking one or more modality images as input, processing it through a trained model, and outputting one label to indicate whether a patient has a certain disease or the severity degree of the disease. In the early stage, deep learning models focused on SAE, DBN, and DBM networks with unsupervised pre-training methods. The research mainly focuses on the analysis of neuroimaging, such as the diagnosis of Alzheimer’s disease (AD) or Mild Cognitive Impairment (MCI). These algorithms usually use multi-modal images as input to extract complementary feature information in modalities such as MRI, PET, and CSF. Suk et al. \cite{suk2013deep} used DBM and SAE to find the expression of potential hierarchical features from 3D neuroimaging images and constructed AD/MCI classification models. The results verified on the ADNI dataset \cite{petersen2010alzheimer} show that the classification performance of the proposed model using SAE is better than using DBM. There are also a small amount of medical image classification research based on unsupervised models. For example, Rahhal et al. \cite{al2018convolutional} used SSAE to learn features in a weakly-supervised manner to classify ECG signals. Abdel-Zaher et al. \cite{abdel2016breast} first tried unsupervised learning on DBN, and then used feedback supervised learning to adjust the network to classify the breast cancer from the Wisconsin dataset. Through the application of DBN and SAE, the performance of various medical image classification tasks has been improved to a certain extent. However, DBN and SAE have disadvantages such as slow convergence speed, long learning time, and easy to fall into local minima.

Nowadays, CNN (Convolutional Neural Network) is gradually becoming the standard technology in image classification. Arevalo et al. \cite{arevalo2015convolutional} proposed a feature learning framework for breast cancer diagnosis. The author applied CNN learning distinguishing features and classifying mammogram lesions. Kooi et al. \cite{kooi2017large} compared the manual design features and automatically extracted features from CNN in CAD. Both of these methods were trained on a large data set of about 45,000 mammograms. The results showed that CNN is superior to traditional manual feature extraction methods with low sensitivity. Xu et al. \cite{xu2016detecting} studied the use of deep CNN to automatically extract features, combined with multi-instance learning methods, to classify histopathological images of colon cancer in the case of few manual annotations. Gao et al. \cite{gao2015mci} discussed the importance of deep learning technology for brain CT image classification, especially the use of CNN to provide supplementary information for early diagnosis of AD. Payan et al. \cite{payan2015predicting} and Hosseiniasl et al. \cite{hosseini2016alzheimer} used 3D CNN to diagnose AD on neuroimaging. Some works combine CNN with RNN. For example, Gao et al. \cite{gao2015automatic} used CNN to extract the low-level local feature information in the slit lamp image, combined with RNN to further extract high-level features and classified nuclear cataracts. CNN has greatly improved the performance of deep models on various tasks, but CNN also has other shortcomings. For example, the pooling layer will lose local features and ignore the correlation between the local image patch and the global image. Also, CNN models lack Interpretability.

\textbf{Medical Image Segmentation} The segmentation of organs and their substructures in medical images has important clinical significance. On the one hand, accurate segmentation can be used to quantitatively analyze clinical parameters related to volume and shape, such as the ventricular volume and contraction ejection rate of the heart. On the other hand, when using radiotherapy technology to treat tumors, accurately segmenting the tumor can ensure that tumor cells are killed during the treatment while protecting normal tissues and organs. An accurate segmentation methodology usually needs to combine multi-modal image information and adaptive context information. Therefore, most of the current studies use multi-modal image information as the network model input, or use multi-scale stream networks, or even 3D kernels to directly extract features. Kamnitsas et al. \cite{kamnitsas2017efficient} used a multi-scale fully 3D CNN network to combine global and local contextual information. The results demonstrated excellent performance in the various challenging segmentation tasks including traumatic brain injury, glioma tumors and ischemic stroke lesion. The performance, especially in terms of the overall segmentation level of glioma tumors, has surpassed the level of human experts. Yu et al. \cite{yu2016automated} combined the residual connection and the fully convolutional network to construct a deep residual FCN network, which automatically segmented melanoma in the dermoscopic image and won second place in the ISBI2016 challenge \cite{li2018skin}.  The U-Net \cite{ronneberger2015u} based framework has also been adopted by many researchers. For example, Choi et al. \cite{choi2019brain} used two pathway CNN to improve the brain tissue segmentation accuracy. The global pathway with a large size kernel determines the approximate position of the striatum, and the local pathway with small size kernels is used to predict all voxel labels. This method achieves the current state-of-the-art performance with Dice similarity coefficients 0.893 in the segmentation of the brain striatum structure. Moeskops et al. \cite{moeskops2016automatic} used a multi-scale CNN method for brain tissue segmentation. This method achieves the best results on 8 tissue classifications. It was verified on 5 different age data sets. The Dice similarity coefficients of the segmentation results were 0.87, 0.82, 0.84, 0.86 and 0.91 respectively. These works use manually designed CNN network structures to effectively learn the features of the target to be segmented. 

In summary, most of the deep learning systems used for medical image analysis are based on deep neural networks. Deep learning can effectively use data to learn high-dimensional discriminative features. This automated feature learning process eliminates the intensive steps of manual feature engineering, enables model tuning and feature learning to be jointly trained, and greatly improves the performance of various medical image analysis tasks. However, in the aforementioned works, there is no work to model and utilize the relational information between different tumor regions or tissues. In contrast, we will propose our novel deep relational learning work in Chapters \ref{Chapter3} and \ref{Chapter4}. Our goal is to improve the accuracy of semantic segmentation by learning the relational information between different semantic regions.

\begin{sidewaystable}
\centering
\footnotesize
\begin{tabular}{|llll|}
\hline
\multicolumn{1}{|c|}{Ref}                       & \multicolumn{1}{c|}{Method} & \multicolumn{1}{c|}{Application} & \multicolumn{1}{c|}{Highlights}                                                   \\ \hline
\multicolumn{4}{|c|}{Disorder Classification}                                                                                                                                                        \\ \hline
Suk and Shen \cite{suk2013deep}           & SAE                         & AD/MCI Classification            & \begin{tabular}[c]{@{}l@{}}Using stacked auto-encoders with supervised fine-tuning for\\ AD/MCI classification\end{tabular} \\ \hline
Hosseini-Asl et al. \cite{hosseini2016alzheimer}    & CNN                         & AD/MCI/HC Classification         & Using 3D CNN to process 3D fMRI data for multi-stage classification.              \\ \hline
Suk and Shen et al. \cite{suk2016deep}    & CNN                         & AD/MCI/HC Classification         & CNN with sparse representations for disorder classification.                      \\ \hline
Lian et al. \cite{lian2018hierarchical}            & FCN                         & AD/MCI/HC Classification         & Hierachically FCN with location proposal for disorder classification.             \\ \hline
\multicolumn{4}{|c|}{Tumor Segmentation}                                                                                                                                                             \\ \hline
Havaei et al. \cite{havaei2017brain}          & CNN                         & Glioma Segmentation              & \begin{tabular}[c]{@{}l@{}}Two-path way CNN with variant path combination for \\ glioma segmentation.\end{tabular}           \\ \hline
Pereira et al. \cite{pereira2016brain}         & CNN                         & Glioma Segmentation              & CNN with multi-modality inputs for MRI glioma segmentation.                       \\ \hline
Kamnitsas et al. \cite{kamnitsas2017efficient}       & FCN                         & Tumor segmentation               & 3D FCN with post-processing CRF for various tumor segmentation.                   \\ \hline
Wang et al. \cite{wang2017automatic}            & FCN                         & Glioma Segmentaion               & Cascaded FCN for progressive glioma region segmentation.                          \\ \hline
Myronenko \cite{myronenko20183d}              & FCN                         & Glioma Segmentation              & \begin{tabular}[c]{@{}l@{}}3D FCN with additional autoencoder as regularization for \\ glioma segmentation.\end{tabular}     \\ \hline
\multicolumn{4}{|c|}{Retinal Image Analysis}                                                                                                                                                         \\ \hline
Fu et al. \cite{fu2016deepvessel}              & CNN                         & Blood Vessel Segmentation        & \begin{tabular}[c]{@{}l@{}}CNN with CRF to capture long-range information for \\ blood vessel segmentation.\end{tabular}     \\ \hline
Wu et al. \cite{wu2016deep}              & CNN                         & Blood Vessel Segmentation        & Patch-based CNN with PCA for last layer feature maps.                             \\ \hline
Zilly et al. \cite{zilly2017glaucoma}           & CNN                         & Optic Disk Segmentation          & CNN with boosting for kernel updating.                                            \\ \hline
\multicolumn{4}{|c|}{Chest X-Ray Image Analysis}                                                                                                                                                     \\ \hline
Bar et al. \cite{bar2015deep}             & CNN                         & Pathology Detection              & Pre-trained CNN with low level features for diseases detection.                   \\ \hline
Cicero et al. \cite{cicero2017training}          & CNN                         & Pathology Detection              & Using GoogleNet for large scale diseases detection validation.                    \\ \hline
Hwang et al. \cite{hwang2016novel}           & CNN                         & Tuberculosis Detection           & CNN for tuberculosis detection with entire radiographs as input.                  \\ \hline
\multicolumn{4}{|c|}{Breast Image Analysis}                                                                                                                                                          \\ \hline
Huynh et al. \cite{huynh2016digital}           & CNN                         & Mass Classification              & Pre-trained network for mass classification.                                      \\ \hline
Akselrod-Ballin et al. \cite{akselrod2016region} & RCNN                        & Mass Classification              & A region proposal based mass localization and classification.                     \\ \hline
Wang et al. \cite{wang2017detecting}                                     & CNN                         & Vessel Calcification             & A detection network based vessel calcification.                                   \\ \hline
Dalmis et al. \cite{dalmics2017using}          & CNN                         & Tissue Segmentation              & A deep CNN based breast and tissue segmentation.                                  \\ \hline
\multicolumn{4}{|c|}{Digital Pathology Image Analysis}                                                                                                                                               \\ \hline
Gao et al. \cite{gao2016hep}             & CNN                         & Nucleus Classification           & CNN based Hep2-cells classification.                                              \\ \hline
Janowczyk et al. \cite{janowczyk2017stain}       & DNN                         & Nucleus Segmentation             & Deep neural network based segmentation with resolution adaptive.                  \\ \hline
Xu and Huang \cite{xu2016detecting}           & DNN                         & Nucleus Detection                & DNN based cell detection with whole-slide image as input.                         \\ \hline
\end{tabular}
\makeatletter\def\@captype{table}\makeatother\caption{Summary of CAD related review articles.}
\label{Tab: DLMIA_Review}
\end{sidewaystable}

\chapter{Context Aware Network for 3D Glioma Segmentation} 
\label{Chapter3} 
\section{Introduction}
Glioma is one of the most common primary brain tumors with fateful health damage impacts and high mortality. To provide sufficient evidence for early diagnosis, surgery planning and post-surgery observation, Magnetic Resonance Imaging (MRI) is a widely used technique to provide reproducible and non-invasive measurement, including structural, anatomical and functional characteristics. Different 3D MRI modalities, such as T1, T1 with contrast-enhanced (T1ce), T2 and Fluid Attenuation Inversion Recover (FLAIR), can be used to examine different biological tissues.

\begin{figure}[t]
    \centering
    \includegraphics[width=0.6\textwidth]{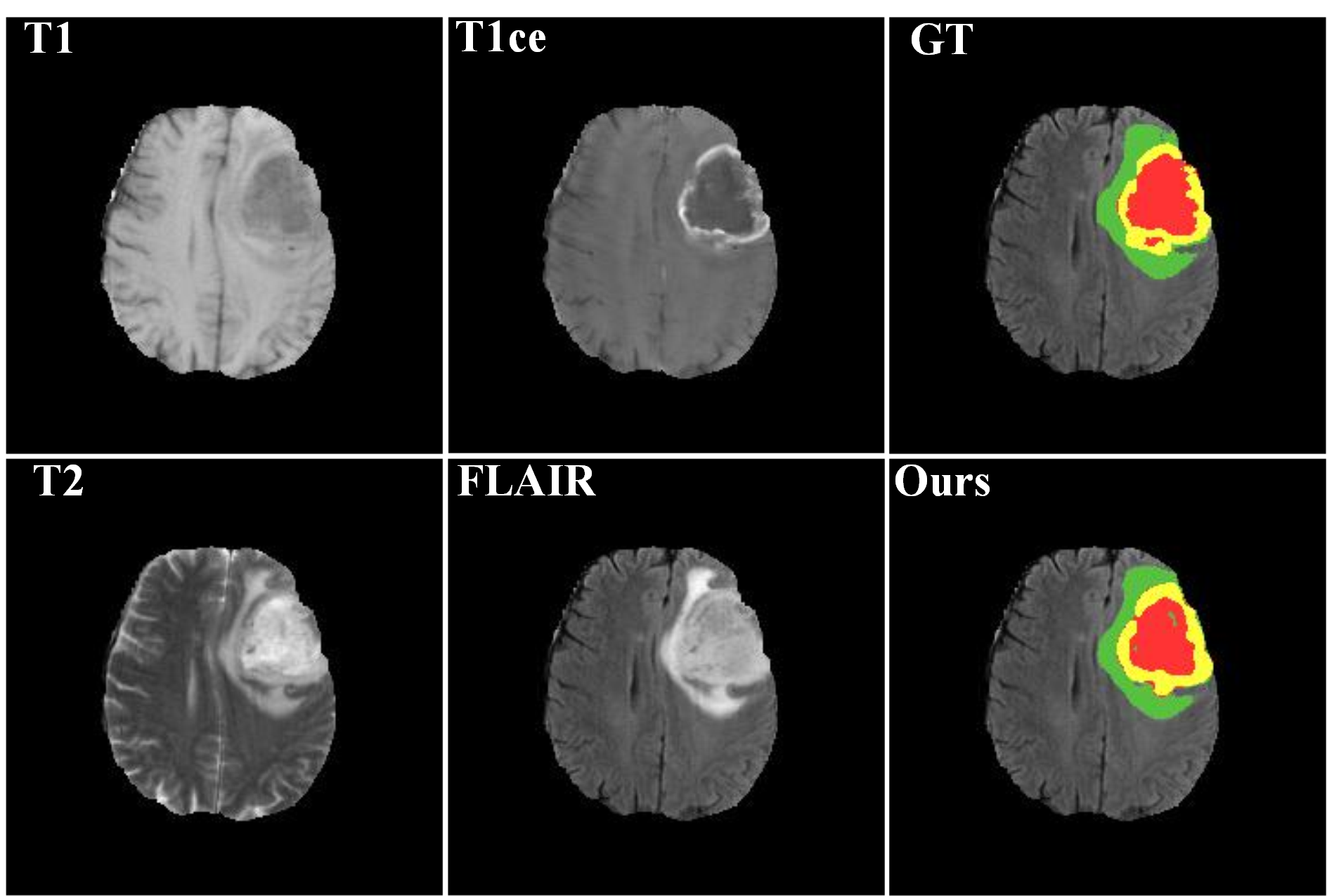}
    \caption{Examples of multi-modality data slices from BraTS17 with ground-truth and our segmentation result. In this figure, green represents GD-Enhancing Tumor, yellow represents Pertumoral Edema and red represents NCR$\backslash$ECT.}
\label{fig:task}
\end{figure}

Medical image segmentation provides fundamental guidance and quantitative assessment for medical professionals to achieve disease diagnosis, treatment planning and follow-up services. However, manual segmentation requires certain professional expertise and usually tends to be time and labor-consuming. Fig. \ref{fig:task} shows a general view of the brain tumor segmentation task. Early research on automated brain tumor segmentation was based on traditional machine learning algorithms \cite{bauer2012segmentation, subbanna2012probabilistic, shin2012hybrid, festa2013automatic}, which rely on hand-crafted features, such as textures \cite{reza2013multi} and local histograms \cite{goetz2014extremely}. However, finding the best hand-crafted features or optimal feature combinations in a high dimensional feature space is impracticable. In recent years, deep learning techniques, especially deep convolutional neural networks (DCNNs), can be used to effectively learn high dimensional discriminative features from data and have been widely used on various computer vision tasks \cite{long2015fully}.

Inter-class ambiguity is a common issue in brain tumor segmentation. This issue makes it hard to achieve accurate dense voxel-wise segmentation if only considering isolated voxels, as different classes' voxels may share similar intensity values or close feature representations. To address this issue, we propose a context-aware network, namely CANet, to achieve accurate dense voxel-wise brain tumor segmentation in MRI images. The proposed CANet contains a novel Hybrid Context Aware Feature Extractor (HCA-FE) and a novel Context Guided Attentive Conditional Random Field (CG-ACRF). Our contributions in this work are summarised below:

\begin{itemize}
    \item We propose a novel HCA-FE built with a 3D feature interaction graph neural network and a 3D encoder-decoder convolutional neural network. Different from previous works that usually extract features in the convolutional space, HCA-FE learns hybrid context guided features both in a convolutional space and a feature interaction graph space (the relationship between neighboring feature nodes is utilised and continuously updated). To our knowledge, this is the first practice on brain tumor segmentation, which incorporates adaptive contextual information with graph convolution updates.
    \item We further propose a novel CG-ACRF based fusion module that attentively aggregates features from the feature interaction graph and convolutional spaces. Moreover, we formulate the mean-field approximation of the inference in the proposed CG-ACRF as a convolution operation, enabling the CG-ACRF to be embedded within any deep neural network seamlessly to achieve end-to-end training.
    \item We conduct extensive evaluations and demonstrate that our proposed CANet outperforms several state-of-the-art technologies using different measure metrics on the Multimodal Brain Tumor Image Segmentation Challenge (BraTS) datasets, i.e. BraTS2017 and BraTS2018.
\end{itemize}

\begin{figure*}[t]
    \centering
    \includegraphics[width=\textwidth]{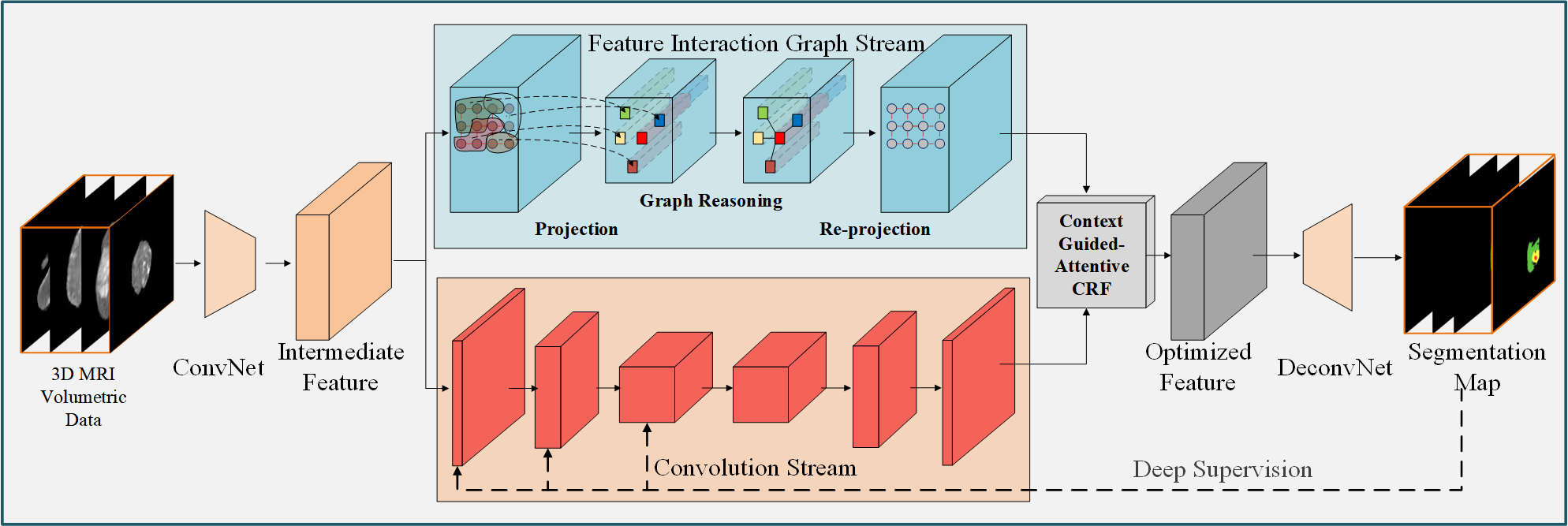}
    \caption{The architecture of the proposed dual stream network. Best viewed in color.}
    \label{fig:totalsystem}
\end{figure*}

\section{Review of Brain Glioma Segmentation methods}
Early research on brain tumor segmentation was based on traditional machine learning algorithms such as clustering \cite{shin2012hybrid}, random decision forests \cite{festa2013automatic}, Bayesian models \cite{corso2008efficient} and graph-cuts \cite{wels2008discriminative}. Shin \cite{shin2012hybrid} used sparse coding for generating edema features and K-means for clustering the tumor voxels. However, how to optimise the size of the sparse coding dictionary is still an intractable problem. Pereira et al. \cite{pereira2016brain} proposed to classify each voxel’s label by using random decision forests, which relied on hand-crafted features and complicated post-processing. Corso et al. \cite{corso2008efficient} used a Bayesian formulation for incorporating soft model assignments into the affinities calculation. This method brought the weighted aggregation of multi-scale features but ignored the relationship between different scales. Wels et al. \cite{wels2008discriminative} proposed a graph-cut based method to learn optimal graph representation for tumor segmentation, leading to superior performance. However, this method required a long inference time for dense segmentation tasks, as the number of vertices in its graph is proportional to the number of the voxels.

Promising achievements have been made on multi-modal MRI brain tumor segmentation using deep convolutional neural networks. Zikic et al. \cite{zikic2014segmentation} is one of the pioneers applying DCNNs onto brain tumor segmentation. Havaei et al \cite{havaei2017brain} further improved DCNNs with different sizes of convolutional kernels in order to capture local and global information. Zhao et al. \cite{zhao2018deep} proposed a modified FCN connected with conditional random fields for refining brain tumor segmentation using three MRI modalities. Dong et al. \cite{dong2017automatic} proposed a modified U-Net for brain tumor segmentation. These previous works used 2D convolutional kernels on 2D MRI slices made from original 3D volumetric MRI data. Methods using 2D slices do decrease the number of the used parameters and require less memory due to dimensionality reduction. However, this pre-processing procedure also leads to the spatial context missing. To minimise the information loss and capture evidence from adjacent slices, Lyksborg et al. \cite{lyksborg2015ensemble} ensembled three 2D CNNs on three orthogonal 2D patches. 

To fully make use of 3D contextual information, recent works applied 3D convolutional kernels on original volume data. Kamnitsas et al. \cite{kamnitsas2017efficient} proposed two pathway 3D CNN followed with dense CRF called DeepMedic for brain tumor segmentation. Authors of \cite{kamnitsas2017efficient} further extended the work by using model ensembling \cite{kamnitsas2017ensembles}. The proposed system EMMA ensembled models from FCN, U-Net and DeepMedic for processing 3D patches. To avoid over-fitting problems in 3D voxel-level segmentation on limited training datasets, Myronenko \cite{myronenko20183d} proposed a 3D CNN with an additional variational autoencoder to regularise the decoder by reconstructing the input image. The architecture built in \cite{myronenko20183d} is further developed in various recent works. Su et al. \cite{su2020multimodal} extends the architecture built in \cite{myronenko20183d} into two sub-networks to fuse the information learned from different modalities. Jiang et al. \cite{jiang2019two} proposed two-stage networks where each stage adopts a similar network in \cite{myronenko20183d}. The first stage network generates a coarse result and the second stage network refines the segmentation result. The final result in \cite{jiang2019two} reaches state-of-the-art by ensemble 12 model instances, which requires huge computational resources. Other works also try to fuse information brought by images in a different modality. Wang et al. \cite{wang2020modality} paired data from a different modality and designed the consistency loss to learn the relationship between features in different modalities. Dorent et al. \cite{dorent2019hetero} utilize the network in \cite{myronenko20183d} for multi-task learning, e.g. joint modality completion and  segmentation together. However, the aforementioned approaches only consider the relationship that lies within the modality and ignores the spatial relationship among features, which is more important to achieve accurate segmentation. 

Recent research works began to focus on using graph neural networks for object semantic segmentation.  Qi et al. \cite{qi20173d} and Landrieu et al. \cite{landrieu2018large} construct graph networks for point cloud semantic segmentation based on energy minimization. However, the data used in these approaches are point clouds. Point cloud contains the point nodes, which can be directly used for building graphs. Lu et al. \cite{lu2019graph} construct a graph-FCN for object semantic segmentation. However, this approach builds the graph by extracting nodes using convolutional kernel and ignores the information regulation between normal convolution and graph convolution. The same issue lies in \cite{liu2020scg}  and \cite{zhang2019dual} as the proposed model cannot adaptively make a preference between features from normal convolution and features from graph convolution.

Medical image datasets (e.g. BraTS) usually have an imbalance and inter-class interference problems. To address these issues whilst maintaining segmentation performance, Chen et al. \cite{chen2018focus} and Wang et al. \cite{wang2017automatic} both applied cascaded network structures for segmenting brain tumors, where the input of the inner region segmentation network is the output of the outer region segmentation network. However, these cascaded structures force the networks to crop data in the cascading stage and hence cause information loss. The summary of MRI based brain tumor segmentation is shown in Table \ref{table:lr}.

\begin{sidewaystable}
\centering
\caption{Summary of existing brain tumor segmentation methods.}
\begin{adjustbox}{width=1\textwidth}
\begin{tabular}{lllll}
\hline
Authors         & Base Model                                                                            & Data Format           & Highlights                                                                                                                                                                                                          & Limitations                                                                                                                                                                        \\ \hline
Wels et al. \cite{wels2008discriminative}     & Graph Cut                                                                             & 3D Volumes            & \begin{tabular}[c]{@{}l@{}}(1) A statistical formulation of \\ brain tumor segmentation.\end{tabular}                                                                                                               & \begin{tabular}[c]{@{}l@{}}(1) The size of graph vertices can be large.\\ (2) Using hand-crafted features.\end{tabular}                                                           \\ \hline
Corso et al. \cite{corso2008efficient}   & \begin{tabular}[c]{@{}l@{}}Bayesian Classifier + \\ the weighted Aggragation\end{tabular} & 2D Single-view slice  & \begin{tabular}[c]{@{}l@{}}(1) Explicitly learned the hierarchical \\ information of tumor tissue structures.\end{tabular}                                                                                            & \begin{tabular}[c]{@{}l@{}}(1) The final segmentation performace \\ heavily relys on the result of weighted \\ aggragation.\end{tabular}                                           \\ \hline
Shin \cite{shin2012hybrid}            & \begin{tabular}[c]{@{}l@{}}Spase coding + \\ K-means Clustering\end{tabular}          & 2D Single-view Slice  & (1) Fast and easy implementation.                                                                                                                                                                                   & \begin{tabular}[c]{@{}l@{}}(1) Clustering performance relys on \\ the quality of sparse coding features.\end{tabular}                                                                  \\ \hline
Festa et al. \cite{festa2013automatic}  & Random Decision Forest                                                                & 3D Volumes            & \begin{tabular}[c]{@{}l@{}}(1) Good interpretation based on\\  the classifier decisions.\end{tabular}                                                                                                               & (1) Hand-crafted features.                                                                                                                                                         \\ \hline
Zikic et al. \cite{zikic2014segmentation}   & 2D CNN                                                                                & 2D Patches            & (1) Computational efficient.                                                                                                                                                                                        & \begin{tabular}[c]{@{}l@{}}(1) Cannot directly learn information \\ from 3D space.\end{tabular}                                                                                    \\ \hline
Pereira et al. \cite{pereira2016brain}  & 2D CNN                                                                                & 2D Single-view slice  & \begin{tabular}[c]{@{}l@{}}(1) Stack small size kernels to \\ capture larger receptive field.\end{tabular}                                                                                                          & \begin{tabular}[c]{@{}l@{}}(1) Patch-classification based segmentation.\\ (2) Requires complicated post-processing.\end{tabular}                                                   \\ \hline
Havaei et al. \cite{havaei2017brain}   & 2D CNN                                                                                & 2D Patches            & \begin{tabular}[c]{@{}l@{}}(1) Replace final fully connected layer \\ with convolution layer, which leads\\  to a significant speed up.\\ (2) Studied the effectiveness \\ of different connections.\end{tabular}   & (1) Patch-classification based segmentation.                                                                                                                                       \\ \hline
Dong et al. \cite{dong2017automatic}     & 2D FCNN                                                                               & 2D Single-view Slice  & \begin{tabular}[c]{@{}l@{}}(1) Introduced a novel soft \\ dice loss function.\end{tabular}                                                                                                                          & \begin{tabular}[c]{@{}l@{}}(1) UNet based FCNN, which \\ used simple concatenation \\ for feature fusion.\end{tabular}                                                             \\ \hline
Kamnitsas et al. \cite{kamnitsas2017efficient} & 3D FCNN + CRF                                                                         & 3D Patches            & \begin{tabular}[c]{@{}l@{}}(1) 3D kernel to learn information \\ from volumetric space.\end{tabular}                                                                                                                & \begin{tabular}[c]{@{}l@{}}(1) Patch-classification based segmentation\\ (2) Cascaded connection between\\  FCNN and CRF.\end{tabular}                                             \\ \hline
Kamnitsas et al. \cite{kamnitsas2017ensembles} & 3D CNN Ensembling                                                                     & 3D Volumes            & \begin{tabular}[c]{@{}l@{}}(1) High accuracy benefited from \\ multiple segmentation models.\end{tabular}                                                                                                           & (1) Computation resource exhausted.                                                                                                                                                \\ \hline
Wang et al. \cite{wang2017automatic}     & 2D Cascaded CNN                                                                       & 2D Single-view Slices & \begin{tabular}[c]{@{}l@{}}(1) Explicitly model the hierachical \\ information of tumor tissue structures.\end{tabular}                                                                                              & \begin{tabular}[c]{@{}l@{}}(1) Information and receptive field\\  loss during crop operation.\\ (2) Over-parameterization by \\ introducing complicated sub-networks.\end{tabular} \\ \hline
Zhao et al. \cite{zhao20173d}     & 2D FCNN + CRF                                                                         & 2D Patches            & \begin{tabular}[c]{@{}l@{}}(1) Fully convolutional network to \\ generate segmentation map directly.\end{tabular}                                                                                                   & \begin{tabular}[c]{@{}l@{}}(1) Cascaded connection between \\ FCNN and CRF.\end{tabular}                                                                                           \\ \hline
Chen et al. \cite{chen2018focus}    & 2D Cascaded CNN                                                                       & 2D Single-view Slices & \begin{tabular}[c]{@{}l@{}}(1) Explicitly model the hierachical \\ information of tumor tissue structures.\end{tabular}                                                                                              & \begin{tabular}[c]{@{}l@{}}(1) Information and receptive field\\  loss during crop operation.\\ (2) Over-parameterization by\\  introducing complicated sub-networks.\end{tabular} \\ \hline
Myronenko \cite{myronenko20183d}      & 3D FCNN                                                                               & 3D Volumes            & \begin{tabular}[c]{@{}l@{}}(1) Additional autoencoder branch \\ for encoder backbone regularization.\end{tabular}                                                                                                   & \begin{tabular}[c]{@{}l@{}}(1) Cannot explicitly learn the hierachical \\ information of tumor tissue structures.\end{tabular}                                                      \\ \hline
Ours            & 3D FCNN                                                                               & 3D Volumes            & \begin{tabular}[c]{@{}l@{}}(1) Effectively modeling the hierachical \\ information of tumor tissue structures \\ by learning feature \\ interaction information.\\ (2) Built-in CRF for feature fusion.\end{tabular} & \begin{tabular}[c]{@{}l@{}}(1) No specific strategy to handle \\ from the data imbalance issue.\end{tabular}                                                                       \\ \hline
\end{tabular}
\end{adjustbox}
\label{table:lr}
\end{sidewaystable}

\section{Proposed Method}
\label{Proposed Method}
\begin{figure*}[!htb]
    \centering
    \includegraphics[width=\textwidth]{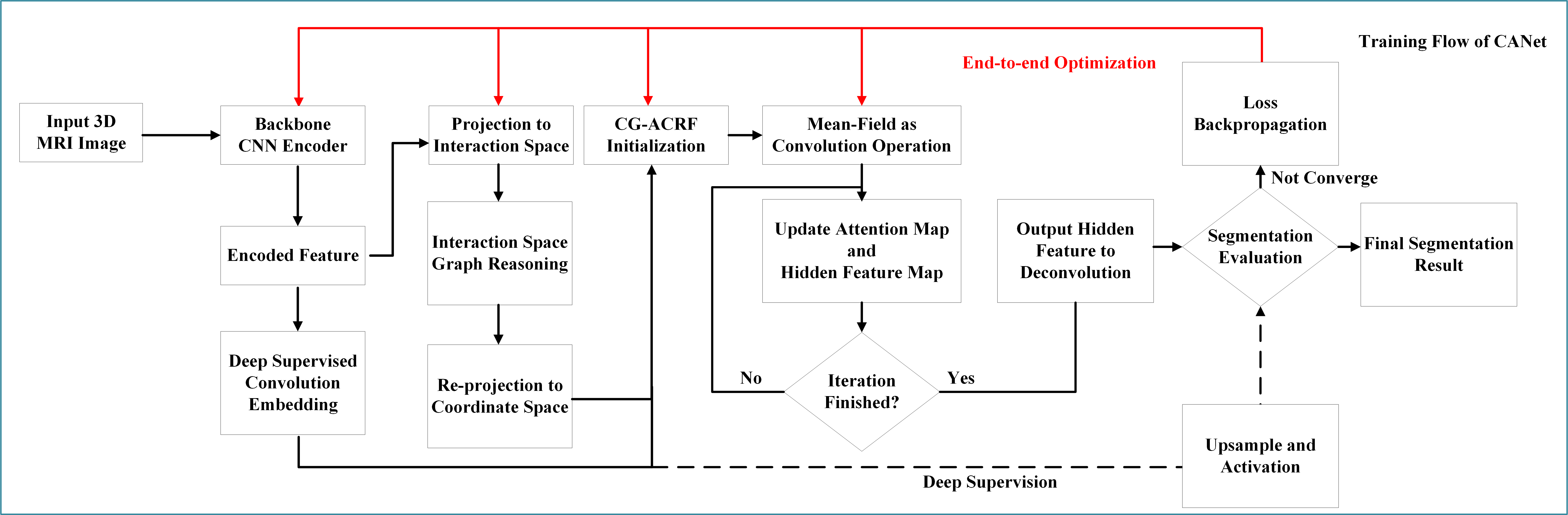}
    \caption{Training flow of the proposed CANet. Best viewed in colors.}
    \label{fig:trainingflow}
\end{figure*}

In this section, we describe our proposed CANet for dense voxel-wise segmentation of 3D MRI brain tumor images. We first describe the proposed HCA-FE with the feature interaction graph and convolutional space contexts in detail. Then we introduce the proposed novel fusion module, CG-ACRF, which deals with the features generated from two branches in HCA-FE and learns to output an optimal feature map. Finally, the formulation of mean-field approximation inference in CG-ACRF as convolutional operations is described, enabling the network to achieve end-to-end training. An illustration of the proposed segmentation framework is shown in Fig. \ref{fig:totalsystem}. Fig. \ref{fig:trainingflow} summarises the training steps of our CANet.

Different from previous works, our proposed HCA-FE can capture long-range contextual information in the feature space by learning the feature interaction, which has not been fully studied in the past. Both streams take the feature map $\mathcal{X} \in \mathbb{R}^{N \times C}$ derived from the shared encoder backbone as input, where $N = H\times W \times D$ is the total number of the voxels in a 3D MRI image. $H$, $W$, and $D$ represent the height, width and depth of the 3D MRI image respectively. $C$ is the number of the feature dimension. The graph stream generates representations in the feature interaction graph space $\mathcal{X}_{\mathcal{G}} \in \mathbb{R}^{N \times C}$ and the convolution stream generates a coordinate space representation $\mathcal{X}_{\mathcal{C}}\in \mathbb{R}^{N \times C}$.

The main concept behind the design of CG-ACRF is to estimate a segmentation map $\mathbf{T} \in \mathcal{T}$ associated with an MRI image $\mathbf{I} \in \mathcal{I}$ by exploiting the relationship between the final representation $\mathcal{X}_{\mathcal{F}} \in \mathbb{R}^{N \times C}$ and the intermediate feature representation $\mathcal{X}$ with auxiliary long-range contextual information $\mathcal{X}_{\mathcal{G}}$, generated from the interaction space with its convolution features $\mathcal{X}_{\mathcal{C}}$. Different from the simple concatenation $\mathcal{X}_{\mathcal{F}} = \textit{concat}(\mathcal{X}, \mathcal{X}_{\mathcal{G}}, \mathcal{X}_{\mathcal{C}})$ or element-wise summation $\mathcal{X}_{\mathcal{F}} = \mathcal{X} + \mathcal{X}_{\mathcal{G}} + \mathcal{X}_{\mathcal{C}}$, we aim to learn a set of latent feature representations $\mathcal{X}_{\mathcal{F}}^\mathcal{H} \in \mathbb{R}^{N \times C}$ through a new CRF. Due to the context information from $\mathcal{X}_{\mathcal{C}}$ and $\mathcal{X}_{\mathcal{G}}$ may contribute differently during the learning $\mathcal{X}_{\mathcal{F}}^\mathcal{H}$, we adopt the idea of an attention mechanism and generalise it into an gate node in CRF. The gate node can regulate the information flow and automatically discover the relevance between different contexts and latent features.
\subsection{Hybrid Context Aware Feature Extractor}
\label{HCG-FA}
\subsubsection{Graph Context Branch}
\textbf{Projection with Adaptive Sampling} We first use the collected feature map to create a feature interaction space by constructing an interaction graph $\mathcal{G} = \{\mathcal{V}, \mathcal{E}, A\}$, where $\mathcal{V}$ represents the set of nodes in the interaction graph, $\mathcal{E}$ represents the edges between the interaction nodes and $A$ represents the adjacency matrix. Given a learned high dimensional feature $X = \{x_{i}\}_{i=1}^{N} \in \mathbb{R}^{N \times C}$  with each $x_{i} \in \mathbb{R}^{1 \times C}$ from the back-bone network, we first project the original feature onto the feature interaction space, generating a projected feature $X_{proj} = \{x_{i}^{proj}\}_{i=1}^{N} \in \mathbb{R}^{K \times C'}$. $K$ is the number of the interaction nodes in the interaction graph and $C'$ is the interaction space dimension. A naive method for getting each element $x_{i}^{proj} \in X_{proj}, i=\{1,...,K\}$ is using the linear combination of its neighbor elements:
\begin{equation}
x_{i}^{proj} = \sum_{\forall j \in \mathcal{N}_{i}} w_{ij}x_{j} A[i,j]
\end{equation}
where $\mathcal{N}_{i}$ denotes the neighbors of pixel $i$. The naive approach normally employs a fully-connected graph with redundant connections and parameters between the interaction nodes, which is very difficult to optimise. More importantly, the linear combination method lacks an ability to perform adaptive sampling because different images contain different contextual information of brain tumors (e.g. location, size and shape). We deal with this issue by performing an adaptive sampling strategy:
\begin{equation}
    \begin{split}
        \triangle \textit{j} &= W_{i,j}x_{i} + b_{i,j}\\
        x_{i}^{proj} &= \sum_{\forall j \in \mathcal{N}_{i}} w_{ij}\rho (x_{j} | \mathcal{V}, j, \triangle j) A[i,j]
    \end{split}
\end{equation}

where $W_{i,j} \in \mathcal{R}^{3 \times (K \times C)}$ and $b_{i,j} \in \mathcal{R}^{3 \times 1}$ are the shift distances which are learned individually for each source feature $x_{i}$ through stochastic gradient decent. $\rho(\dot)$ is the trilinear interpolation sampler which can sample a shifted interaction node $x_{j}^{d}$ around feature node $x_{j}$, given the learned deformation $\triangle j$ and the total set of interaction graph nodes $\mathcal{V}$.

\textbf{Interaction Graph Reasoning} After having projected the input features into the interaction graph $\mathcal{G}$ with $K$ interaction nodes $\mathcal{V} = \{v_{1},...,v_{k}\}$ and edges $\mathcal{E}$, we follow the definition of the graph convolution network \cite{DBLP:conf/iclr/KipfW17}. In particular, we define $A_{\mathcal{G}}$ as the adjacency matrix on $K \times K$ nodes and $W_{\mathcal{G}} \in \mathbb{R}^{D \times D}$ as the weight matrix, and the formulation of the graph convolution operation is formulated as follows:
\begin{equation}
\begin{split}
    X_{\mathcal{GC}} &= \sigma(A X_{proj} W_{\mathcal{G}})\\
    &= \sigma((I - \hat{A}) X_{proj} W_{\mathcal{G}})
\end{split}
\end{equation}
where $\sigma()$ is sigmoid activation function. We first apply Laplacian smoothing and update the adjacency matrix to $(I - \hat{A})$ so as to propagate the node feature over the entire graph.
In practice, we implement $\hat{A}$ and $W_{\mathcal{G}}$ using a $1 \times 1$ convolution layer. We also achieve the implementation of $I$ as a residual connection which can maximise the gradient flow with a faster convergence speed.

\textbf{Re-Projection} Once the feature propagation has been finished, we re-project the features back to the original coordinate space with output $\mathcal{X}_{\mathcal{G}} \in \mathbb{R}^{N \times D}$. Similar to the projection step, we use trilinear interpolation here to calculate each elements $x_{\mathcal{G}}^{i} \in \mathcal{X}_{\mathcal{G}}, i \in \{1,...,N\}$ after having transformed the feature from the interaction space to the coordinate space. As a result, we have the feature $\mathcal{X}_{\mathcal{G}}$ with feature dimension $D$ at all $N$ grid coordinates.

\begin{figure*}[t]
    \centering
    \includegraphics[width=\textwidth]{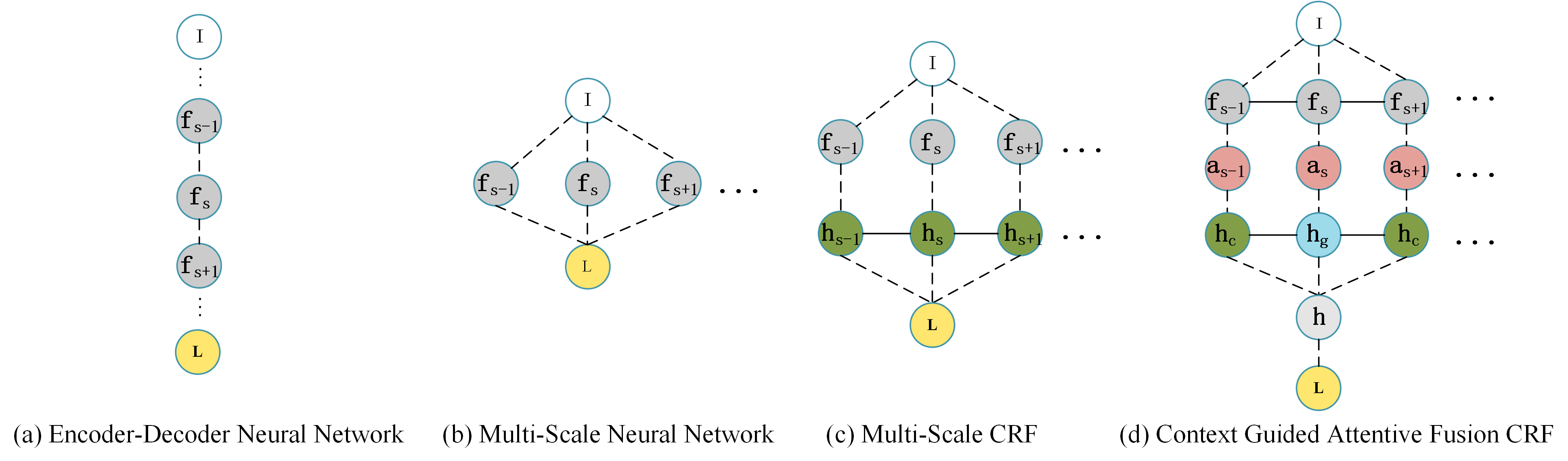}
    \caption{{{A graph model illustration of previous fusion schemes: (a) basic encoder-decoder neural network, (b) multi-scale neural network, (c) multi-scale CRF, and  our proposed (d) context guided attentive fusion CRF. $I$ denotes the input 3D MRI image. $f_s$ denotes the feature map at scale $s$. $a_s$ indicates the attention map generated from the corresponding feature $f$ at scale $s$. $h_c$ and $h_g$ represent the hidden feature generated from convolutional features and graph convolutional features respectively. $L$ means the final segmentation labeling output. Best viewed in color.}}}
    \label{fig:crfcomp}
\end{figure*}

\subsubsection{Convolution Context Branch}
The convolution context branch is composed of a contracting path (encoder) and an expansive path (decoder) with skip connections between these two paths. The contracting path reduces the spatial dimensionality of the pooling layer in a pyramidal scale whilst the expansive path recovers the spatial dimensionality and the details of the object with the corresponding pyramid scale. One of the advantages of using this architecture is that it fully utilises the features with different scales of contextual information, where large scale features can be used to localise objects and small scale but high dimensionality features can provide more detailed and accurate information for classification.

However, 3D volumetric images require more parameters to learn during feature extraction. It is often observed that training such 3D model often fails for various reasons such as over-fitting and gradient vanishing or exploding. Besides, simple or complicated augmentation technologies used to extend the training dataset may result in a slow convergence speed. To address the issues mentioned above, we develop a deep supervised mechanism that inherits the advantages of the convolution context branch. The proposed deep supervision mechanism thus reinforces the gradient flow and improves the discriminative capability during the training procedure.

Specifically, we use additional upsampling layers to reshape the features created at the deep supervised layer to be of the resolution of the final input. For each transformed layer, we apply the softmax function to obtain additional dense segmentation maps. For these additional segmentation results, we calculate the segmentation errors with regards to the ground-truth segmentation maps. The auxiliary losses are integrated with the loss from the output layer of the whole network and we further back-propagate the gradient for parameter updating during each iteration in the training stage.

We denote the set of the parameters in the deep supervised layers as $W_{S} = \{w^{i}\}_{i=1}^{S}$ and $w^{s}$ as the parameters of the upsampling layer correspond to layer $s$. The auxiliary loss for a deep supervision layer $s$ is formulated using cross-entropy:
\begin{equation}
    \mathcal{L}_{s}(\mathcal{X};W_{S}) = \sum_{i=1}^{S}\sum_{j=1}^{N}-\log  \mathbbm{1}(p(y_{j}|x_{j}^{s};w^{s}))
\end{equation}
where $\mathbbm{1}$ is the indicator function which is 1 if the segmentation result is correct, otherwise 0.  $\mathcal{Y} = \{y_{i}\}_{i=1}^{N}$ is the ground-truth of voxel $i$ and $\mathcal{X}_{S} = \{x^{s}_{i}\}_{i=1,s=1}^{N,S}$ is the predicted segmentation label of voxel $i$ generated from the upsampling layer $s$. Finally, the deep supervision loss $\mathcal{L}_{s}$ can be integrated with the loss $\mathcal{L}_{T}$ from the final output layer. The parameters of the deep supervised layers $W_{S}$ can be updated with the rest parameters $W$ from the whole framework simultaneously using back-propagation:
\begin{equation}
\begin{split}
      \mathcal{L} = \mathcal{L}_{T}(\mathcal{Y}| & \mathcal{X};W, W_{S}) + \sum_{s=1}^{S}\delta_{s}\mathcal{L}_{s}(\mathcal{X};w^{s})\\
      &+ \lambda(||W||^{2} + \sum_{s=1}^{S}(||w^{s}||^{2}))
\end{split}
\label{cnnstreamtogetherloss}
\end{equation}
where $\delta_{s}$ represents the weight factor for the supervision loss of each upsampling layer. As the training procedure continues to approach to the optimal parameter sets, $\delta_{s}$ reduces gradually. The final operation of Eq. (\ref{cnnstreamtogetherloss}) is the $L$2-regularisation of the total trainable weights with the weight factor $\lambda$.


\subsection{Context Guided Attentive Conditional Random Field}
We further propose a novel context guided attentive CRF module to perform feature fusion, motivated from two perspectives. A graph model of our proposed CG-ACRF is illustrated in Fig. \ref{fig:crfcomp}. There are two reasons to use CG-ACRF for feature fusion. Firstly, assigning segmentation labels by maximising probabilities may result in blurry boundaries due to the neighboring voxels sharing similar spatial contexts. Secondly, previous works fuse information from different sources (e.g. multi-scale or multi-stage) by using simple channel-wise concatenation or element-wise summation mechanism. However, these mechanisms do not take into account the heterogeneity between different feature maps (e.g. shallow layers tend to focus on low-level visual features while deep layers tend to attend abstract features). Simplifying the relationship between different source feature maps (e.g. feature maps of large kernels tend to represent the object outline while feature maps of small kernels tend to encode the details of the object structure) results in information loss. Different from previous related works and using the inference ability of the probabilistic graphical model, we employ the conditional random field model to learn optimised latent fusion features for final segmentation. As information from different contexts may contribute to the final results at different degrees, we integrate the attention gates of the CRF to regulate how much information should flow between features generated from different contexts. We further show the convolution formulation of CG-ACRF mean-field approximation inference, which allows our attentive CRF fusion module to be integrated into neural networks as a layer and trained in an end-to-end fashion. Compared with previous architectures such as an encoder-decoder neural network (Fig. \ref{fig:crfcomp} (a)) and multi-scale neural network (Fig. \ref{fig:crfcomp} (b)), our proposed CG-ACRF (Fig. \ref{fig:crfcomp} (d)) has a strong inference ability and can jointly learn the hidden representation of features encoded by the neural network backbone, improving the generalisation ability of the segmentation model. Compared with previous architectures such as multi-scale CRF (Fig. \ref{fig:crfcomp} (c)), our proposed CG-ACRF model first uses an attention gate by directly modeling the cost energy in the network (Eq. (\ref{EnergyDefinition})). The attention gate thus regulates the information flow from the features encoded by the backbone neural network to the latent representations by minimising the total energy cost. Moreover, our proposed CG-ACRF learns to project the features into two spaces, i.e. a convolutional space and a feature interaction graph. Hidden representations from different spaces can further boost feature fusion performance. We evaluate the effectiveness of each component in the experiment section.

\subsubsection{Definition}
Given the feature map $\mathcal{X}_{\mathcal{C}} = \{x_{\mathcal{C}}^{i}\}_{i=1}^{N}$ from the convolution context branch and the feature map $\mathcal{X}_{\mathcal{G}} = \{x_{\mathcal{G}}^{i}\}_{i=1}^{N}$ from the interaction graph branch, our goal is to estimate the fusion representation $H_{\mathcal{G}} = \{h_{\mathcal{G}}^{i}\}_{i=1}^{N}, H_{\mathcal{C}} = \{h_{\mathcal{C}}^{i}\}_{i=1}^{N}$ and the attention variable $A = \{a_{\mathcal{G}\mathcal{C}}^{i}\}_{i=1}^{N}$. We formalise the problem by designing a Context Guided Attentive Conditional Random Field with a Gibbs distribution as follows:
\begin{equation}
    P(H,A|I,\Theta) = \frac{1}{Z(I, \Theta)}exp\{-E(H,A,I,\Theta)\}
\end{equation}
where $E(H,A,I,\Theta)$ is the associated energy, and
\begin{equation}
     E(H,A,I,\Theta) = \Phi_{\mathcal{G}}(H_{\mathcal{G}}, X_{\mathcal{G}}) + \Phi_{\mathcal{C}}(H_{\mathcal{C}}, X_{\mathcal{C}}) + \Psi_{\mathcal{G}\mathcal{C}}(H_{\mathcal{G}}, H_{\mathcal{C}}, A)
\label{EnergyDefinition}
\end{equation}
where $I$ is the input 3D MRI image and $\Theta$ is the set of parameters. In Eq.(\ref{EnergyDefinition}), $\Phi_{\mathcal{G}}$ is the unary potential between the latent graph representation $h_{\mathcal{G}}^{i}$ and the graph features $x_{\mathcal{G}}^i$. $\Phi_{\mathcal{C}}$ is the unary potential related to latent convolution representation $h_{\mathcal{C}}^i$ and convolution feature $x_{\mathcal{C}}^i$. In order to enable the estimated latent representation $h^i$ to be close to the observation $x^i$, we use the Gaussian function created in previous works \cite{krahenbuhl2011efficient}:

\begin{equation}
    \Phi(H, X) = \sum_{i}\phi(h^i, x^i) = -\sum_{i=1}^{N}\frac{1}{2}||h^{i} - x^{i}||^{2}.
\end{equation}

The final term shown in Eq. (\ref{EnergyDefinition}) is the attention guided pairwise potential between the latent convolution representation $h_{\mathcal{C}}^i$ and the latent graph representation $h_{\mathcal{G}}^i$. The attention term $a_{\mathcal{GC}}^{i}$ controls the information flow between the two latent representations where the graph representation may or may not contribute to the estimated convolution representation. We define:
\begin{equation}
\begin{split}
    \Psi_{\mathcal{G}\mathcal{C}}(H_{\mathcal{G}}, H_{\mathcal{C}}, & A) = \sum_{i=1}^{N}\sum_{j \in \mathcal{N}_{i}}\psi(a_{\mathcal{GC}}^{j}, h_{\mathcal{C}}^{i}, h_{\mathcal{G}}^{j})\\
    & = \sum_{i=1}^{N}\sum_{j \in \mathcal{N}_{i}}a_{\mathcal{GC}}^{j} h_{\mathcal{C}}^{i} \Upsilon_{\mathcal{GC}}^{i,j} h_{\mathcal{G}}^{j}
\end{split}
\end{equation}
where $\Upsilon_{\mathcal{GC}}^{i,j} \in \mathbb{R}^{D_{\mathcal{G}} \times D_{\mathcal{C}}}$ and $D_{\mathcal{G}}, D_{\mathcal{C}}$ represent the dimensionality of the features $X_{\mathcal{G}}$ and $X_{\mathcal{C}}$ respectively.

\subsubsection{Inference}
\label{MFCRF}
By learning latent feature representations to minimise the total segmentation energy, the system can produce an appropriate segmentation map, \textit{e.g.} the maximum a posterior $P(H,A|I,\Theta)$. However, the optimisation of $P(H, A|I,\Theta)$ is intractable due to the computational complexity in normalising constant $Z(I, \Theta)$, which is exponentially proportional to the cardinality of $h \in H$ and $a \in A_{\mathcal{GC}}$. Therefore, in order to derive the maximum a posterior in an efficient way, we adopt mean-field approximation to approximate a complex posterior probability distribution. We have:
\begin{equation}
    P(H,A | I,\Theta) \approx Q(H,A) = \prod_{i=1}^{N} q_{i}(h_{\mathcal{G}}^i)q_{i}(h_{\mathcal{C}}^i)q_{i}(a_{\mathcal{G}\mathcal{C}}^i)
\label{mfdef}
\end{equation}

Here we use the product of independent marginal distributions $q(h_{\mathcal{G}}^i)$, $q(h_{\mathcal{C}}^i)$ and $q(a_{\mathcal{G}\mathcal{C}}^i)$ to approximate the complex distribution $P(H,A,I,\Theta)$. To achieve a satisfactory approximation result, we minimise the Kullback-Leibler (KL) divergence $D_{KL}(Q||P)$ between the two distributions $Q$ and $P$. By replacing the definition of the energy $E(H,A,I,\Theta)$, we formulate the KL divergence in Eq. (\ref{mfdef}) as follows:
\begin{equation}
\begin{split}
    D_{KL}(Q||P) &= \sum_{h}Q(h)\ln (\frac{Q(h)}{P(h)})\\
    &=\sum_{h}Q(h)E(h) + \sum_{h}Q(h)\ln Q(h) + \ln Z
\end{split}
\label{kl1}
\end{equation}
From Eq.(\ref{kl1}), we can minimise the KL divergence by directly minimising the free energy $FE(Q) = \sum_{h}Q(h)E(h) + \sum_{h}Q(h)\ln (Q(h))$. In $FE(Q)$, the first item represents the cross-entropy between two distributions $Q$ and $E$ and the second item represents the entropy of distribution $Q$. We can further expand the expression of $FE(Q)$ by replacing $Q$ and $E$ with Eqs. (\ref{mfdef}) and (\ref{EnergyDefinition}) respectively:

\begin{equation}
\begin{split}
    FE(Q) &= \sum_{i=1}^{N}q_{i}(h_{\mathcal{G}}^i)q_{i}(h_{\mathcal{C}}^i)q_{i}(a_{\mathcal{G}\mathcal{C}}^i)(\Phi_{\mathcal{G}}+\Phi_{\mathcal{C}}+\Psi_{\mathcal{GC}})\\
    &+ \sum_{i=1}^{N}q_{i}(h_{\mathcal{G}}^i)q_{i}(h_{\mathcal{C}}^i)q_{i}(a_{\mathcal{G}\mathcal{C}}^i)(\ln (q_{i}(h_{\mathcal{G}}^i)q_{i}(h_{\mathcal{C}}^i)q_{i}(a_{\mathcal{G}\mathcal{C}}^i)))
\end{split}
\label{FEDef}
\end{equation}

Eq. (\ref{FEDef}) shows that the problem of minimising $FE(Q)$ can be transferred to a constrained optimisation problem with multiple variables, which can be formally formulated below:

\begin{equation}
\begin{split}
    & \min_{q_{i}(h_{\mathcal{G}}^i), q_{i}(h_{\mathcal{C}}^i), q_{i}(a_{\mathcal{G}\mathcal{C}}^i)} FE(Q), \forall i  \in N\\
    & \textrm{s.t.} \sum_{l=1}^{L}q_{i}(h_{\mathcal{G}}^i) = 1, \sum_{l=1}^{L}q_{i}(h_{\mathcal{C}}^i) = 1, \int_{0}^{1} q_{i}(a_{\mathcal{GC}}^i)da^{i}_{\mathcal{GC}} = 1
\end{split}
\label{optDef}
\end{equation}
where $l$ represents the index of the segmentation label. We can calculate the first order partial derivative by differentiating $FE(Q)$ w.r.t each variable. For example, we have:

\begin{equation}
\begin{split}
    \frac{\partial FE}{\partial q_{i}(h^{i}_{\mathcal{C}})} = \phi_{\mathcal{C}}(h_{\mathcal{C}}^{i}, x_{\mathcal{C}}^{i}) +& \sum_{j\in \mathcal{N}_{i}}\mathbb{E}_{q_{j}(a_{\mathcal{GC}}^{j})} \{ a_{\mathcal{GC}}^{j} \} \mathbb{E}_{q_{j}(h_{\mathcal{G}}^j)} \psi (h_{\mathcal{C}}^i, h_{\mathcal{G}}^j)\\
    &- \ln q_{i}(h^{i}_{\mathcal{C}}) + \text{const}
\end{split}
\label{partialDer}
\end{equation}

By assigning 0 to the left of Eq. (\ref{partialDer}), we reach:
\begin{equation}
\begin{split}
     q_{i}(h_{\mathcal{C}}^{i}) \propto & \exp \Big\{ \phi_{\mathcal{C}}(h_{\mathcal{C}}^{i}, x_{\mathcal{C}}^{i}) + \\
     & \sum_{j \in N_{i}} \mathbb{E}_{q_{j}(a_{\mathcal{GC}}^{j})} \{ a_{\mathcal{GC}}^{j} \} \mathbb{E}_{q_{j}(h_{\mathcal{G}}^j)} \psi (h_{\mathcal{C}}^i, h_{\mathcal{G}}^j) \Big \}
\end{split}
\label{qhcdef}
\end{equation}

Eq. (\ref{qhcdef}) shows that, once the other two independent variables $q(h_{\mathcal{G}})$ and  $q(a_{\mathcal{GC}})$ are fixed, how $ q(h_{\mathcal{C}})$ is updated during the mean-field approximation inference. Further more, we follow the above procedure and obtain the updating of the remaining two variable as follows:
\begin{equation}
\begin{split}
     q_{i}(h_{\mathcal{G}}^{i}) \propto & \exp \Big\{ \phi_{\mathcal{G}}(h_{\mathcal{G}}^{i}, x_{\mathcal{C}}^{i}) + \\
     & \mathbb{E}_{q_{j}(a_{\mathcal{GC}}^{j})} \{ a_{\mathcal{GC}}^{j} \} \sum_{j \in N_{i}} \mathbb{E}_{q_{j}(h_{\mathcal{C}}^j)} \psi (h_{\mathcal{C}}^i, h_{\mathcal{G}}^j) \Big \}
\end{split}
\label{qhgdef}
\end{equation}
\begin{equation}
    q_{i}(a_{\mathcal{GC}}^{i}) \propto \exp \Big\{a_{\mathcal{GC}}^{i} \mathbb{E}_{q_{i}(h_{\mathcal{C}}^{i})} \{ \sum_{j \in N_{i}} \mathbb{E}_{q_{j}(h_{\mathcal{G}}^{j})} \{ \psi(h_{\mathcal{C}}^{i}, h_{\mathcal{G}}^{j})\} \} \Big \}
\label{qatdef}
\end{equation}
where $\mathbb{E}_{q()}$ represents the expectation with respect to the distribution $q()$. Eqs. (\ref{qhcdef}-\ref{qatdef}) shown above denote the computational procedure of seeking an optimal posterior distributions of $h_{\mathcal{C}}$, $h_{\mathcal{G}}$ and $a_{\mathcal{GC}}$ during the mean-field approximation. Intuitively, Eq. (\ref{qhcdef}) shows that, the latent convolution feature $h_{\mathcal{C}}^{i}$ for voxel $i$ can be  used to describe the observation, referred to feature $x_{\mathcal{C}}^{i}$. Afterwards, we use the re-weighted messages from the latent features of the neighboring voxels to learn the co-occurrent relationship of the pixels. The attention weight between the latent convolution and the graph features for voxel $i$ allows us to re-weight the pairwise potential message from the neighbours of voxel $i$, and then use the attention variable to re-weight the total value of voxel $i$. By denoting $\Bar{a}_{\mathcal{G}\mathcal{C}}^{i} = \mathbb{E}_{q(a_{\mathcal{G}\mathcal{C}}^{i})}\{a_{\mathcal{G}\mathcal{C}}^{i}\}$ and $\Bar{h}^{i} = \mathbb{E}_{q(h^{i})}\{h^{i}\}$, we have the feature update as follows:
\begin{equation}
    \Bar{h}_{\mathcal{G}}^{i} = x_{\mathcal{G}}^{i} + \Bar{a}_{\mathcal{G}\mathcal{C}}^{i} \sum_{j \in N_{i}} \Upsilon_{\mathcal{G}\mathcal{C}}^{i,j} \Bar{h}_{\mathcal{C}}^{j}
\end{equation}

\begin{equation}
    \Bar{h}_{\mathcal{C}}^{i} = x_{\mathcal{C}}^{i} + \sum_{j \in N_{i}} \Bar{a}_{\mathcal{G}\mathcal{C}}^{j}  \Upsilon_{\mathcal{G}\mathcal{C}}^{i,j} \Bar{h}_{\mathcal{G}}^{j}
\end{equation}

$\Bar{a}_{\mathcal{G}\mathcal{C}}^{i}$ is also derived from the probabilistic distribution, \textit{i.e.} its value lies in $[0,1]$. Here, we choose the Sigmoid function to formulate the updates for $\Bar{a}_{\mathcal{G}\mathcal{C}}^{i}$:
\begin{equation}
    \Bar{a}_{\mathcal{G}\mathcal{C}}^{i} = \sigma(- \sum_{j \in N_{i}}a_{\mathcal{GC}}^{j} h_{\mathcal{C}}^{i} \Upsilon_{\mathcal{GC}}^{i,j} h_{\mathcal{G}}^{j})
\end{equation}
where $\sigma(.)$ denotes the Sigmoid activation function.

\subsubsection{Mean Field Inference as Convolution Operation}
\begin{figure}[t]
    \centering
    \includegraphics[width=0.7\textwidth]{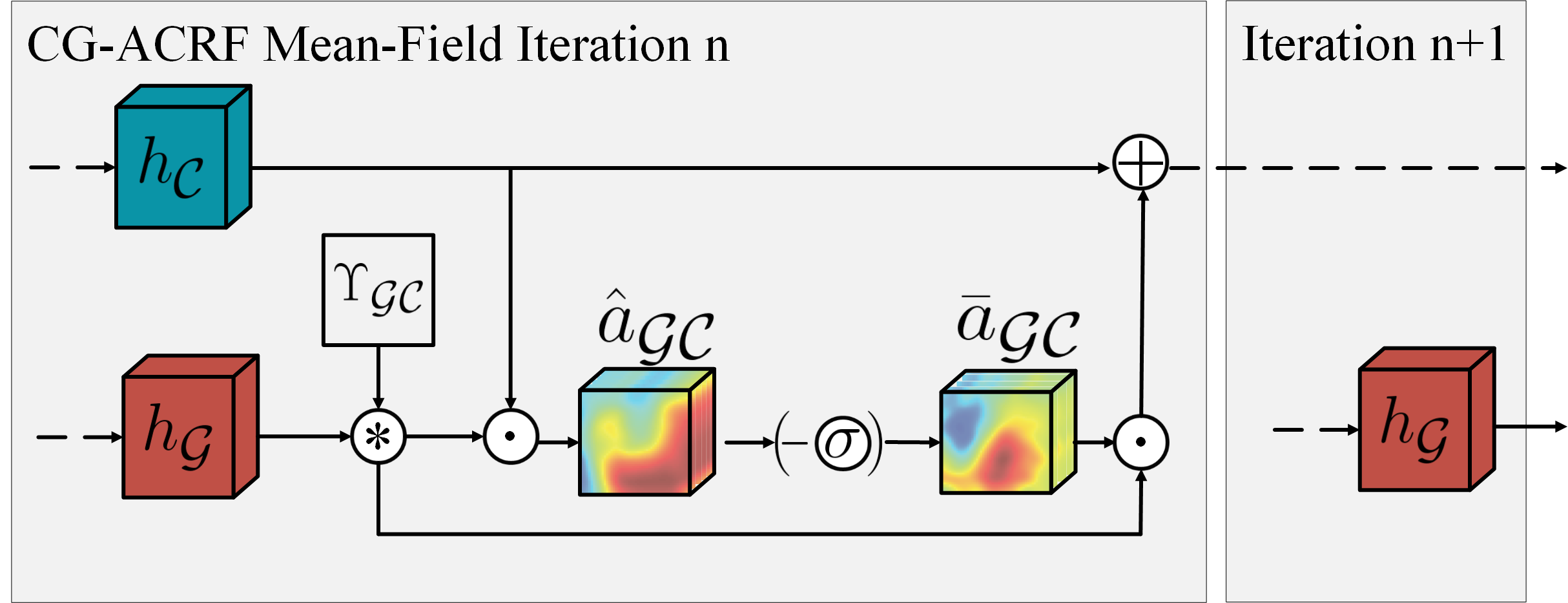}
    \caption{Details of the mean-field updates within CG-ACRF. The circled symbols indicate message-passing operations within the CG-ACRF block. Best viewed in colors.}
    \label{fig:crfmf}
\end{figure}

To achieve joint training and end-to-end optimisation of the proposed CRF with the backbone network, we implement the mean-field approximation of the proposed CRF in neural networks. We aim to perform the updating of the latent feature and attention maps according to the derivation described in Section \ref{MFCRF}. The algorithm for implementing mean-field approximation using convolutional operations is described in Algorithm \ref{Alg:convmf}. A graph illustration of Algorithm \ref{Alg:convmf} is shown in Fig. \ref{fig:crfmf}.

\begin{algorithm}
\caption{Algorithm for Mean-Field Approximation.}
\begin{algorithmic}[1]
\renewcommand{\algorithmicrequire}{\textbf{Input:}}
\renewcommand{\algorithmicensure}{\textbf{Output:}}
\REQUIRE Feature interaction graph output $x_{\mathcal{G}}$ and convolution output $x_{\mathcal{C}}$. Initialize hidden graph feature map $h_{\mathcal{G}}$ with $x_{\mathcal{G}}$. Initialize hidden convolutional feature map $h_{\mathcal{C}}$ with $x_{\mathcal{C}}$
\ENSURE Estimated optimised hidden convolution feature map $h$.
\WHILE {in iteration number}
 \STATE $\hat{a}_{\mathcal{G}\mathcal{C}} \leftarrow h_{\mathcal{C}} \odot (\Upsilon_{\mathcal{G}\mathcal{C}} \ast h_{\mathcal{G}})$;
 \STATE $\Bar{a}_{\mathcal{G}\mathcal{C}} \leftarrow \sigma(-(\hat{a}_{\mathcal{G}\mathcal{C}}))$;
 \STATE $h_{\mathcal{G}} \leftarrow \Upsilon_{\mathcal{G}\mathcal{C}} \ast h_{\mathcal{G}}$;
 \STATE $\Bar{h}_{\mathcal{C}} \leftarrow \Bar{a}_{\mathcal{G}\mathcal{C}} \odot h_{\mathcal{G}}$;
 \STATE $h \leftarrow x \oplus \Bar{h}_{\mathcal{C}}$;
\ENDWHILE
\RETURN Optimised hidden feature map $h$.
\end{algorithmic}
\label{Alg:convmf}
\end{algorithm}

where $\ast, \odot, \oplus$ represent the convolution, element-wise dot product, and element-wise summation respectively. First, the latent feature map is initialised with corresponding observation inputs $x_{\mathcal{G}}$ and $x_{\mathcal{C}}$, while the attention map is initialised from the message passing on the two latent feature maps. Then, we activate and normalise the attention map. The latent convolutional feature map is updated from the message passing on the latent graph feature map. Finally, the updated attention map is used to refine the latent convolutional feature map $h$. We output $h$ with the unary term $x$ by establishing residual connections.

\section{Experiment Setup}
\label{Experimental Setup}
To demonstrate the effectiveness of the proposed CANet for brain tumor segmentation, we conduct experiments on two publicly available datasets: the Multimodal Brain Tumor Segmentation Challenge 2017 (BraTS2017) and the Multimodal Brain Tumor Segmentation Challenge 2018 (BraTS2018). 

\subsection{Datasets and Evaluation Metrics}
\textbf{Datasets.} The \textbf{BraTS2017}\footnote{https://www.med.upenn.edu/sbia/brats2017.html} consists of 285 cases of patients in the training set and 44 cases in the validation set. \textbf{BraTS2018}\footnote{https://www.med.upenn.edu/sbia/brats2018.html} shares the same training set with BraTS2017 and includes 66 cases in the validation set. Each case is composed of four MR sequences, namely native T1-weighted (T1), post-contrast T1-weighted (T1ce), T2-weighted (T2) and Fluid Attenuated Inversion Recovery (FLAIR). Each sequence has a 3D MRI volume of 240$\times$240$\times$155. Ground-truth annotation is only provided in the training set, which contains the background and healthy tissues (label 0), necrotic and non-enhancing tumor (label 1), peritumoral edema (label 2) and GD-enhancing tumor (label 4). We first consider the 5-fold cross-validation on the training set where each fold contains (random division) 228 cases for training and 57 cases for validation. We then evaluate the performance of the proposed method on the validation set. The validation result is generated from the official server of the contest to determine the segmentation accuracy of the proposed methods.	

\textbf{Evaluation Metrics.} Following previous works \cite{wang2017automatic}, \cite{kamnitsas2017efficient}, \cite{bakas2017advancing}, the segmentation accuracy is measured by Dice score, Sensitivity, Specificity and Hausdorff95 distance respectively. In particular,
\begin{itemize}
    \item Dice score: $Dice(P,T) = \frac{|P_1 \cap T_1|}{(|P_1|+|T_1|)/2}$
    \item Sensitivity: $Sens(P,T) = \frac{|P_1 \cap T_1|}{|T_1|}$
    \item Specificity: $Spec(P,T) = \frac{|P_0 \cap T_0|}{|T_0|}$
    \item Hausdorff Distance: $Haus(P,T) = \\
    max\{sup_{p\in P_1}inf_{t\in T_1} d(p,t), sup_{t\in T_1}inf_{p\in P_1} d(t,p)\}$
\end{itemize}
where $P$ represents the model prediction and $T$ represents the ground-truth annotation. $T_1$ and $T_0$ are the subset voxels predicted as positives and negatives for the tumor region. Similar set-ups are made for $P_1$ and $P_0$. Furthermore, the Hausdorff95 distance measures how far the model prediction deviates from the ground-truth annotation. $sup$ represents the supremum and $inf$ represents the infimum. For each metric, three regions namely enhancing tumor (ET, label 1), whole tumor (WT, labels 1, 2 and 4) and the tumor core (TC, labels 1 and 4) are evaluated individually.
\subsection{Data Augmentation and Implementation Details}
\textbf{Data Augmentation} For each sequence in each case, we set all the voxels outside the brain to zero and normalise the intensity of the non-background voxels to be of zero mean and unit variance. During the training, we use randomly cropped images of size 128$\times$128$\times$128.
We further set up a common augmentation strategy for each sequence in each case: (i) randomly rotate an image with the angle between [-20$^{\circ}$, +20$^{\circ}$]; (ii) randomly scale an image with a factor of 1.1; (iii) randomly mirror flip an image across the axial coronal and sagittal planes with the probability of 0.5; (iv) random intensity shift between [-0.1, +0.1]; (v) random elastic deformation with $\sigma = 10$.

\textbf{Implementation Details} We implement the proposed CANet and other benchmark experiments using the PyTorch framework and deploy all the experiments on 2 parallel Nvidia Tesla P100 GPUs for 200 epochs with a batch size of 4. We use the Adam optimizer with an initial learning rate $\alpha_0 = 1\mathrm{e}{-4}$. The learning rate is decreased by a factor of 5 after 100, 125 and 150 epochs. We use a $L$2 regulariser with a weight decay of $1\mathrm{e}{-5}$. We store the weights for each epoch and use the weights that lead to the best dice score for inference.

\section{Experimental Result} 
\label{resultdiscussion}
In this section, we present both quantitative and qualitative experimental results of different evaluations. We first conduct an ablation study of our method to show the effective impact of HCA-FE and CG-ACRF on the segmentation performance. We also perform additional analysis of the encoder backbone and different iteration numbers of approximation for CG-ACRF. Afterward, we compare our approach with several state-of-the-art methods on different datasets. Finally, we present the analysis of failure cases.

\subsection{Ablation Studies}
\begin{figure*}[!ht]
    \centering
    \includegraphics[width=\textwidth]{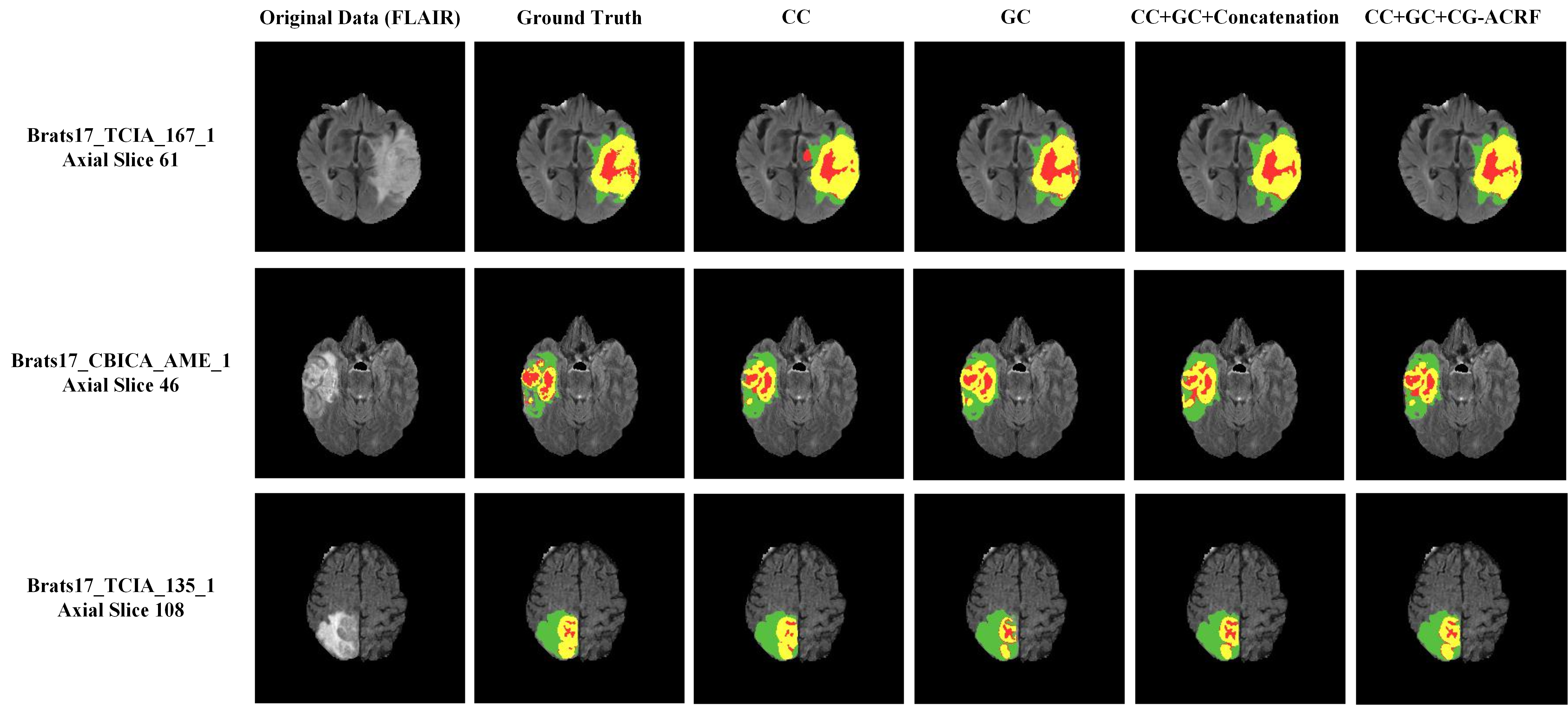}
    \caption{Qualitative comparison of different baseline models and the proposed CANet by cross-validation on BraTS2017 training set. From left to right, each column represents the input FLAIR data, ground truth annotation, segmentation result of CANet with only the convolution branch, segmentation result of CANet with only the graph convolution branch, segmentation output of CANet with HCA-FE and concatenation fusion scheme, segmentation output of CANet with HCA-FE and CG-ACRF fusion module. Best viewed in colors.}
    \label{fig:ablation}
\end{figure*}

We first evaluate the effect of HCA-FE and CG-ACRF. To this end, we apply a 5-fold cross-evaluation on the BraTS2017 training set and report the mean result. Table \ref{table:ablationstudy} shows the quantitative results, while the qualitative results can be found in Fig. \ref{fig:ablation} as an example of the segmentation outputs. We start from two baselines. The first baseline is in the fully convolution format with deep supervision on the backbone convolution encoder (CC). The second baseline only uses graph convolution in the convolution encoder without deep supervision (GC). We then evaluate the proposed whole HCA-FE (CC+GC) without any feature fusion method, \textit{i.e.} concatenating feature maps from CC and GC together. Finally, we evaluate the proposed feature fusion module CG-ACRF, which takes a feature map with different contexts from HCA-FE and outputs the optimal latent feature map for the final segmentation. For the experiments are shown in Table \ref{table:ablationstudy} and Fig. \ref{fig:ablation}, we use the encoder of UNet as the backbone network with 5 iterations in CG-ACRF. The experiments described later include the analysis of different backbones and iteration numbers.

\begin{table*}[!ht]
\centering
\caption{Quantitative results of the CANet components by five fold cross-validation for the BraTS2017 training set (dice, sensitivity and specificity). All the methods are based on CANet with UNet as the backbone. The best result is shown in bold text and the runner-up result is underlined.}
\begin{adjustbox}{width=1\textwidth}
\begin{tabular}{|c|c|c|c|c|c|c|c|c|c|c|c|c|}
\hline
              & \multicolumn{3}{c|}{DICE}                              & \multicolumn{3}{c|}{Sensitivity}                       & \multicolumn{3}{c|}{Specificity}                       & \multicolumn{3}{c|}{Hausdorff95}                       \\ \hline
Backbone+         & ET               & WT               & TC               & ET               & WT               & TC               & ET               & WT               & TC               & ET               & WT               & TC               \\ \hline
CC            & \textbf{0.68628} & 0.87467          & 0.82068          & 0.85684          & {\underline{0.92495}}    & 0.86324          & {\underline{0.99708}}    & {\underline{0.99094}}    & 0.99562          & \textbf{6.79149} & 6.88633          & 7.93923          \\ \hline
GC            & 0.6373           & {\underline{0.89365}}    & {\underline{0.82246}}    & \textbf{0.97704} & \textbf{0.96964} & \textbf{0.94428} & 0.98723          & 0.98742          & \textbf{0.99665} & 9.89949          & {\underline{6.40312}}    & {\underline{5.81216}}    \\ \hline
CC+GC+Concatenation         & 0.68194          & 0.86073          & 0.80306          & {\underline{0.85725}}    & 0.92243          & 0.86085          & 0.99672          & 0.98913          & 0.99351          & {\underline{7.75539}}    & 9.37745          & 11.43241         \\ \hline
CC+GC+CG-ACRF & {\underline{0.68489}}    & \textbf{0.90338} & \textbf{0.87291} & 0.80651          & 0.92363          & {\underline{0.86989}}    & \textbf{0.99746} & \textbf{0.99307} & {\underline{0.99592}}    & 7.80448          & \textbf{3.56898} & \textbf{4.03629} \\ \hline
\end{tabular}
\end{adjustbox}
\label{table:ablationstudy}
\end{table*}

From Table \ref{table:ablationstudy}, we observe that the GC obtains better performance than CC. For the dice score, GC achieves 0.89365 for the entire tumor and 0.82246 for the tumor core. CC only achieves a dice score of 0.87467 on the entire tumor and 0.82068 on the tumor core, which is 2\% and 0.2\% lower than those by GC respectively. For hausdorff95, GC achieves 6.40312 on the entire tumor and 5.81216 on the tumor core. CC achieves 6.88633 and 7.93923, which are 0.49321 and 2.12707 higher than those of GC on the entire tumor and the tumor core, respectively. From Fig. \ref{fig:ablation}, we observe that GC can accurately predict individual regions. For example, the GD-enhanced tumor region normally does not appear at the outside of the tumor region. This superior performance may benefit from the information learned from the feature interactive graph as the feature nodes of different tumor regions have a strong structural association between them. Learning the relationship may help the system to predict correct labels of the tumor regions. However, the sensitivity of GC is much higher than that of CC. In Table \ref{table:ablationstudy}, for example, the sensitivity score of GC is higher than that of CC: 12.02\% higher on the enhancing tumor, 4.469\% higher on the entire tumor, 8.104\% higher on tumor core, respectively. We observe poor segmentation results at the NCR/ECT region by GC, worse than CC and the ground truth shown in Fig. \ref{fig:ablation}.

We then evaluate the complete HCA-FE with the extracted feature maps by CC and GC simultaneously. Here, we fuse the feature maps of CC and GC using a naive concatenation method. The HCA-FE has less over-segmentation results, depicted in Table \ref{table:ablationstudy}, where the sensitivity of CC+GC is much lower than that of GC. The sensitivity of CC+GC is  0.85725 on the enhancing tumor (ET), 0.92243 on the whole tumor (WT) and 0.86085 on the tumor core (TC), respectively. From Fig. \ref{fig:ablation}, we witness that by introducing the complete HCA-FE, the segmentation model can correct some misclassified regions produced by CC. However, the concatenation fusion method does not demonstrate any benefit to the overall segmentation. CC+GC has a dice score of 0.86073 on the whole tumor and 0.80306 on the tumor core, which is 3.292\% and 1.94\% lower than those of GC respectively. We also observe the loss of the boundary information in Fig. \ref{fig:ablation}, especially the boundaries of NCR/ECT and GD-enhancing tumors excessively shrinks compared with those of GC and CC.

We finally evaluate the effectiveness of our proposed CG-ACRF. By introducing the CG-ACRF fusion module, our segmentation model outperforms the other methods. Benefiting from the inference ability of CG-ACRF, it presents a satisfactory segmentation output. For the whole tumor and the tumor core, its Dice scores are 0.90338 and 0.87291 respectively, which are the top scores in the leader-board. Its Hausdorff95 also is the lowest. For the whole tumor and the tumor core, its hausdorff95 values are 3.56898 and 4.03629 respectively. Referring to much lower sensitivity scores reported in Table \ref{table:ablationstudy}, we conclude that the superior performance has been achieved by the complete CANet. The same conclusion can be drawn from Fig. \ref{fig:ablation} where CG-ACRF can detect optimal feature maps that benefit the downstream deconvolution networks and outline small tumor cores and edges, which may be lost when we use a down-sampling operation in the encoder backbone.

\subsection{Iteration Test}
\begin{figure*}[!ht]
    \centering
    \includegraphics[width=\textwidth]{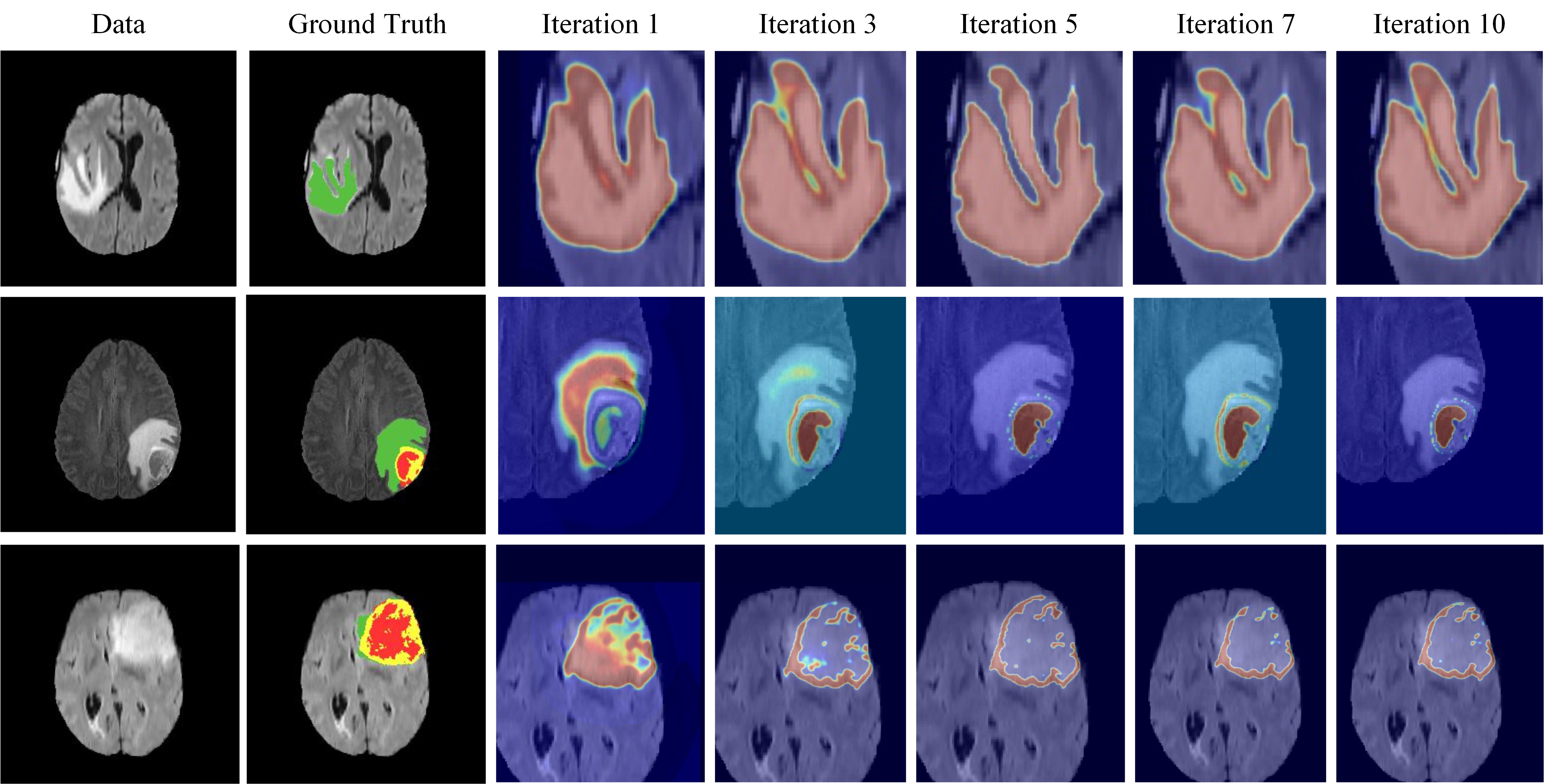}
    \caption{Examples to illustrate the effectiveness of different iteration numbers by mean-field approximation in CG-ACRF. Columns from top to bottom represent different patient cases. Rows from left to right indicate FLAIR data, ground truth annotation, attentive map generated by CANet with different iteration numbers (from 1 to 10) in CG-ACRF respectively. Best viewed in colors.}
    \label{fig:ProbMap}
\end{figure*}

As described in Algorithm 1, we manually set the iteration number in the mean-field approximation of CG-ACRF. Since the mean-field approximation cannot guarantee a convergence point, we examine the effectiveness of different iteration numbers. Table \ref{table:convmf} reports the quantitative result of using different iteration numbers, i.e. 1, 3, 5, 7, and 10. With the increase of iterations, our proposed model performs better. However, no additional benefit is gained when the iteration number becomes over 5. Fig. \ref{fig:ProbMap} presents the probability map during segmentation, where the light color represents the region with a lower probability while the dark color represents the area with a higher probability. We observe that using only one iteration, CANet can outline the region of interest using the fused feature maps. By increasing the iteration number to 3 or 5, CG-ACRF can gradually extract an optimal feature map, leading to accurate segmentation. We further increase the iteration number to 7 and 10 but no further improvement has been made. Therefore, we set the iteration number to 5 as a good trade-off between the segmentation performance and the number of the engaged parameters.

\begin{table*}[!ht]
\centering
\caption{Quantitative results of different iteration numbers by CG-ACRF mean-field approximation on the five fold cross-validation of the BraTS2017 training set with respect to Dice, Sensitivity, Specificity and Hausdorff95. The best result is in bold and the runner-up result is underlined.}
\begin{adjustbox}{width=1\textwidth}
\begin{tabular}{|c|c|c|c|c|c|c|c|c|c|c|c|c|}
\hline
             & \multicolumn{3}{c|}{Dice}                              & \multicolumn{3}{c|}{Sensitivity}                       & \multicolumn{3}{c|}{Specificity}                       & \multicolumn{3}{c|}{Hausdorff95}                       \\ \hline
Iteration \# & ET               & WT               & TC               & ET               & WT               & TC               & ET               & WT               & TC               & ET               & WT               & TC               \\ \hline
1            & 0.65697          & 0.86066          & 0.79005          & \textbf{0.90075} & 0.92017          & 0.85205          & 0.9953           & 0.98993          & {\underline{0.99426}}    & 7.99666          & 7.74909          & 10.48848         \\ \hline
3            & 0.68131          & {\underline{0.87267}}    & {\underline{0.80679}}    & {\underline{0.87265}}    & 0.92288          & {\underline{0.86948}}    & 0.99638          & {\underline{0.99007}}    & 0.99397          & \textbf{7.61352} & {\underline{6.8011}}     & {\underline{8.94057}}    \\ \hline
7            & 0.6643           & 0.85534          & 0.76902          & 0.85384          & 0.92108          & 0.86033          & 0.99644          & 0.98955          & 0.99336          & 9.84976          & 9.72             & 12.04193         \\ \hline
10           & {\underline{0.68484}}    & 0.85043          & 0.7839           & 0.83708          & \textbf{0.93128} & 0.85847          & {\underline{0.99675}}    & 0.98757          & 0.99383          & 8.06683          & 11.14894         & 11.64947         \\ \hline
Ours(5)    & \textbf{0.68489} & \textbf{0.90338} & \textbf{0.87291} & 0.80651          & {\underline{0.92363}}    & \textbf{0.86989} & \textbf{0.99746} & \textbf{0.99307} & \textbf{0.99592} & {\underline{7.80448}}    & \textbf{3.56898} & \textbf{4.03629} \\ \hline
\end{tabular}
\end{adjustbox}
\label{table:convmf}
\end{table*}

\subsection{Comparison with State-of-The-Art methods}
We choose several state-of-the-art deep learning model based brain tumor segmentation methods, including 3D UNet \cite{cciccek20163d}, Attention UNet \cite{oktay2018attention}, PRUNet \cite{brugger2019partially}, NoNewNet \cite{isensee2018no} and 3D-ESPNet \cite{mehta20193despnet}. We first consider the 5-fold cross-validation on the BraTS2017 training set. Each fold contains randomly chosen 228 cases for training and 57 cases for validation. In these cross-validation experiments on the training set, we consider CANet with complete HCA-FE and CG-ACRF fusion module with 5-iteration, which leads to the best performance in the ablation tests. As shown in Table \ref{table:SOTACompare}, our CANet outperforms the rest State-of-The-Art methods on several metrics while the results of the other metrics are competitive. The Dice score of CANet is 0.90338 and 0.87291 for the whole tumor and the tumor core respectively. The former is 8\% higher and the latter is 3\% higher than individual runner-up results. The Hausdorff95 values of CANet are 3.56898 and 4.03629 for the whole tumor and the tumor core, which are much lower than the runner up scores, i.e. 4.15649 and 5.77847, respectively.

\begin{figure*}[!ht]
    \centering
    \includegraphics[width=\textwidth]{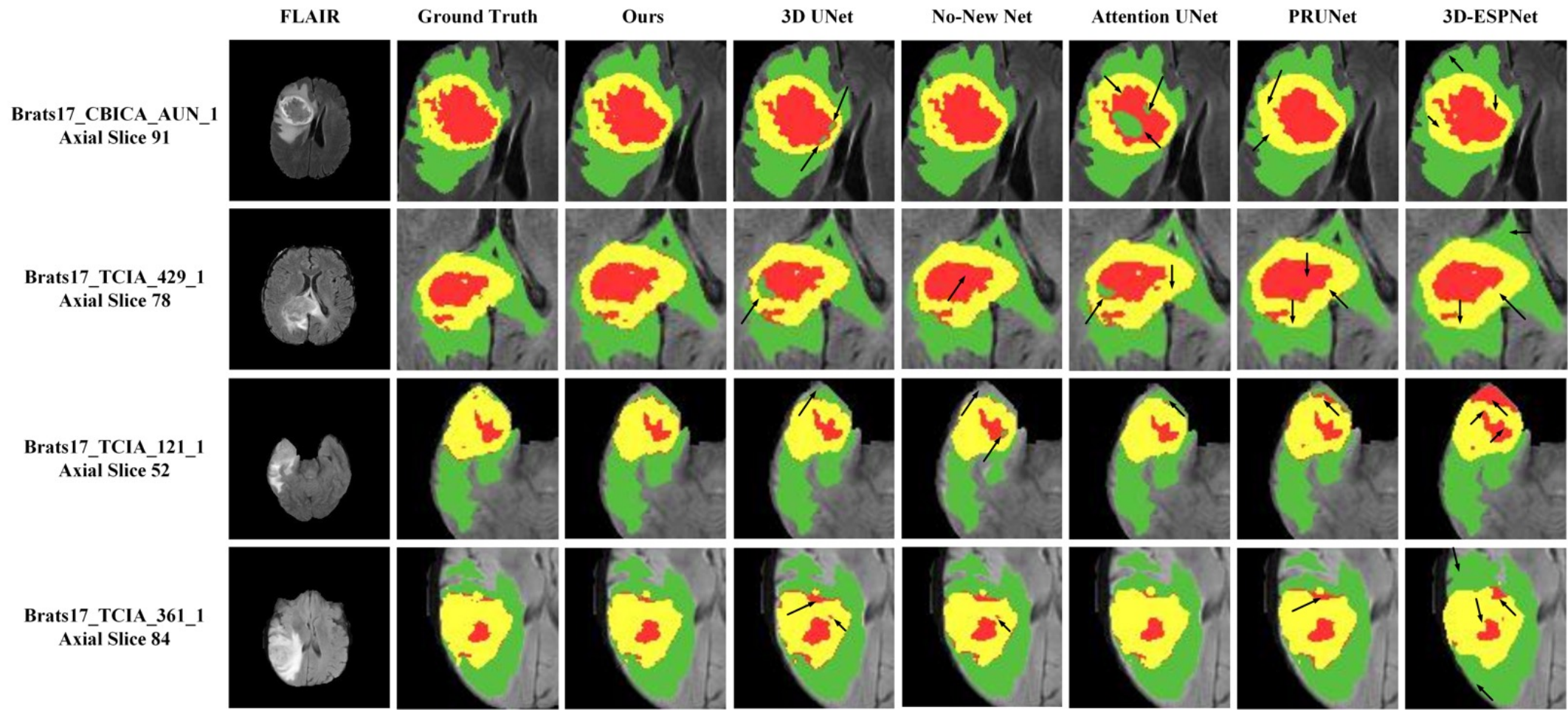}
    \caption{Examples of segmentation results by cross validation for the BraTS2017 training set. Qualitative comparisons with other brain tumor segmentation methods are presented. The eight columns from left to right show the frames of the input FLAIR data, the ground truth annotation, the results generated from our CANet (UNet encoder backbone with HCA-FE and 5-iteration CG-ACRF), 3DUNet \cite{cciccek20163d}, NoNewNet \cite{isensee2018no}, Attention UNet \cite{oktay2018attention}, PRUNet \cite{mehta20193despnet}, respectively. Black arrows indicate the failure in these comparison methods. Best viewed in colors.}
    \label{fig:SOTA}
\end{figure*}

\begin{figure*}[!ht]
    \centering
    \includegraphics[width=0.5\textwidth]{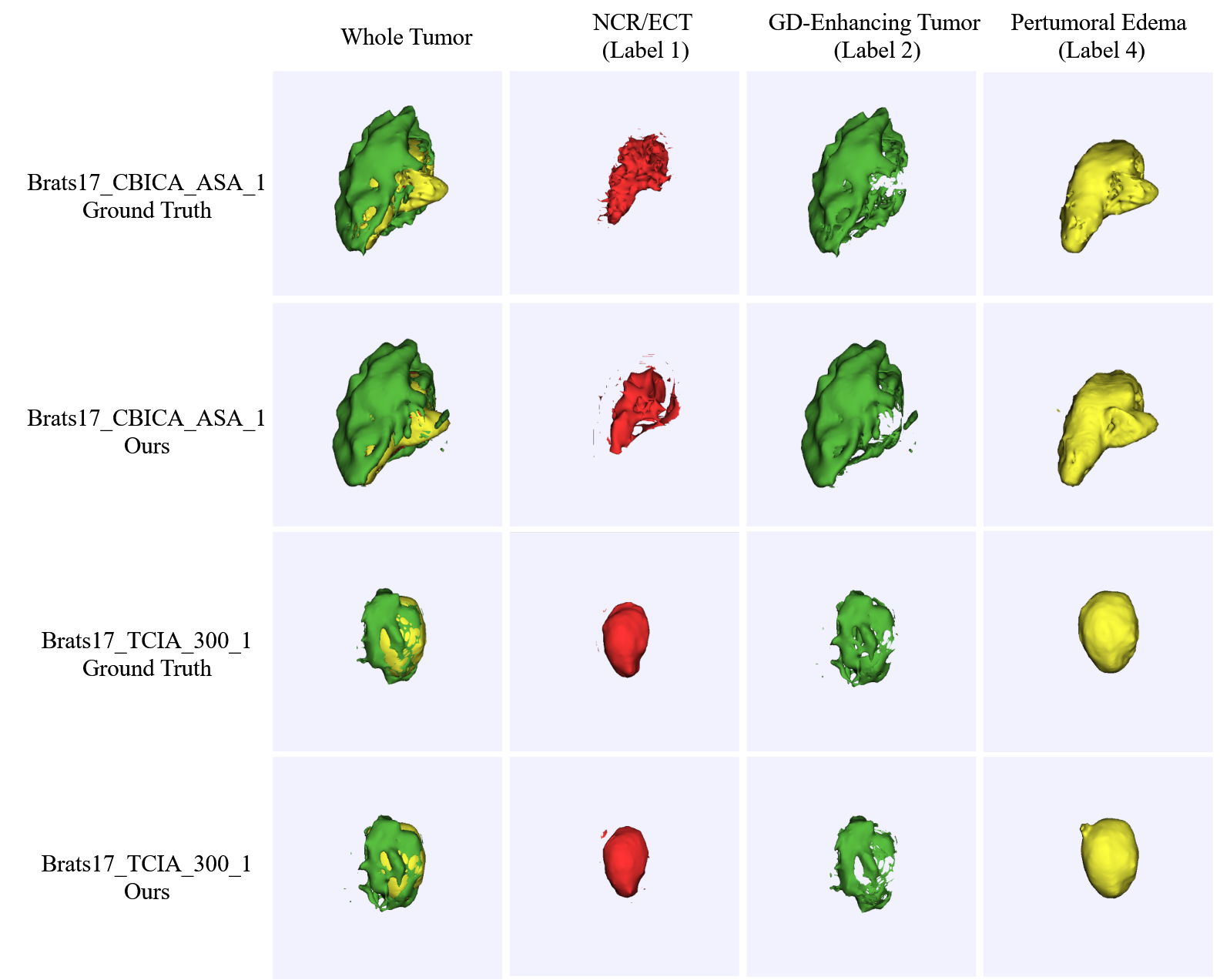}
    \caption{3D segmentation results of two volume cases by cross-validation on the BraTS2017 training set. The first and the third rows indicate the ground truth annotation. The second and the fourth rows indicate the segmentation result of our proposed CANet with HCA-FE and 5-iteration CG-ACRF. Rows from left to right indicate the qualitative comparison for the whole tumor, NCR/ECT, GD-enhancing tumor and Pertumoral Edema respectively. Best viewed in colors.}
    \label{fig:3D}
\end{figure*}

To further evaluate the segmentation output, we compare the segmentation output of the proposed approach against the ground-truth. Fig. \ref{fig:SOTA} shows that the proposed CANet can effectively predict the correct regions including small tumor cores and complicated edges while the other state of the art methods fail to do so. Fig. \ref{fig:3D} presents the example segmentation result and the ground-truth annotation in 3D visualisation. From Fig. \ref{fig:3D}, we can observe that our proposed CANet effectively captures 3D forms and shape information in all different circumstances.

\begin{figure*}[!ht]
    \centering
    \includegraphics[width=0.5\textwidth]{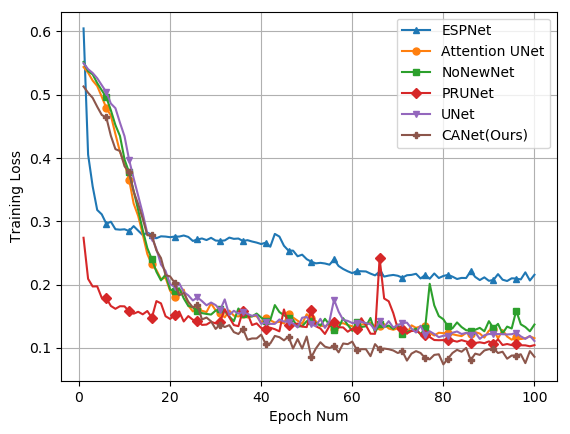}
    \caption{The learning curve of the state of the art methods and our proposed CANet with HCA-FE and 5-iteration CG-ACRF. Best viewed in color.}
    \label{fig:training record}
\end{figure*}

Fig. \ref{fig:training record} reports the training curve of CANet and the other state-of-the-art methods. Our proposed method converges to a lower training loss using fewer epochs. Taking the advantage of the powerful HCA-FE and the proposed fusion module CG-ACRF, CANet achieves satisfactory outlining for the brain tumors. With the training epoch increasing, CANet can fine-tune the segmentation map and successfully detect small tumor cores and boundaries.\\

\begin{table*}[!ht]
\centering
\caption{Quantitative results of the state-of-the-art models by cross-validation for the BraTS2017 training set with respect to dice, sensitivity, specificity and hausdorff. The best result is shown in bold and the runner-up result is underlined.}
\begin{adjustbox}{width=\textwidth}
\begin{tabular}{|c|c|c|c|c|c|c|c|c|c|c|c|c|}
\hline
               & \multicolumn{3}{c|}{Dice}                              & \multicolumn{3}{c|}{Sensitivity}                       & \multicolumn{3}{c|}{Specificity}                       & \multicolumn{3}{c|}{Hausdorff95}                       \\ \hline
Model          & ET               & WT               & TC               & ET               & WT               & TC               & ET               & WT               & TC               & ET               & WT               & TC               \\ \hline
3D-UNet\cite{cciccek20163d}        & 0.70646          & 0.86492          & 0.81032          & 0.80275          & 0.9064           & 0.82906          & 0.99791          & 0.99005          & 0.99493          & 6.62407          & 8.19351          & 8.95848          \\ \hline
No-New Net\cite{isensee2018no}     & {\underline{0.74108}}    & 0.87083          & 0.8125           & 0.76688          & 0.89296          & 0.83115          & \textbf{0.99853} & {\underline{0.99243}}    & 0.99528          & \textbf{3.93033} & 7.05536          & 7.64098          \\ \hline
Attention UNet\cite{oktay2018attention} & 0.67174          & 0.8634           & 0.77837          & \textbf{0.84741} & 0.9001           & 0.86171          & 0.99591          & 0.98961          & 0.99186          & 9.34711          & 9.67562          & 10.66793         \\ \hline
PRUNet\cite{brugger2019partially}         & 0.71015          & 0.89072          & 0.81447          & 0.78826          & 0.90028          & 0.84056          & {\underline{0.99804}}    & 0.99002          & 0.99586          & 7.20534          & 7.41411          & 9.1874           \\ \hline
3D-ESPNet\cite{mehta20193despnet}      & 0.68949          & {\underline{0.89548}}    & {\underline{0.84397}}    & 0.80535          & \textbf{0.94666} & \textbf{0.88085} & 0.99671          & 0.99026          & {\underline{0.99677}}    & 6.89359          & {\underline{4.15649}}    & {\underline{5.77847}}    \\ \hline
CANet (Ours)   & 0.68489          & \textbf{0.90338} & \textbf{0.87291} & 0.80651          & {\underline{0.92363}}    & {\underline{0.86989}}    & 0.99746          & \textbf{0.99307} & 0.99592          & 7.80448          & \textbf{3.56898} & \textbf{4.03629} \\ \hline
\end{tabular}
\end{adjustbox}
\label{table:SOTACompare}
\end{table*}

\begin{table*}[!ht]
\centering
\caption{Quantitative results comparison between CANet and other state of the art results on the BraTS2017 validation set for Dice and Hausdorff95. The best results of these methods are underlined. The bold shows the best score of each tumor region by single prediction approaches. '-' depicts that the result of the associated method has not been reported yet.}
\begin{adjustbox}{width=1\textwidth}
\begin{tabular}{|cc|ccc|ccc|}
\hline
                  &                  & \multicolumn{3}{c|}{\textbf{Dice}}               & \multicolumn{3}{c|}{\textbf{Hausdorff95}}        \\ \hline
\textbf{Approach} & \textbf{Method}  & \textbf{ET}    & \textbf{WT}    & \textbf{TC}    & \textbf{ET}    & \textbf{WT}    & \textbf{TC}    \\ \hline
                  & Kamnitsas et al. \cite{kamnitsas2017ensembles} & 0.738          & 0.901          & 0.797          & 4.500          & 4.230          & 6.560          \\
                  & Wang et al. \cite{wang2017automatic}      & {\underline {0.786}}    & {\underline {0.905}}    & {\underline {0.838}}    & {\underline {3.282}}    & {\underline {3.890}}    & {\underline {6.479}}    \\
Ensemble          & Zhao et al. \cite{zhao20173d}     & 0.754          & 0.887          & 0.794          & -              & -              & -              \\
                  & Isensee et al. \cite{isensee2017brain}  & 0.732          & 0.896          & 0.797          & 4.550          & 6.970          & 9.480          \\
                  & Jungo et al. \cite{jungo2017towards}    & 0.749          & 0.901          & 0.790          & 5.379          & 5.409          & 7.487          \\ \hline
                  & Islam et al. \cite{islam2017multi}     & 0.689          & 0.876          & 0.761          & 12.938         & 9.820          & 12.361         \\
                  & Jesson et al. \cite{jesson2017brain}    & 0.713          & \textbf{0.899} & 0.751          & 6.980          & \textbf{4.160} & \textbf{8.650} \\
Single Prediction & Roy et al. \cite{roy2018recalibrating}      & 0.716          & 0.892          & 0.793          & 6.612          & 6.735          & 9.806          \\
                  & Pereira er al. \cite{pereira2019adaptive}   & 0.719          & 0.889          & 0.758          & 5.738          & 6.581          & 11.100         \\
                  & CANet (Ours)    & \textbf{0.728} & 0.892          & \textbf{0.821} & \textbf{5.496} & 7.392          & 10.122         \\ \hline
\end{tabular}
\end{adjustbox}
\label{table:17val}
\end{table*}

\begin{figure*}[!ht]
    \centering
    \includegraphics[width=0.6\textwidth]{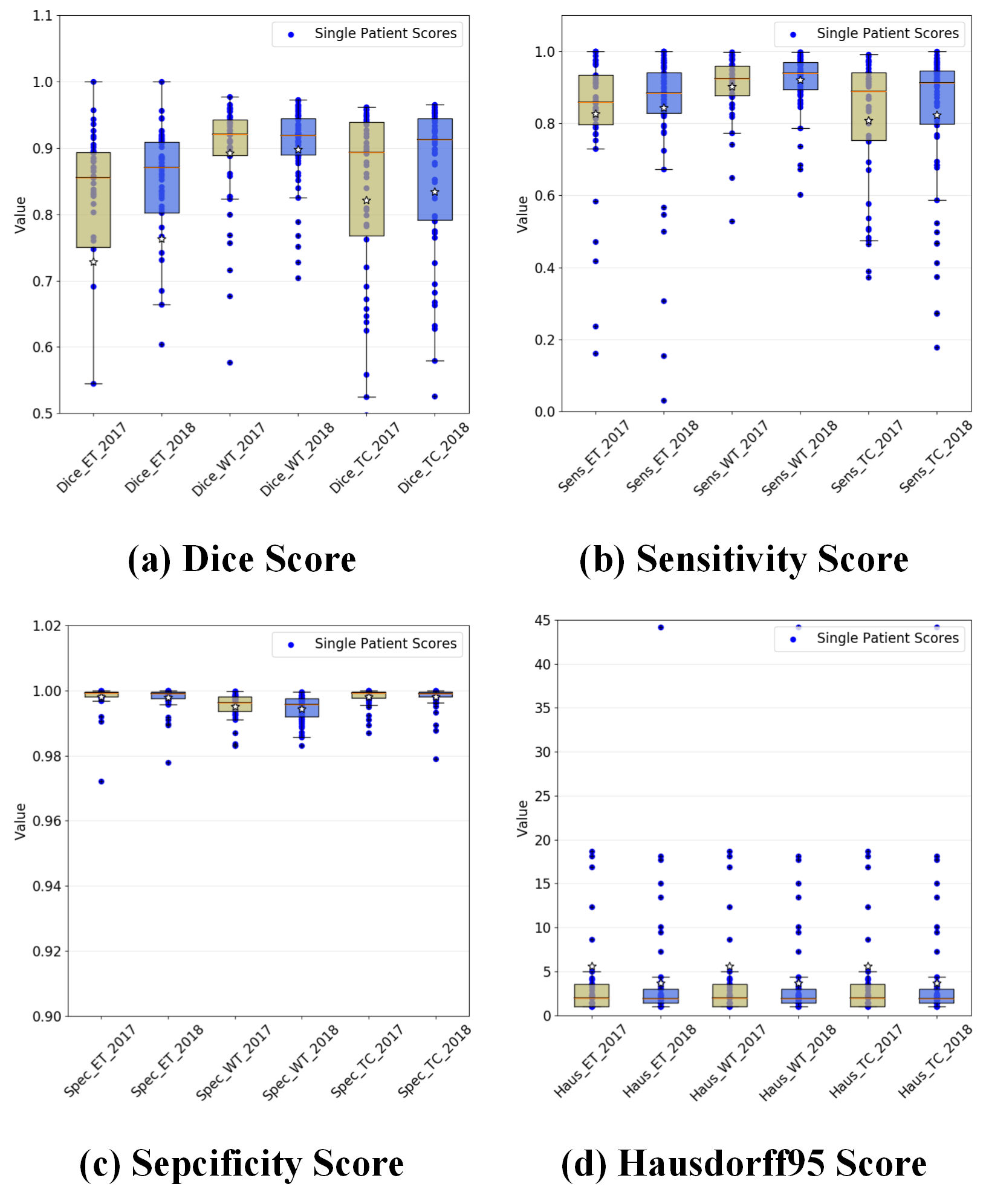}
    \caption{Boxplot of the segmentation results by CANet with HCA-FE and 5-iteration CG-ACRF. Dots within yellow boxes are individual segmentation results generated for the BraTS2017 validation set. Dots within blue boxes are individual segmentation results generated for the BraTS2018 validation set. Best viewed in color.}
    \label{fig:boxplot}
\end{figure*}

\begin{table*}[!ht]
\centering
\caption{Quantitative results of the BraTS2018 validation set with respect to Dice and Hausdorff95. The best results of these methods are underlined. The bold results show the best score of each tumor region using single prediction approaches. '-' represents the result of the associated method has not been reported yet.}
\begin{adjustbox}{width=1\textwidth}
\begin{tabular}{|cc|ccc|ccc|}
\hline
                                  &                                  & \multicolumn{3}{c|}{\textbf{Dice}}                                                                           & \multicolumn{3}{c|}{\textbf{Hausdorff95}}                                                                    \\ \hline
\textbf{Approach} & \textbf{Method} & \textbf{ET}    & \textbf{WT}    & \textbf{TC}    & \textbf{ET}    & \textbf{WT}    & \textbf{TC}    \\ \hline
                                  & Isensee et al. \cite{isensee2018no}                   & \uline{0.796}  & 0.908                           & 0.843                           & 3.120                           & 4.790                           & 8.020                           \\
                                  & McKinley et al. \cite{mckinley2018ensembles}                  & 0.793                           & 0.901                           & \uline{0.847}  & 3.603                           & 4.062                           & \uline{4.988}  \\
Ensemble                           & Zhou et al. \cite{zhou2018learning}                      & 0.792                           & \uline{0.907}  & 0.836                           & \uline{2.800}  & 4.480                           & 7.070                           \\
                                  & Cabezas et al. \cite{cabezas2018survival}                   & 0.740                           & 0.889                           & 0.726                           & 5.304                           & 6.956                           & 11.924                          \\
                                  & Feng et al. \cite{feng2018brain}                      & 0.787                           & 0.906                           & 0.834                           & 3.964                           & \uline{4.018}  & 5.340                           \\ \hline
                                  & Sun et al. \cite{sun2018tumor}                       & 0.751                           & 0.865                           & 0.720                           & -                               & -                               & -                               \\
                                  & Myronenko \cite{myronenko20183d}                        & \textbf{0.816} & \textbf{0.904} & \textbf{0.860} & \textbf{3.805} & \textbf{4.483} & 8.278                           \\
Single Prediction                  & Weninger et al. \cite{weninger2018segmentation}                  & 0.712                           & 0.889                           & 0.758                           & 8.628                           & 6.970                           & 10.910                          \\
                                  & Gates et al. \cite{gates2018glioma}                    & 0.678                           & 0.806                           & 0.685                           & 14.523                          & 14.415                          & 20.017                          \\
                                  & CANet(Ours)                     & 0.767                           & 0.898                           & 0.834                           & 3.859                           & 6.685                           & \textbf{7.674} \\ \hline
\end{tabular}
\end{adjustbox}
\label{table:18val}
\end{table*}
We further investigate the segmentation results on the BraTS2017 and BraTS2018 validation sets, where the quantitative result of each patient case is generated from the online evaluation server. The mean result is reported in Table \ref{table:17val} and Table \ref{table:18val}. Box plot in Fig. \ref{fig:boxplot} shows the distribution of the segmentation result among all the patient cases in the validation set. For the BraTS2017 validation set, our proposed CANet with complete HCA-FE and 5-iteration CG-ACRF achieves the state-of-the-art results of Dice on ET, Dice on TC and Hasdorff95 on ET among the single model segmentation benchmarks. Our CANet has Dice on ET of 0.728, which is higher than the approach reported in \cite{pereira2019adaptive}. The Dice on TC by CANet is 0.821, which is higher than the runner-up result reported in \cite{roy2018recalibrating}. The Hausdorff95 on ET of CANet is 5.496, which is much lower than the runner-up generated in \cite{pereira2019adaptive}. For the BraTS2018 validation set, our proposed CANet achieves the state-of-the-art result for Hausdorff95, i.e. 7.674, on the tumor core, while the other results are all runner-ups. Note that the method proposed by Myronenko \cite{myronenko20183d} has the best performance using most of the evaluation metrics. In Myronenko's method, they set up an additional branch using autoencoder to regularise the encoder backbone by reconstructing the input 3D MRI image. This autoencoder branch greatly enhances the feature extraction capability of the backbone encoder. In our framework, we regularise the network weights using a L2-regularisation without any additional branch, and the result of our proposed CANet is better than the other single prediction methods. Be reminded that the standard single prediction models generate the segmentation only using one network, and do not need much computational resources and a complicated voting scheme. Compared with the ensemble methods, the result of our proposed CANet is still very competitive.

Note that even we incorporate our designed CG-ACRF as iterative convolutional blocks within the system, our model still maintains a relatively small space. Our final model, i.e. HCA-FE with five times convolution approximation for CG-ACRF maintains a parameter size at 1.84E7. Compared with baseline methods such as Isensee et al. \cite{isensee2018no} with a parameter size at 1.45E7 and Myronenko et al. \cite{myronenko20183d} with a parameter size at 2.01E7 \cite{zhang2020exploring}, our system maintains the parameter size at an intermediate level, which prevents the occurrence of over-fitting.

\subsection{Failure Case Studies}
\begin{figure*}[!ht]
    \centering
    \includegraphics[width=\textwidth]{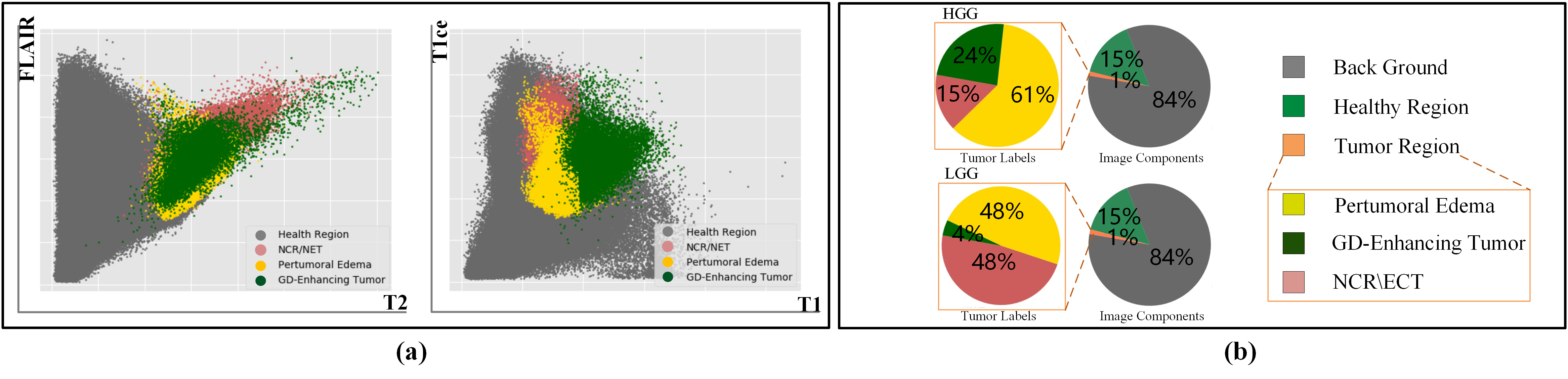}
    \caption{Statistics of the BraTS2017 training set. The left-hand side figure of (a) shows the FLAIR and T2 intensity projection, and the right-hand side figure shows the T1ce and T1 intensity projection. (b) is the pie chart of the training data with labels, where the top figure shows the HGG data labels while the bottom figure shows the LGG labels. There are a large region and label imbalance cases here. Best viewed in colors.}
    \label{fig:imbalance}
\end{figure*}

\begin{figure*}[!ht]
    \centering
    \includegraphics[width=\textwidth]{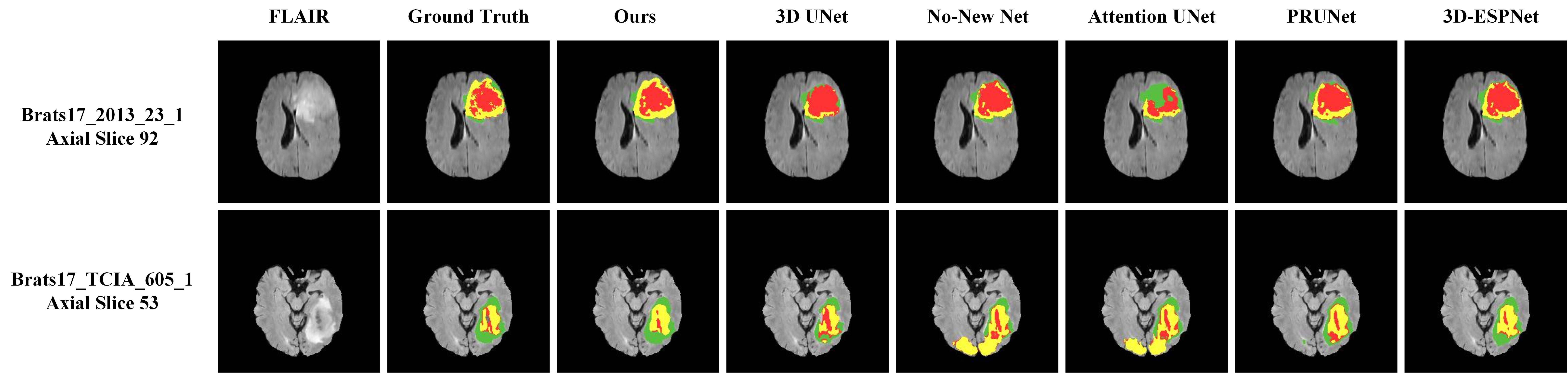}
    \caption{Qualitative comparisons in the failure cases. Rows from left to right indicate the input data of the FLAIR modality, ground truth annotation, segmentation result from our CANet, segmentation result from the other SOTA methods respectively. Our results look better than the SOTA methods results. Best viewed in colors.}
    \label{fig:FailureCase}
\end{figure*}

Fig. \ref{fig:imbalance} shows the statistical information of the BraTS2017 training set. As an example, we here report two failure segmentation cases by our proposed approach, which are shown in Fig. \ref{fig:FailureCase}. During the whole training process, CANet focuses on extracting feature maps with different contextual information, e.g. convolutional and graph contexts. However, we have not designed specific strategies for handling the imbalanced issue of the training set. The imbalanced issue is presented in two aspects. Firstly, there exists an unbalanced number of voxels in different tumor regions. As the exemplar case named "Brats17\_TCIA\_605\_1" is shown in Fig. \ref{fig:FailureCase}, the NCR/ECT region is much smaller than the other two regions, suggesting the poor performance of segmenting NCR/ECT. Secondly, there exists an unbalanced number of patient cases from different institutions. This imbalance introduces an annotation bias where some annotations tend to connect all the small regions into a large region while the other annotation tends to label the voxels individually. As the exemplar case named "Brats17\_2013\_23\_1" is shown in Fig. \ref{fig:FailureCase}, the ground truth annotation tends to be sparse while the segmentation output tends to be connected. In future work, we will consider an effective training scheme based on active/transfer learning which can effectively handle the imbalance issue in the dataset. Despite the imbalance issue, our segmentation method on the overall cases qualitatively outperforms the other state-of-the-art methods.

\section{Discussion}
In summary, we have proposed a novel 3D MRI brain tumor segmentation approach called CANet. Considering different contextual information in standard and graph convolutions, we proposed a novel hybrid context-aware feature extractor combined with a deep supervised convolution and a graph convolution stream. Different from previous works that used naive feature fusion schemes such as element-wise summation or channel-wise concatenation, we here designed a novel feature fusion model based on the conditional random field called context guided attentive conditional random field (CG-ACRF), which effectively learns the optimal latent features for downstream segmentations. Furthermore, we formulated the mean-field approximation within CG-ACRF as convolutional operations, which incorporate the CG-ACRF in a segmentation network to perform end-to-end training. We conducted extensive experiments to evaluate the effectiveness of the proposed HCA-FE, CG-ACRF and the complete CANet frameworks. The results have shown that our proposed CANet achieved state-of-the-art results in several evaluation metrics. In the future, we consider combining the proposed network with novel training methods that can better handle the imbalance issue in the datasets.


\chapter{Hierarchical Homography Estimation Network for Medical Image Mosaicing} 

\label{Chapter4} 


\section{Introduction}
\label{C4:S1}

In Chapter \ref{Chapter3}, we introduced an example of implicit deep relational learning paradigm on medical image analysis, i.e. use graph convolution kernel to learn the correlation information between different features on the built feature graph. One of the shortcomings of this implicit relational learning is that it only focuses on the relationship of features on one image while ignoring the relation relevance of multiple images along the time dimension. To this end, we propose a novel Hierarchical Homography Estimation Network (HHEN) for medical image mosaicing, which can explicitly model the spatial relationship between frames in both short and long ranges.

Image mosaicing refers to sequentially transferring and combining a set of narrow field-of-view images to form a new image or video with a wider field-of-view. Traditional image mosaicing has been widely used in various areas such as satellite photographing \cite{bu2016map2dfusion}, augmented reality \cite{azzari2008markerless} and panoramic photo edition \cite{li2018deep}. In the last few years, the interest in medical image mosaicing has grown in the medical image analysis community, especially in fetoscopic laser photography mosaicing \cite{bano2020deep, bano2020deep2}. Clinical physicians use fetoscopic video to detect vascular anastomoses, make treatment plans and control surgical robots. Thus clinicians can be able to treat various diseases such as abnormal vascular anastomoses in twin-to-twin transfusion syndrome (TTTS). However, it is a challenging problem for clinicians to fully use the fetoscope due to its narrow FoV and low visibility. Image mosaicing can compose a set of fetoscopic images with correlation transformation to generate a new image with wider FoV, thus it can offer computer-assisted diagnosis evidence to effectively detect the vascular anastomoses location.

Traditional mosaicing methods commonly use low-level features, i.e. point-based mosaicing and HoG based mosaicing \cite{kanazawa2004image, zagrouba2009efficient, kourogi1999real}. However, traditional mosaicing methods do not perform well due to poor medical image quality, e.g. turbid amniotic fluid in TTTS treatment. Recently, several deep-learning-based mosaicing methods have been proposed \cite{lv2019improved, nguyen2018unsupervised}. These methods learned deep features or homography estimation automatically and achieved superior performance on various datasets. Despite the remarkable progress that has been made, one open challenge is that these methods only focus on learning local information between adjacent frames while ignoring the non-local information in long series, which is important for time serious tasks. To overcome this, we propose a novel hierarchical medical image mosaicing network using neural homography estimation. Our contributions are summarized as follows:

\begin{enumerate}
\item We propose a new deep hierarchical homography estimation network to automatically and hierarchically estimate the 8 degree-of-freedom homography, upon which multi-scale (local level between adjacent frames and non-local level between long-range frames) homography are jointly learned and optimized in a data-driven manner.

\item Inspired by recent works on color homography \cite{finlayson2017color}, we propose a new data generation method called Partially Image Generation (PIG). PIG only perturb the color, rotation, and translation movement between adjacent frames. The generated frames by PIG can be used for evaluating model performance during network training.

\item We conduct extensive experiments on five different video clips from UCL Fetoscopy Placenta dataset. The results show that our method achieves state-of-the-art performance and generalizes well on different clips under different numbers of frames and different acquisition methods.
\end{enumerate}

The rest of this Chapter is organized as follows. In Section \ref{C4:S2}, we briefly review related works on image mosaicing. In Section \ref{C4:S3}, we introduced our proposed method, including HHEN in Section \ref{C4:S3:SS1} and PIG in Section \ref{C4:S3:SS2} respectively. In Section \ref{C4:S4}, we demonstrate the dataset and implementation details. In Section \ref{C4:S5}, our proposed HHEN is evaluated and compared with the state-of-the-art methods. In addition, the results are analysed in detail. This Chapter is finally concluded in Section \ref{C4:S6}.

\section{Review of Recent Image Mosaicing Methods}
\label{C4:S2}
In this chapter, we briefly review related works on image mosaicking, which refers to the alignment of multiple overlapping images into a large combination with larger FoV.

\begin{figure}[t]
    \centering
    \includegraphics[width=\textwidth]{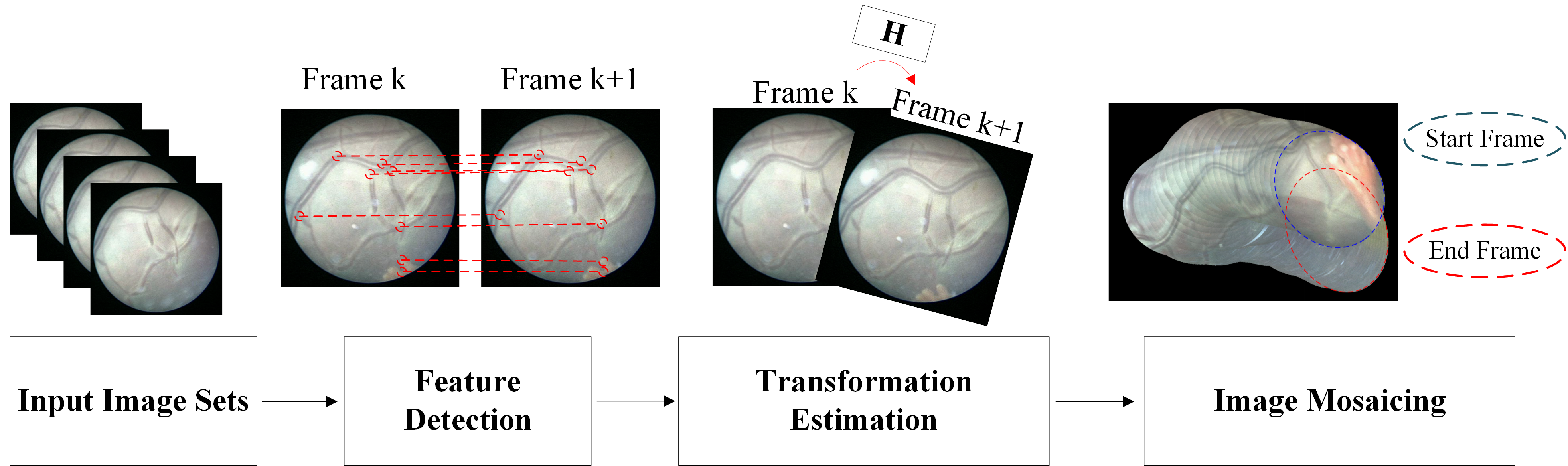}
    \caption{A schematic pipeline for image mosaicing.}
\label{fig:mosaicingflow}
\end{figure}

Finding the point-to-point or other feature correlation between two images is the fundamental part of conventional image mosaicking, which can be found in the mosaicing schematic pipeline shown in Fig. \ref{fig:mosaicingflow}. Traditional image mosaicing methods commonly use low-level features such as edge and point to find the correlation between two images. Kanazawa et al. \cite{kanazawa2004image} extract 100 feature points generated from the Harris operator to find the correlation between two images. Zagrouba et al. \cite{zagrouba2009efficient} extract both Harris points and the semantic segmentation as multi-level features for correlation learning. Kourogi et al. \cite{kourogi1999real} first generated the pseudo-motion vector on each pixel, and then estimate the affine motion parameters by using pixel-wise motion vector matching. Heikkila et al. \cite{heikkila2005image} extract the correlation feature using SIFT first and then reject the outlier points by applying RANSAC algorithms. Botterill et al. \cite{botterill2010real} first represent the image with the Bag-of-Words (BoW) method to find the adjoining frames and then use a perspective transformation to register adjoining frames to achieve video mosaicing.

Different from natural images, the quality of medical images is interfered with by many factors, such as fetoscopic photography captured within a turbid amniotic fluid environment. And traditional feature extraction algorithms for points or edges cannot extract effective and meaningful features. Although the aforementioned algorithms are simple and fast with low demands on computing resources. However, the aforementioned algorithms are difficult to be applied to tasks that require higher precision, such as medical image mosaicing.

Recently, various deep-learning-based image mosaicing methods have been presented. Lv et al. \cite{lv2019improved} proposed a CNN based Speed-up Robust Feature (SURF) extraction method for image mosaicing. Nguyen et al. \cite{nguyen2018unsupervised} proposed an unsupervised learning network for homography estimation in image mosaicing. However, these two methods’ input is still hand-crafted features, which cannot be jointly optimized with the mosaicing network. Zhang et al. \cite{zhang2019convolutional} use a CNN for registration based microscopic image mosaicing. Bano et al. \cite{bano2020deep} use a VGG-stylized CNN for homography estimation between two adjacent frames. However, these methods only focus on adjacent frame homography estimation (local information), while ignoring the correlation between long-range frames, which has been proven as an important factor for sequential frame-based tasks \cite{wang2018non, jiang2018modeling, anderson2018bottom}. To address this open challenge, we propose our novel hierarchical neural homography estimation network (HHEN) for fetoscopic photography mosaicing, which can automatically and hierarchically estimate the 6 degree-of-freedom homography. Besides, we propose a new generation method called Partially Image Generation (PIG). PIG only perturbs the color, rotation, and translation movement between adjacent frames. The generated frames by PIG can be used for evaluating model performance during training.

\section{Proposed Method}
\label{C4:S3}
Recently, deep image homography (DIH) estimation \cite{detone2016deep} have been proposed that use a regression neural network to estimate the homography between two local patches in one image. Based on DIH, several image mosaicing methods \cite{bano2020deep, bano2020deep2} have been proposed to stitching a set of images by estimating homography between every adjacent frame. However, DIH and proposed mosaicing work only focused on adjacent frame feature learning (local information) while ignored the correlation between long-range frames (non-local information), which has been proved as an important factor for achieving video and serious tasks. In this chapter, we first propose our novel image homography estimation framework called Hierarchical Homography Estimation Network (HHEN) in Chapter \ref{C4:S3:SS1}. HHEN treat the target frame as a query frame while transformed both local and non-local template frame as key and value frame. Thus the homography can be learned and optimize hierarchically with local and non-local information in a data-driven manner. We further propose a novel generation method called Partially Image Generation (PIG). PIG only controls the rotation, translation movement and color restoration between adjacent frames as the incision point of fetoscopic photography are fixed. Without any external sensors like electromagnetic tracker (EMT) used in \cite{tella2018probabilistic}, our proposed HHEN with PIG minimizes the drift error effectively and achieve the state-of-the-art result on fetoscopic photography mosaicing.

\subsection{Hierarchical Homography Estimation Network}
\label{C4:S3:SS1}

\begin{figure}[t]
    \centering
    \includegraphics[width=\textwidth]{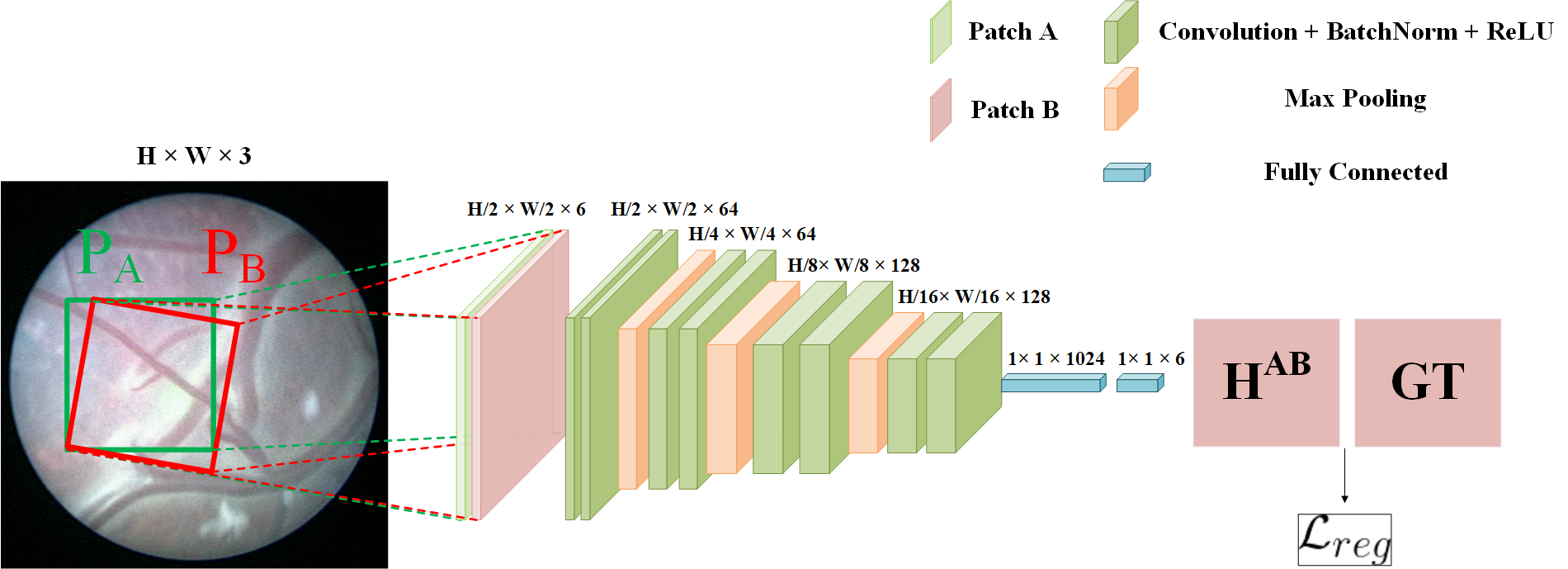}
    \caption{A schematic pipeline for Deep Image Homography (DIH) estimation.}
\label{fig:DIH}
\end{figure}

\begin{figure}[t]
    \centering
    \includegraphics[width=\textwidth]{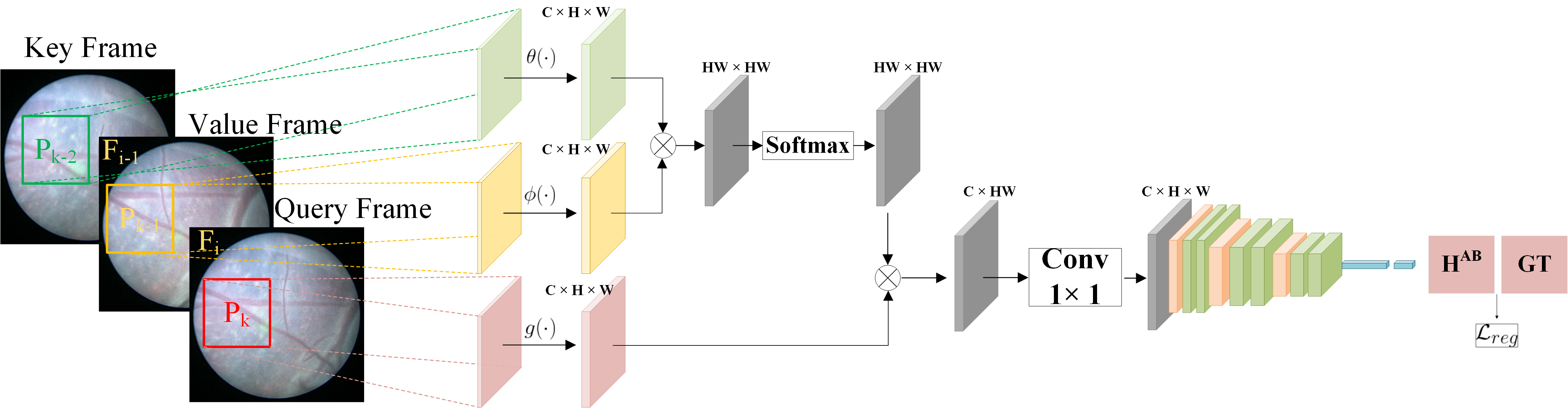}
    \caption{A schematic pipeline for Hierarchical Homography Estimation Netowork (HHEN) estimation.}
\label{fig:HHEN}
\end{figure}

A prilimary of our proposed HHEN is deep image homography (DIH) shown in Fig. \ref{fig:DIH}, where DIH estimates the homography between two relative image local patches $\left( P_{A} and P_{B} \right)$ from a single image. Similar to DIH, we also use a VGG style backbone network to estimate homography $\textbf{H}$. Instead of using $\textbf{H}$ with 9 degree-of-freedom, we here define the target homography is 3 point related due to the rotation and shear operation in \cite{detone2016deep} has little effect in our scenario (Due to the fetoscopic photography data is obtained through a fixed incision point). By defining arbitrary three corner points coordinate $\left( x_i, y_i \right)$ of $P_{A}$ and $\left( x_{i}^{\prime}, y_{i}^{\prime} \right)$ of $P_{B}$ with $i=1,2,3$, the 3 point related homography $\textbf{H}$ is defined as:

\[
\textbf{H} =
\begin{bmatrix}
\Delta x_1 & \Delta x_2 & \Delta x_3\\
\Delta y_1 & \Delta y_2 & \Delta y_3
\end{bmatrix}^{\top}
\]

where $\Delta x_i = x_{i}^{\prime} - x_{i}$, $\Delta y_i = y_{i}^{\prime} - y_{i}$. The fourth corner can be calculated due to the patch is extracted as a rectangle. Note that DIH generates $P_B$ by randomly displacing $P_A$ as illustrated in Fig. \ref{fig:DIH}, thus the drift error is not acceptable for image-based mosaicing tasks as the drift error will be accumulated. Mosaicing requires relative homography between adjacent frames with respect to the template frame. For minimizing the drift error in relative homography, several related works \cite{bano2020deep2, bano2020deep} have been proposed to let the backbone network learn the homography between patches extracted from adjacent frames. However, the homography estimation networks in \cite{bano2020deep2, bano2020deep} only considered the local information between adjacent frames while ignoring the non-local information between long-range consecutive frames, which has been proven as an important factor for video applications \cite{wang2018non, jiang2018modeling, anderson2018bottom}. Therefore, we propose a hierarchical homography estimation network (HHEN) that learns long-range information between a set of consecutive frames. As shown in Fig. \ref{fig:HHEN}, we extend the definition of a pair of adjacent frames to a hierarchical composition. We set the frame that needs to be moved (target frame) as \textit{Query Frame} while the template frame is assembled with long-range \textit{Key Frame} and short-range \textit{Value Frame}. Intuitively, for the target \textit{Query Frame}, the network estimates the homography with respect how \textit{Value Frame} moves based on previous non-local \textit{Key Frame}.

Inspired by non-local image de-noising operation \cite{buades2005non} and non-local neural network for video classification \cite{wang2018non}, we define the generic non-local frame fusion head in HHEN as:

\begin{equation}
y_{i} = G(x_{i}^{Q})\frac{1}{N(x)}\sum_{\forall j < i} F(x_{i}^{V}, x_{j}^{K})G(x_{j}^{Q})
\label{eqn:nonlocal}
\end{equation}

wherein Eqn. \ref{eqn:nonlocal}, $i$ is the position of the output in the spatial domain and $j$ is the neighborhood position of $i$ (we use the 8-neighbor pixels in this paper). $x^{Q}, x^{K}, x^{V}$ denotes the input image patch from the query frame, key frame and value frame respectively. $y$ is the output fusion feature map with the same size as $x$. $F(\cdot)$ is the pairwise function that takes the pairwise input $(x_{i}^{V} , x_{j}^{K})$. The output of $F(\cdot)$ is a scalar weight between two inputs, which represents the affinity transformation between $x_{i}^{V}$ and $x_{j}^{K}$. $G(\cdot)$ is the unary function that outputs the representation of the input signal. $N(\cdot)$ is the normalization factor. Overall, Eqn. \ref{eqn:nonlocal} denotes that, for a given input $x_{i}^{Q}$ at position $j$ from query frame $x^{Q}$, $F(\cdot)$ learns how the related input from value frame is updated based on a key frame, then the non-local operator uses the learned affinity transformation between value frame and key frame to update the feature computed by $G(\cdot)$ in query frame.

The natural instantiation choice  of $F(\cdot)$ is the Gaussian function as implemented in the non-local mean operator \cite{buades2005non}, which is formalized as:

\begin{equation}
F(x_{i}^{V}, x_{j}^{K}) = e^{{x_{i}^{V}}^\top x_{j}^{K}}
\label{eqn:gaussian}
\end{equation}

In Eqn. \ref{eqn:gaussian}, the ${x_{i}^{V}}^\top x_{j}^{K}$ is the dot product to measure the similarity between $x_{i}^{V}$ and $x_{j}^{K}$. Correspondingly, the normalization factor is $N(x) = \sum_{\forall j < i} F(x_{i}^{V}, x_{j}^{K})$. Since the entire system is trained in an end-to-end fashion, in order to enable the final regression loss to directly adjust the parameter update of the non-local operator, we expand the Gaussian function (Eqn. \ref{eqn:gaussian}) to an embedding manner:

\begin{equation}
F(F(x_{i}^{V}, x_{j}^{K})) = e^{\theta(x_{i}^{V}) \top \phi(x_{j}^{K})}
\label{eqn:gaussianembedding}
\end{equation}

In Eqn. \ref{eqn:gaussianembedding}, $\theta(\cdot)$ and $\phi(\cdot)$ are two linear embedding functions, i.e. $\theta(x_{i}^{V}) = W_{\theta}x_{i}^{V}$ and $\phi(x_{j}^{K}) = W_{\phi}x_{j}^{K}$. By using the embedding Gaussian function, given the normalization factor is $N(x) = \sum_{\forall j < i} F(x_{i}^{V}, x_{j}^{K})$, the implementation of the non-local fusion head can be regarded as a softmax activation with inputs from two convolution features generated from $1 \times 1$ convolution kernel:

\begin{equation}
\textbf{y} = G(\textbf{x}) \textit{softmax}(\textbf{x}^{\top} W_{\theta}^{\top} W_{\phi} \textbf{x})G(\textbf{x})
\label{eqn:nonlocalimpl}
\end{equation}

By choosing the same implementation using $1 \times 1$ convolution for $G(\cdot)$, the non-local fusion head can be jointly trained with the downstream network. The output of the HHEN is the homography $H^{Q}$ with respect to the target frame, i.e. \textit{Query Frame}. By iterate every frame in an image set as \textit{Query Frame} and compute every frame's $H^{Q}$, we can complete mosaicing all the image frames in the given image set. 

\subsection{Partially Image Generation}
\label{C4:S3:SS2}
Note that in \cite{detone2016deep}, the author used the MS-COCO dataset \cite{lin2014microsoft} for training and testing. Compared with the MS-COCO dataset which contains natural real images, fetoscopic photography contains particular artifacts such as color distortion caused by amniotic fluid particles. Also, the shear and scale contribute little effect when HHEN is trained with fetoscopic images as the fetoscopic image is obtained through a fixed incision point with constrained distance from the placenta. Thus to minimize the drift error during mosaicing, we propose our Partially Image Generation (PIG) generate the training dataset. PIG only assumes that only rotation, translation and color homography are related between the pair of key-value and value-query frames. For a given image $I_{A}$ and its extracts patch $P_{A}$ with three corners coordinates ($x_{i}, y_{i}$), ($i=1,2,3$). We applied:

\begin{itemize}
\item Rotation with angle $\alpha$ and translation with movement distance $\delta x$ on x coordinate and $\delta y$ on y coordinate.\\
\[
\begin{bmatrix}
x_{i}^{\prime} \\
y_{i}^{\prime}
\end{bmatrix} = \begin{bmatrix}
cos(\alpha) & sin(\alpha) \\
-sin(\alpha) & cos(\alpha)
\end{bmatrix} \begin{bmatrix}
x_{i} \\
y_{i}
\end{bmatrix} + \begin{bmatrix}
\Delta x \\
\Delta y
\end{bmatrix}
\]

\item Color homography transformation: we transfer the original color image with every pixel $\textit{R, G, B}$ value to the greyscale image with every pixel's intensity $\textit{Y}$:\\
\[ 
\textit{Y} = \frac{max(\textit{R, G, B}) + min(\textit{R, G, B})}{2}
\label{Eqn:PIAGrey}
\]
\end{itemize}

to obtain the corresponding pair image $I_{B}$ with perturbed patch $P_{B}$. We empirically set the rotation angle $\alpha$ and translation movement distance $\Delta x$, $\Delta y$ (Sec. \ref{C4:S4:SS2}). By implementing the PIG, the HHEN can learn the homography ($\alpha$, $\Delta x$ and $\Delta y$) without interference from color distortion (the color with artifacts are transformed into greyscale images).

\section{Experiment Setup}
\label{C4:S4}
We conduct extensive experiments to evaluate our proposed fetoscopic photography mosaicing method using HHEN with PIG. We first introduce the datasets and evaluation metrics first (Sec. \ref{C4:S4:SS1}). Then we express the details of method implementations (Sec. \ref{C4:S4:SS2}).

\subsection{Datasets and Evaluation Metrics}
\label{C4:S4:SS1}
We use the 5 fetoscopic video clips from UCL Fetoscopy Placenta Data \cite{bano2020vessel} \footnote{https://www.ucl.ac.uk/interventional-surgical-sciences/fetoscopy-placenta-data}, which includes two synthetic video clips (SYN1 and SYN2), one TTTS Phantom in the water video clip (TTTS1) and two in-vivo TTTS procedure video clips (INVT1 and INVT2). We show the details of all 5 datasets in Table \ref{table:mosaicingdata}. We can observe from Table \ref{table:mosaicingdata} that the datasets are varied among different factors such as object texture, ambient reflected light and capture motion, which post the challenges for mosaicing methods.

\begin{sidewaystable}
\begin{tabular}{|l|c|c|c|c|c|}
\hline
Frame Example      &   \raisebox{-\totalheight}{\includegraphics[width=0.15\textwidth, height=30mm]{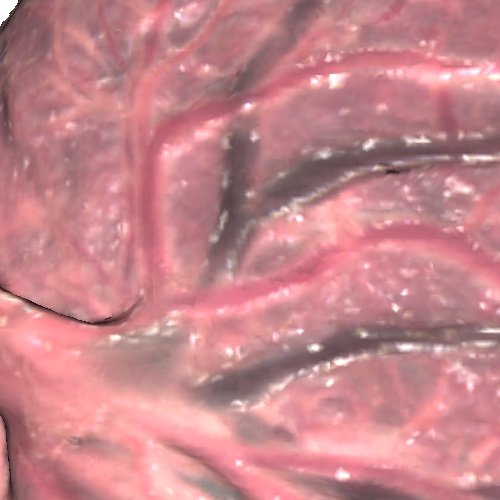}}         &\raisebox{-\totalheight}{\includegraphics[width=0.15\textwidth, height=30mm]{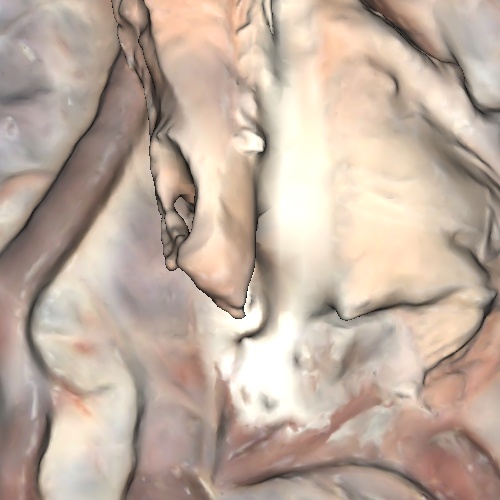}}           &\raisebox{-\totalheight}{\includegraphics[width=0.15\textwidth, height=30mm]{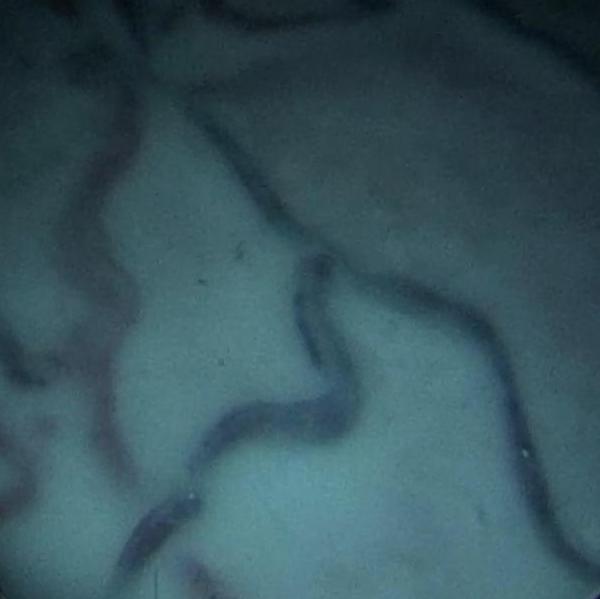}}                       & \raisebox{-\totalheight}{\includegraphics[width=0.15\textwidth, height=30mm]{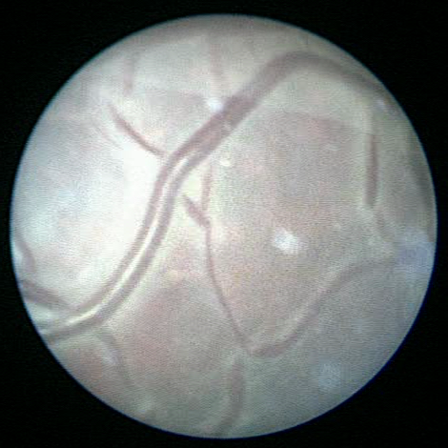}}                       & \raisebox{-\totalheight}{\includegraphics[width=0.15\textwidth, height=30mm]{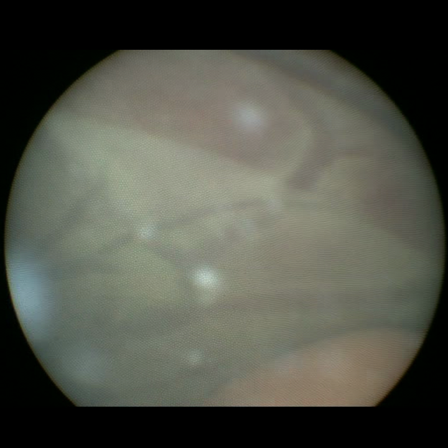}}                       \\ \hline
Data Type          & \begin{tabular}[c]{@{}c@{}}Synthetic\\ (SYN1)\end{tabular} & \begin{tabular}[c]{@{}c@{}}Synthetic\\ (SYN2)\end{tabular} & \begin{tabular}[c]{@{}c@{}}TTTS Phantom \\ in water\\ (TTTS1)\end{tabular} & \begin{tabular}[c]{@{}c@{}}In-vivo \\ TTTS Procedure\\ (INVT1)\end{tabular} & \begin{tabular}[c]{@{}c@{}}In-vivo \\ TTTS Procedure\\ (INVT2)\end{tabular} \\ \hline
Total Frame Num & 500       & 200       & 200                   & 400                    & 100                    \\ \hline
Image Resolution   & 500 x 500 & 500 x 500 & 600 x 600             & 448 x 448              & 448 x 448              \\ \hline
Camera View        & Planar    & Planar    & Planar                & Non Planar             & Planar                 \\ \hline
Motion Type        &Spiral & Circular & Circular                                                           & \begin{tabular}[c]{@{}c@{}}Exploratory \\ Freehand\end{tabular}   & \begin{tabular}[c]{@{}c@{}}Exploratory \\ Freehand\end{tabular}   \\ \hline
\end{tabular}
\caption{Details of the selected datasets for experimental analysis}
\label{table:mosaicingdata}
\end{sidewaystable}

We compare our method with state-of-the-art deep-learning-based methods: deep image homography (DIH) \cite{detone2016deep} and deep sequential mosaicking (DSM) \cite{bano2020deep} and a hand-crafted feature-based method (FEAT with SURF feature matching for homography estimation) \cite{brown2007automatic}. For the synthetic datasets SYN1 and SYN2, we evaluate our HHEN with PIG against all baseline methods and report the mean residual error (MSE) against ground truth data. For datasets obtained from TTTS clinical surgery (TTTS1, INVT1, INVT2), we evaluate the performance of our HHEN with PIG against all baseline methods and report the average root mean square error (RMSE) between image and generated image. Following \cite{brasch2018semantic, godard2019digging}, we also evaluate the average photometric error (APE) by using the $L1$ distance between two images generated by the estimated homography.

\subsection{Implementation Details}
\label{C4:S4:SS2}
We first randomly select 435 images from all datasets except the INVT1 to compose a training set. The image frames in the training set are not considered in the testing set as they are not a full video clip. INVT1 thus is unseen for the model during the training. By using PIG, all images are transferred into greyscale images where the intensity in Eqn. \ref{Eqn:PIAGrey} represent the brightness converted from RGB color to avoid color distortion. Our proposed network and other baseline methods are complied in Keras \footnote{https://keras.io/} platform with a Tensorflow \footnote{https://www.tensorflow.org/} backend. We train our proposed network on a single Nvidia Tesla T4 GPU (16GB) for 2000 epochs. We use the Stochastic Gradient Descent (SGD) as the optimizer with a learning rate $lr = 0.0001$ and momentum with $0.9$. Due to the limitation of GPU memory, we set the batch size as 16. We use the Euclidean loss function as the main goal of the homography is to estimate the distance of rotation and transition between adjacent frames. During training with PIG, we manually set the rotation angle $\alpha$ between ($-8^{\circ}, +8^{\circ}$) and the translation distance $\Delta x$, $\Delta y$ between ($-15, +15$) pixels. The manually perturbed rotation angle and transition distance are stored as ground truth. The network takes the original image patch and the perturbed image patch as inputs and outputs the numerical results. The loss can be estimated between the numerical outputs and the manually settled angel and transition distance.

\section{Experiment Results}
\label{C4:S5}

We conduct sufficient experiments on 5 video clips from UCL Fetoscopy Placenta Data against several baseline methods including FEAT, DIH and DSM. We discuss the experimental results from various perspectives in the following sections.

\begin{figure}[t]
    \centering
    \includegraphics[width=0.6\textwidth]{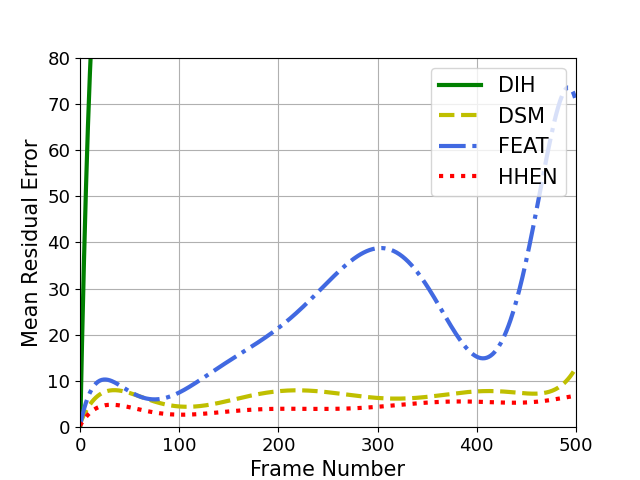}
    \caption{Quantitative result comparison on mean residual error between HHEN and baseline methods.}
\label{fig:MRE}
\end{figure}

\begin{figure}[t]
    \centering
    \includegraphics[width=0.6\textwidth]{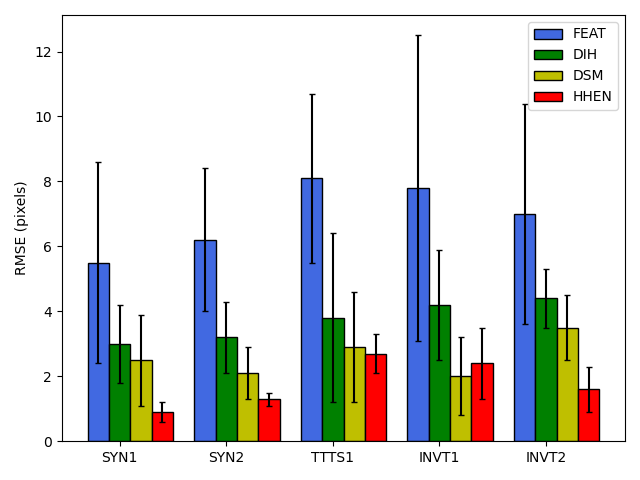}
    \caption{Quantitative result comparison on average RMSE between HHEN and baseline methods.}
\label{fig:RMSE}
\end{figure}

\begin{figure}[t]
    \centering
    \includegraphics[width=0.6\textwidth]{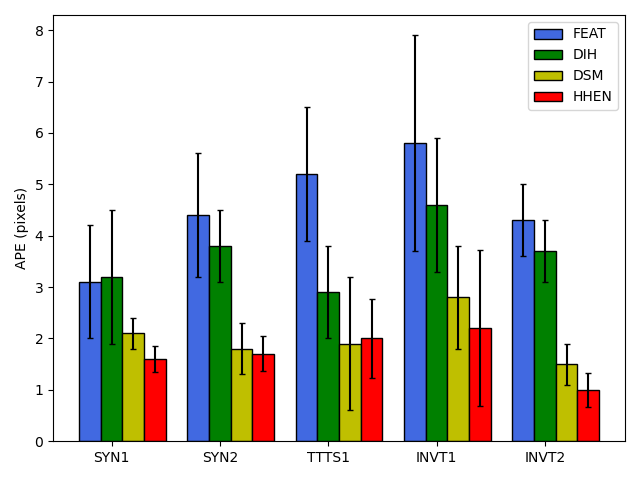}
    \caption{Quantitative result comparison on average photometric error between HHEN and baseline methods.}
\label{fig:APE}
\end{figure}

\subsection{Comparison with Hand-Crafted Methods}
\label{C4:S5:SS1}
We first compare the result between our proposed HHEN and FEAT. Note that FEAT  achieves the mosaicing based on the Speed Up Robust Features (SURF). SURF constructs a Hessian matrix to generate the point-of-interests on the edges. However, such kind of traditional-feature-based mosaicing methods fail on fetoscopic photography mosaicing due to the vessels are blurred and surrounded with turbid amniotic fluid. Compared with synthetic data (SYN1 and SYN2) which ignored these kinds of environmental interference (Table \ref{table:mosaicingdata}), the FEAT failed to achieve robust mosaicing on real data (TTTS1, INVT1, INVT2). As shown in Fig. \ref{fig:RMSE}, FEAT got relatively low RMSE on TTTS1, INVT1 and INVT2 with scores 8.13, 7.86, 7.11 respectively. However, the HHEN benefits from deep feature learning while PIG avoids the color restoration during training, which help HHEN outperforms the 
FEAT with a large margin (2.72 on TTTS1, 2.45 on INVT1, 1.61 on INVT2). The same result is also shown on the evaluation of APE (Fig. \ref{fig:APE}), while HHEN outperforms the FEAT on all clips. Besides, the MRE of FEAT generated mosaicing rise up after 150 frames while the HHEN continuously maintains the MRE at a relatively low level due to the non-local information learned from long-range frames. We can also observe the same phenomenon from the qualitative results shown in Fig. \ref{fig:SYNGT} as the FEAT mosaicing starts drifting away at 150 frames.

\subsection{Comparison with State-of-The-Art Deep Learning Based Methods}
\label{C4:S5:SS2}

\begin{figure}[t]
    \centering
    \includegraphics[width=\textwidth]{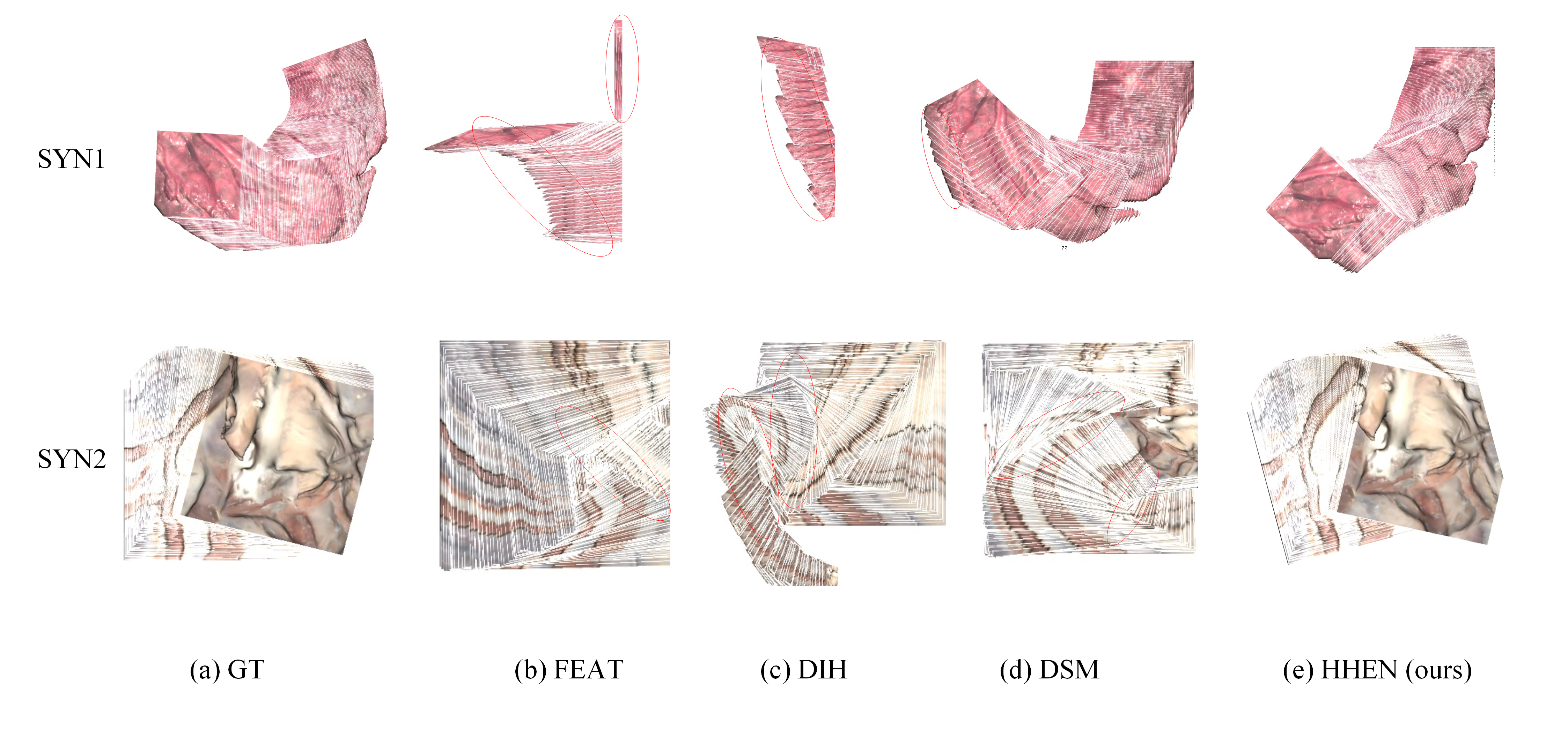}
    \caption{Qualitative result visualization comparison on SYN1 and SYN2 dataset between HHEN and baseline methods. We highlight the mosaicing drift error with red ellipse. Best viewed in colors.}
\label{fig:SYNGT}
\end{figure}

\begin{figure}[t]
    \centering
    \includegraphics[width=\textwidth]{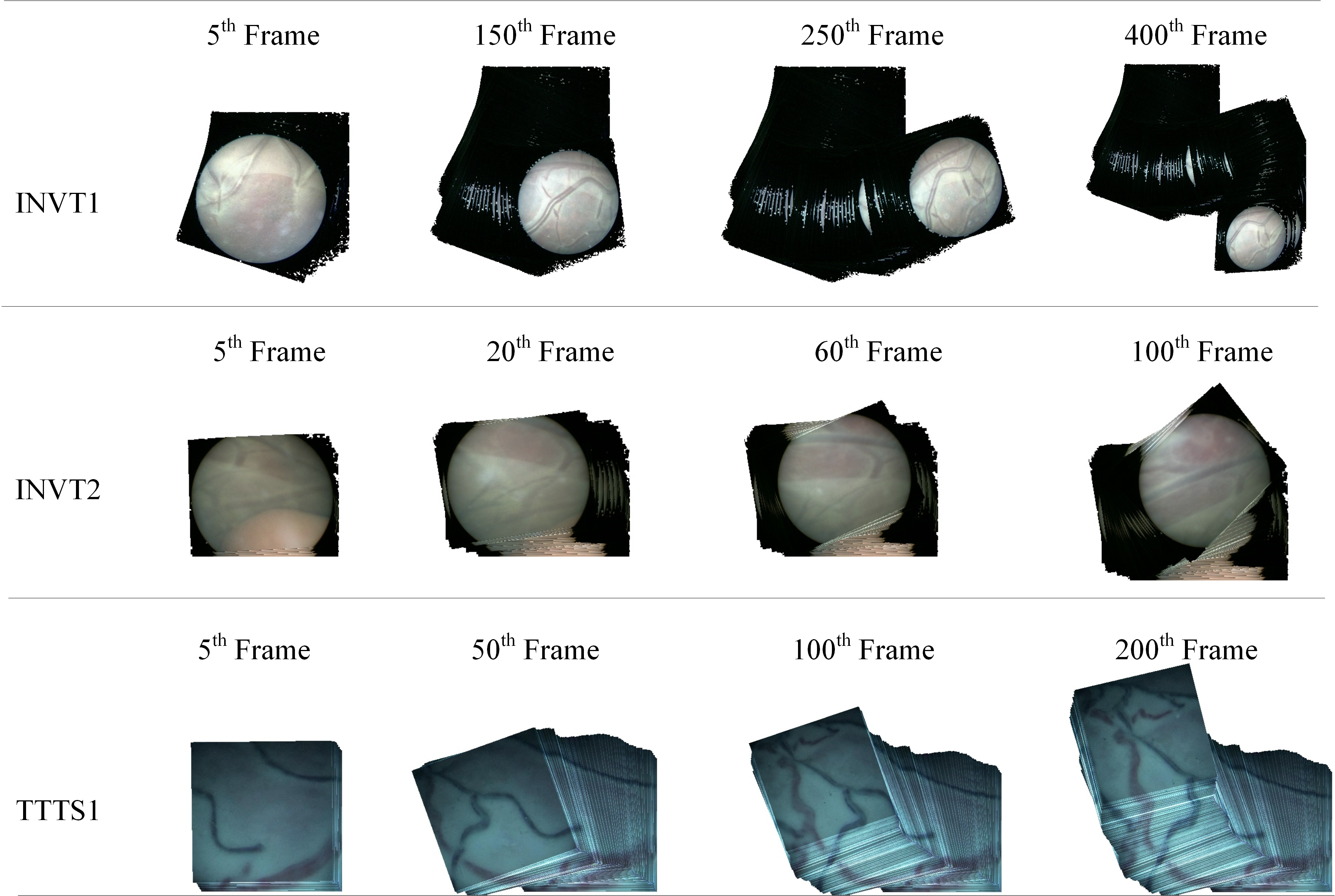}
    \caption{Qualitative result visualization on TTTS1, INVT1 and ONVT2 dataset using generated homography from HHEN. We can observe that HHEN generates meaningful and stable mosaicing even for unseen data INVT1. Best viewed in colors.}
\label{fig:testresult}
\end{figure}

We also compared our proposed HHEN with two state-of-the-art image mosaicing methods, i.e. DIH and DSM. For MSE, DIH explodes very quickly as there are random rotation and translation during training data generation. DSM performs well on short frames due to it learns adjacent homography between adjacent frames. However, the error of DSM starts to accumulate after 300 frames and the mosaicing starts drifting (Fig. \ref{fig:SYNGT}) due to the local information is insufficient to achieve a stable mosaicing during long-range videos. In contrast, our proposed HHEN observed a continuous low MRE even for long-range frames that benefited from non-local information. HHEN also outperforms these two methods on both RMSE and APE. On RMSE, HHEN achieves 0.92, 1.33, 2.72, 2.45 and 1.61 on SYN1, SYN2, TTTS1, INVT1 and INVT2 respectively, while the other two methods obtained higher error (DIH: 3.08, 3.27, 3.86, 4.29, 4.41, DSN: 2.53, 2.17, 2.95, 2.08 3.53). On APE, HHEN also outperforms these two baseline methods on SYN1, SYN2, TTTS1, INVT1 and INVT2 (HHEN: 1.64, 1.76, 2.03, 2.28, 1.04, DIH: 3.27, 3.88, 2.91, 4.65, 3.74, DSM: 2.13, 1.86, 1.90, 2.85, 1.53). We also visualize the mosaicing result of testing sequence TTTS1, INVT1 and INVT2 in Fig. \ref{fig:testresult}. We can observe from Fig. \ref{fig:testresult} that our proposed HHEN with PIG generates the most meaningful mosaicing result even for unseen clip INVT1, which shows the high robustness of our method dealing with different fetoscopic photography scenarios.

\section{Discussion}
\label{C4:S6}
In this project, we utilize explicit deep relation learning on medical image mosaicing. Specifically, we propose a novel hierarchical homography estimation network to effectively learn the homography between adjacent frames that benefited from both local and non-local information. To further generate sufficient training data, we propose a novel image generation method called partially image generation which only perturbs the image with rotation, translation and color transformation. PIG can accelerate HHEN to learn relative homography while ignoring external interference such as color restoration. We conduct extensive experiments on 5 video data clips and the result shows our proposed method performs superior compared with state-of-the-art mosaicing methods. However, there are some blank spaces left for us to improve the mosaicing method. First, the whole network parameter is shared and updated across the data frames, which can be replaced by a recurrent neural network or a long-short term memory network to model the relationship between frames more effectively. Another future direction is to utilize the vessel segmentation mask as auxiliary information, which represents the saliency information, to achieve more accurate image mosaicing.

\chapter{Conclusion and Future Works} 
\label{Chapter5}

\section{Conclusion}
Relational information is a key point for both computational models and clinical physicians to understand the diseases in medical images. In this research project, our goal is to explore the potential of deep relational learning applied to various medical image analysis tasks. We conclude the contribution as follows:
\begin{itemize}
\item \textbf{Context Aware Network for 3D Glioma Segmentation:} We propose an implicit paradigm of relational learning and utilize it to model the relational information between features. Specifically, we propose a novel Hybrid Context Aware Feature Extractor (HCA-FE) built with a 3D feature interaction graph neural network and a 3D encoder-decoder convolutional neural network. To aggregate the features from interaction graph space and convolution space effectively, we also design a Context Guided Attentive Conditional Random Field (CG-ACRF) to regulate the information flow from the HCA-FE. Experimental results on two BraTS dataset, BraTS2017 and BraTS2018, show that our proposed method outperforms state-of-the-art methods.

\item \textbf{Hierarchical Homography Estimation Network for Medical Image Mosaicing} We propose a novel Hierarchical Homography Estimation Network (HHEN) for medical image mosaicing. Specifically, HHEN can utilize both local information between adjacent frames and non-local information between long-range frames. HHEN explicitly learns the spatial relationship between two adjacent image frames, i.e. homography using 6 degree-of-freedom in a data-driven manner. We also propose the novel Partially Image Generation (PIG) to generate perturb patches during HHEN training. PIG can let HHEN learns the rotation and translation while avoiding interferences such as color restoration. Experimental results on 5 different video clips from UCL fetoscopic photography dataset show that our proposed method achieves the state-of-the-art mosaicing results on testing clips and high robustness on unseen clips using generated homography.
\end{itemize}
\section{Future Works}
We list several future research direction as follows:
\begin{enumerate}
\item \textbf{Designing the network automatically:} Like the other state-of-the-art methods, we propose our deep neural network model by manually designing the structure, the layer combination, even the training parameters such as learning rate and the batch size. However, designing an effective neural network model and finding the best combination of different operation layers requires a high-level expertise experience and complicated validation experiments, which makes the networking designing as hard as to win a lottery \cite{frankle2018lottery}. In the future, we hope to explore the automatic network design process, specifically the neural architecture search algorithms, and try to propose novel search algorithms and construct novel search space to search one or more effective network structure candidates on medical image analysis tasks.

\item \textbf{Learning the vision and language information with multi-modal inputs:}  At present, we only focus on learning information purely from medical images. However, in clinical scenarios, other modality data such as medical records and drug prescriptions, also play an important role for physicians. In the future, we may utilize the technologies from image caption \cite{xu2015show} and visual question answering \cite{anderson2018bottom} to explore how the relational information represents in both visual and textual format with multi-modal inputs.

\item \textbf{Deep relational learning on multi-task medical image analysis:} Currently we only complete improving the performance of a particular task, e.g. only for glioma segmentation and only for fetoscopic photography mosaicing. Recent research shows that different vision task contains auxiliary information for each other \cite{zamir2018taskonomy}. In the future, we may explore how to effectively fuse different relational information from different tasks and promote the final performance on multi-task learning (for example, how to use the relation information between different glioma segmented regions to improve the accuracy of glioma degree classification).
\end{enumerate}



\appendix 




\printbibliography[heading=bibintoc]

@article{van2001computer,
  title={Computer-aided diagnosis in chest radiography: a survey},
  author={Van Ginneken, Bram and Romeny, BM Ter Haar and Viergever, Max A},
  journal={IEEE Transactions on medical imaging},
  volume={20},
  number={12},
  pages={1228--1241},
  year={2001},
  publisher={IEEE}
}

@article{li2007improve,
  title={Improve computer-aided diagnosis with machine learning techniques using undiagnosed samples},
  author={Li, Ming and Zhou, Zhi-Hua},
  journal={IEEE Transactions on Systems, Man, and Cybernetics-Part A: Systems and Humans},
  volume={37},
  number={6},
  pages={1088--1098},
  year={2007},
  publisher={IEEE}
}

@article{doi2007computer,
  title={Computer-aided diagnosis in medical imaging: historical review, current status and future potential},
  author={Doi, Kunio},
  journal={Computerized medical imaging and graphics},
  volume={31},
  number={4-5},
  pages={198--211},
  year={2007},
  publisher={Elsevier}
}

@article{miranda2015computer,
  title={Computer-aided diagnosis system based on fuzzy logic for breast cancer categorization},
  author={Miranda, Gisele Helena Barboni and Felipe, Joaquim Cezar},
  journal={Computers in biology and medicine},
  volume={64},
  pages={334--346},
  year={2015},
  publisher={Elsevier}
}

@article{lee2015image,
  title={Image based computer aided diagnosis system for cancer detection},
  author={Lee, Howard and Chen, Yi-Ping Phoebe},
  journal={Expert Systems with Applications},
  volume={42},
  number={12},
  pages={5356--5365},
  year={2015},
  publisher={Elsevier}
}

@article{litjens2015clinical,
  title={Clinical evaluation of a computer-aided diagnosis system for determining cancer aggressiveness in prostate MRI},
  author={Litjens, Geert JS and Barentsz, Jelle O and Karssemeijer, Nico and Huisman, Henkjan J},
  journal={European radiology},
  volume={25},
  number={11},
  pages={3187--3199},
  year={2015},
  publisher={Springer}
}

@article{de2018clinically,
  title={Clinically applicable deep learning for diagnosis and referral in retinal disease},
  author={De Fauw, Jeffrey and Ledsam, Joseph R and Romera-Paredes, Bernardino and Nikolov, Stanislav and Tomasev, Nenad and Blackwell, Sam and Askham, Harry and Glorot, Xavier and O’Donoghue, Brendan and Visentin, Daniel and others},
  journal={Nature medicine},
  volume={24},
  number={9},
  pages={1342--1350},
  year={2018},
  publisher={Nature Publishing Group}
}

@article{ahmad2019artificial,
  title={Artificial intelligence and computer-aided diagnosis in colonoscopy: current evidence and future directions},
  author={Ahmad, Omer F and Soares, Antonio S and Mazomenos, Evangelos and Brandao, Patrick and Vega, Roser and Seward, Edward and Stoyanov, Danail and Chand, Manish and Lovat, Laurence B},
  journal={The Lancet Gastroenterology \& Hepatology},
  volume={4},
  number={1},
  pages={71--80},
  year={2019},
  publisher={Elsevier}
}

@inproceedings{suk2013deep,
  title={Deep learning-based feature representation for AD/MCI classification},
  author={Suk, Heung-Il and Shen, Dinggang},
  booktitle={International Conference on Medical Image Computing and Computer-Assisted Intervention},
  pages={583--590},
  year={2013},
  organization={Springer}
}

@article{hosseini2016alzheimer,
  title={Alzheimer's disease diagnostics by a deeply supervised adaptable 3D convolutional network},
  author={Hosseini-Asl, Ehsan and Gimel'farb, Georgy and El-Baz, Ayman},
  journal={arXiv preprint arXiv:1607.00556},
  year={2016}
}

@inproceedings{suk2016deep,
  title={Deep ensemble sparse regression network for Alzheimer’s disease diagnosis},
  author={Suk, Heung-Il and Shen, Dinggang},
  booktitle={International Workshop on Machine Learning in Medical Imaging},
  pages={113--121},
  year={2016},
  organization={Springer}
}

@article{lian2018hierarchical,
  title={Hierarchical fully convolutional network for joint atrophy localization and Alzheimer's Disease diagnosis using structural MRI},
  author={Lian, Chunfeng and Liu, Mingxia and Zhang, Jun and Shen, Dinggang},
  journal={IEEE transactions on pattern analysis and machine intelligence},
  year={2018},
  publisher={IEEE}
}

@article{havaei2017brain,
  title={Brain tumor segmentation with deep neural networks},
  author={Havaei, Mohammad and Davy, Axel and Warde-Farley, David and Biard, Antoine and Courville, Aaron and Bengio, Yoshua and Pal, Chris and Jodoin, Pierre-Marc and Larochelle, Hugo},
  journal={Medical image analysis},
  volume={35},
  pages={18--31},
  year={2017},
  publisher={Elsevier}
}

@article{pereira2016brain,
  title={Brain tumor segmentation using convolutional neural networks in MRI images},
  author={Pereira, S{\'e}rgio and Pinto, Adriano and Alves, Victor and Silva, Carlos A},
  journal={IEEE transactions on medical imaging},
  volume={35},
  number={5},
  pages={1240--1251},
  year={2016},
  publisher={IEEE}
}

@article{kamnitsas2017efficient,
  title={Efficient multi-scale 3D CNN with fully connected CRF for accurate brain lesion segmentation},
  author={Kamnitsas, Konstantinos and Ledig, Christian and Newcombe, Virginia FJ and Simpson, Joanna P and Kane, Andrew D and Menon, David K and Rueckert, Daniel and Glocker, Ben},
  journal={Medical image analysis},
  volume={36},
  pages={61--78},
  year={2017},
  publisher={Elsevier}
}

@inproceedings{wang2017automatic,
  title={Automatic brain tumor segmentation using cascaded anisotropic convolutional neural networks},
  author={Wang, Guotai and Li, Wenqi and Ourselin, S{\'e}bastien and Vercauteren, Tom},
  booktitle={International MICCAI brainlesion workshop},
  pages={178--190},
  year={2017},
  organization={Springer}
}

@inproceedings{myronenko20183d,
  title={3D MRI brain tumor segmentation using autoencoder regularization},
  author={Myronenko, Andriy},
  booktitle={International MICCAI Brainlesion Workshop},
  pages={311--320},
  year={2018},
  organization={Springer}
}

@inproceedings{fu2016deepvessel,
  title={Deepvessel: Retinal vessel segmentation via deep learning and conditional random field},
  author={Fu, Huazhu and Xu, Yanwu and Lin, Stephen and Wong, Damon Wing Kee and Liu, Jiang},
  booktitle={International conference on medical image computing and computer-assisted intervention},
  pages={132--139},
  year={2016},
  organization={Springer}
}

@inproceedings{wu2016deep,
  title={Deep vessel tracking: A generalized probabilistic approach via deep learning},
  author={Wu, Aaron and Xu, Ziyue and Gao, Mingchen and Buty, Mario and Mollura, Daniel J},
  booktitle={2016 IEEE 13th International Symposium on Biomedical Imaging (ISBI)},
  pages={1363--1367},
  year={2016},
  organization={IEEE}
}

@article{zilly2017glaucoma,
  title={Glaucoma detection using entropy sampling and ensemble learning for automatic optic cup and disc segmentation},
  author={Zilly, Julian and Buhmann, Joachim M and Mahapatra, Dwarikanath},
  journal={Computerized Medical Imaging and Graphics},
  volume={55},
  pages={28--41},
  year={2017},
  publisher={Elsevier}
}

@inproceedings{bar2015deep,
  title={Deep learning with non-medical training used for chest pathology identification},
  author={Bar, Yaniv and Diamant, Idit and Wolf, Lior and Greenspan, Hayit},
  booktitle={Medical Imaging 2015: Computer-Aided Diagnosis},
  volume={9414},
  pages={94140V},
  year={2015},
  organization={International Society for Optics and Photonics}
}

@article{cicero2017training,
  title={Training and validating a deep convolutional neural network for computer-aided detection and classification of abnormalities on frontal chest radiographs},
  author={Cicero, Mark and Bilbily, Alexander and Colak, Errol and Dowdell, Tim and Gray, Bruce and Perampaladas, Kuhan and Barfett, Joseph},
  journal={Investigative radiology},
  volume={52},
  number={5},
  pages={281--287},
  year={2017},
  publisher={LWW}
}

@inproceedings{hwang2016novel,
  title={A novel approach for tuberculosis screening based on deep convolutional neural networks},
  author={Hwang, Sangheum and Kim, Hyo-Eun and Jeong, Jihoon and Kim, Hee-Jin},
  booktitle={Medical imaging 2016: computer-aided diagnosis},
  volume={9785},
  pages={97852W},
  year={2016},
  organization={International Society for Optics and Photonics}
}

@article{huynh2016digital,
  title={Digital mammographic tumor classification using transfer learning from deep convolutional neural networks},
  author={Huynh, Benjamin Q and Li, Hui and Giger, Maryellen L},
  journal={Journal of Medical Imaging},
  volume={3},
  number={3},
  pages={034501},
  year={2016},
  publisher={International Society for Optics and Photonics}
}

@incollection{akselrod2016region,
  title={A region based convolutional network for tumor detection and classification in breast mammography},
  author={Akselrod-Ballin, Ayelet and Karlinsky, Leonid and Alpert, Sharon and Hasoul, Sharbell and Ben-Ari, Rami and Barkan, Ella},
  booktitle={Deep learning and data labeling for medical applications},
  pages={197--205},
  year={2016},
  publisher={Springer}
}

@article{wang2017detecting,
  title={Detecting cardiovascular disease from mammograms with deep learning},
  author={Wang, Juan and Ding, Huanjun and Bidgoli, Fatemeh Azamian and Zhou, Brian and Iribarren, Carlos and Molloi, Sabee and Baldi, Pierre},
  journal={IEEE transactions on medical imaging},
  volume={36},
  number={5},
  pages={1172--1181},
  year={2017},
  publisher={IEEE}
}

@article{dalmics2017using,
  title={Using deep learning to segment breast and fibroglandular tissue in MRI volumes},
  author={Dalm{\i}{\c{s}}, Mehmet Ufuk and Litjens, Geert and Holland, Katharina and Setio, Arnaud and Mann, Ritse and Karssemeijer, Nico and Gubern-M{\'e}rida, Albert},
  journal={Medical physics},
  volume={44},
  number={2},
  pages={533--546},
  year={2017},
  publisher={Wiley Online Library}
}

@article{gao2016hep,
  title={HEp-2 cell image classification with deep convolutional neural networks},
  author={Gao, Zhimin and Wang, Lei and Zhou, Luping and Zhang, Jianjia},
  journal={IEEE journal of biomedical and health informatics},
  volume={21},
  number={2},
  pages={416--428},
  year={2016},
  publisher={IEEE}
}

@article{janowczyk2017stain,
  title={Stain normalization using sparse autoencoders (StaNoSA): application to digital pathology},
  author={Janowczyk, Andrew and Basavanhally, Ajay and Madabhushi, Anant},
  journal={Computerized Medical Imaging and Graphics},
  volume={57},
  pages={50--61},
  year={2017},
  publisher={Elsevier}
}

@inproceedings{xu2016detecting,
  title={Detecting 10,000 cells in one second},
  author={Xu, Zheng and Huang, Junzhou},
  booktitle={International conference on medical image computing and computer-assisted intervention},
  pages={676--684},
  year={2016},
  organization={Springer}
}

@article{litjens2017survey,
  title={A survey on deep learning in medical image analysis},
  author={Litjens, Geert and Kooi, Thijs and Bejnordi, Babak Ehteshami and Setio, Arnaud Arindra Adiyoso and Ciompi, Francesco and Ghafoorian, Mohsen and Van Der Laak, Jeroen Awm and Van Ginneken, Bram and S{\'a}nchez, Clara I},
  journal={Medical image analysis},
  volume={42},
  pages={60--88},
  year={2017},
  publisher={Elsevier}
}

@article{kooi2017large,
  title={Large scale deep learning for computer aided detection of mammographic lesions},
  author={Kooi, Thijs and Litjens, Geert and Van Ginneken, Bram and Gubern-M{\'e}rida, Albert and S{\'a}nchez, Clara I and Mann, Ritse and den Heeten, Ard and Karssemeijer, Nico},
  journal={Medical image analysis},
  volume={35},
  pages={303--312},
  year={2017},
  publisher={Elsevier}
}

@inproceedings{ghafoorian2016non,
  title={Non-uniform patch sampling with deep convolutional neural networks for white matter hyperintensity segmentation},
  author={Ghafoorian, Mohsen and Karssemeijer, Nico and Heskes, Tom and Van Uder, IWM and de Leeuw, Frank-Erik and Marchiori, Elena and van Ginneken, Bram and Platel, Bram},
  booktitle={2016 IEEE 13th International Symposium on Biomedical Imaging (ISBI)},
  pages={1414--1417},
  year={2016},
  organization={IEEE}
}

@article{charbonnier2017improving,
  title={Improving airway segmentation in computed tomography using leak detection with convolutional networks},
  author={Charbonnier, Jean-Paul and Van Rikxoort, Eva M and Setio, Arnaud AA and Schaefer-Prokop, Cornelia M and van Ginneken, Bram and Ciompi, Francesco},
  journal={Medical image analysis},
  volume={36},
  pages={52--60},
  year={2017},
  publisher={Elsevier}
}

@article{van2016fast,
  title={Fast convolutional neural network training using selective data sampling: Application to hemorrhage detection in color fundus images},
  author={Van Grinsven, Mark JJP and van Ginneken, Bram and Hoyng, Carel B and Theelen, Thomas and S{\'a}nchez, Clara I},
  journal={IEEE transactions on medical imaging},
  volume={35},
  number={5},
  pages={1273--1284},
  year={2016},
  publisher={IEEE}
}

@article{litjens2014evaluation,
  title={Evaluation of prostate segmentation algorithms for MRI: the PROMISE12 challenge},
  author={Litjens, Geert and Toth, Robert and van de Ven, Wendy and Hoeks, Caroline and Kerkstra, Sjoerd and van Ginneken, Bram and Vincent, Graham and Guillard, Gwenael and Birbeck, Neil and Zhang, Jindang and others},
  journal={Medical image analysis},
  volume={18},
  number={2},
  pages={359--373},
  year={2014},
  publisher={Elsevier}
}

@article{setio2017validation,
  title={Validation, comparison, and combination of algorithms for automatic detection of pulmonary nodules in computed tomography images: the LUNA16 challenge},
  author={Setio, Arnaud Arindra Adiyoso and Traverso, Alberto and De Bel, Thomas and Berens, Moira SN and van den Bogaard, Cas and Cerello, Piergiorgio and Chen, Hao and Dou, Qi and Fantacci, Maria Evelina and Geurts, Bram and others},
  journal={Medical image analysis},
  volume={42},
  pages={1--13},
  year={2017},
  publisher={Elsevier}
}

@article{bejnordi2017diagnostic,
  title={Diagnostic assessment of deep learning algorithms for detection of lymph node metastases in women with breast cancer},
  author={Bejnordi, Babak Ehteshami and Veta, Mitko and Van Diest, Paul Johannes and Van Ginneken, Bram and Karssemeijer, Nico and Litjens, Geert and Van Der Laak, Jeroen AWM and Hermsen, Meyke and Manson, Quirine F and Balkenhol, Maschenka and others},
  journal={Jama},
  volume={318},
  number={22},
  pages={2199--2210},
  year={2017},
  publisher={American Medical Association}
}

@article{esteva2017dermatologist,
  title={Dermatologist-level classification of skin cancer with deep neural networks},
  author={Esteva, Andre and Kuprel, Brett and Novoa, Roberto A and Ko, Justin and Swetter, Susan M and Blau, Helen M and Thrun, Sebastian},
  journal={nature},
  volume={542},
  number={7639},
  pages={115--118},
  year={2017},
  publisher={Nature Publishing Group}
}

@article{yang2017cascade,
  title={Cascade of multi-scale convolutional neural networks for bone suppression of chest radiographs in gradient domain},
  author={Yang, Wei and Chen, Yingyin and Liu, Yunbi and Zhong, Liming and Qin, Genggeng and Lu, Zhentai and Feng, Qianjin and Chen, Wufan},
  journal={Medical image analysis},
  volume={35},
  pages={421--433},
  year={2017},
  publisher={Elsevier}
}

@incollection{hinton2012practical,
  title={A practical guide to training restricted Boltzmann machines},
  author={Hinton, Geoffrey E},
  booktitle={Neural networks: Tricks of the trade},
  pages={599--619},
  year={2012},
  publisher={Springer}
}

@article{rumelhart1986learning,
  title={Learning representations by back-propagating errors},
  author={Rumelhart, David E and Hinton, Geoffrey E and Williams, Ronald J},
  journal={nature},
  volume={323},
  number={6088},
  pages={533--536},
  year={1986},
  publisher={Nature Publishing Group}
}

@article{hinton2006fast,
  title={A fast learning algorithm for deep belief nets},
  author={Hinton, Geoffrey E and Osindero, Simon and Teh, Yee-Whye},
  journal={Neural computation},
  volume={18},
  number={7},
  pages={1527--1554},
  year={2006},
  publisher={MIT Press}
}

@article{petersen2010alzheimer,
  title={Alzheimer's disease neuroimaging initiative (ADNI): clinical characterization},
  author={Petersen, Ronald Carl and Aisen, PS and Beckett, Laurel A and Donohue, MC and Gamst, AC and Harvey, Danielle J and Jack, CR and Jagust, WJ and Shaw, LM and Toga, AW and others},
  journal={Neurology},
  volume={74},
  number={3},
  pages={201--209},
  year={2010},
  publisher={AAN Enterprises}
}

@article{al2018convolutional,
  title={Convolutional neural networks for electrocardiogram classification},
  author={Al Rahhal, Mohamad M and Bazi, Yakoub and Al Zuair, Mansour and Othman, Esam and BenJdira, Bilel},
  journal={Journal of Medical and Biological Engineering},
  volume={38},
  number={6},
  pages={1014--1025},
  year={2018},
  publisher={Springer}
}

@article{abdel2016breast,
  title={Breast cancer classification using deep belief networks},
  author={Abdel-Zaher, Ahmed M and Eldeib, Ayman M},
  journal={Expert Systems with Applications},
  volume={46},
  pages={139--144},
  year={2016},
  publisher={Elsevier}
}

@inproceedings{arevalo2015convolutional,
  title={Convolutional neural networks for mammography mass lesion classification},
  author={Arevalo, John and Gonz{\'a}lez, Fabio A and Ramos-Poll{\'a}n, Ra{\'u}l and Oliveira, Jose L and Lopez, Miguel Angel Guevara},
  booktitle={2015 37th Annual international conference of the IEEE engineering in medicine and biology society (EMBC)},
  pages={797--800},
  year={2015},
  organization={IEEE}
}

@inproceedings{gao2015mci,
  title={MCI identification by joint learning on multiple MRI data},
  author={Gao, Yue and Wee, Chong-Yaw and Kim, Minjeong and Giannakopoulos, Panteleimon and Montandon, Marie-Louise and Haller, Sven and Shen, Dinggang},
  booktitle={International Conference on Medical Image Computing and Computer-Assisted Intervention},
  pages={78--85},
  year={2015},
  organization={Springer}
}

@article{payan2015predicting,
  title={Predicting Alzheimer's disease: a neuroimaging study with 3D convolutional neural networks},
  author={Payan, Adrien and Montana, Giovanni},
  journal={arXiv preprint arXiv:1502.02506},
  year={2015}
}

@article{gao2015automatic,
  title={Automatic feature learning to grade nuclear cataracts based on deep learning},
  author={Gao, Xinting and Lin, Stephen and Wong, Tien Yin},
  journal={IEEE Transactions on Biomedical Engineering},
  volume={62},
  number={11},
  pages={2693--2701},
  year={2015},
  publisher={IEEE}
}

@article{yu2016automated,
  title={Automated melanoma recognition in dermoscopy images via very deep residual networks},
  author={Yu, Lequan and Chen, Hao and Dou, Qi and Qin, Jing and Heng, Pheng-Ann},
  journal={IEEE transactions on medical imaging},
  volume={36},
  number={4},
  pages={994--1004},
  year={2016},
  publisher={IEEE}
}

@article{li2018skin,
  title={Skin lesion analysis towards melanoma detection using deep learning network},
  author={Li, Yuexiang and Shen, Linlin},
  journal={Sensors},
  volume={18},
  number={2},
  pages={556},
  year={2018},
  publisher={Multidisciplinary Digital Publishing Institute}
}

@inproceedings{ronneberger2015u,
  title={U-net: Convolutional networks for biomedical image segmentation},
  author={Ronneberger, Olaf and Fischer, Philipp and Brox, Thomas},
  booktitle={International Conference on Medical image computing and computer-assisted intervention},
  pages={234--241},
  year={2015},
  organization={Springer}
}

@article{choi2019brain,
  title={Brain tissue segmentation based on MP2RAGE multi-contrast images in 7 T MRI},
  author={Choi, Uk-Su and Kawaguchi, Hirokazu and Matsuoka, Yuichiro and Kober, Tobias and Kida, Ikuhiro},
  journal={PloS one},
  volume={14},
  number={2},
  pages={e0210803},
  year={2019},
  publisher={Public Library of Science San Francisco, CA USA}
}

@article{moeskops2016automatic,
  title={Automatic segmentation of MR brain images with a convolutional neural network},
  author={Moeskops, Pim and Viergever, Max A and Mendrik, Adri{\"e}nne M and De Vries, Linda S and Benders, Manon JNL and I{\v{s}}gum, Ivana},
  journal={IEEE transactions on medical imaging},
  volume={35},
  number={5},
  pages={1252--1261},
  year={2016},
  publisher={IEEE}
}

@article{azar2013decision,
  title={Decision tree classifiers for automated medical diagnosis},
  author={Azar, Ahmad Taher and El-Metwally, Shereen M},
  journal={Neural Computing and Applications},
  volume={23},
  number={7-8},
  pages={2387--2403},
  year={2013},
  publisher={Springer}
}

@article{rojas2017optimal,
  title={Optimal hyper-parameter tuning of SVM classifiers with application to medical diagnosis},
  author={Rojas-Dom{\'\i}nguez, Alfonso and Padierna, Luis Carlos and Valadez, Juan Mart{\'\i}n Carpio and Puga-Soberanes, Hector J and Fraire, H{\'e}ctor J},
  journal={IEEE Access},
  volume={6},
  pages={7164--7176},
  year={2017},
  publisher={IEEE}
}

@inproceedings{khachnaoui2018review,
  title={A review on deep learning in thyroid ultrasound computer-assisted diagnosis systems},
  author={Khachnaoui, Hajer and Guetari, Ramzi and Khlifa, Nawres},
  booktitle={2018 IEEE International Conference on Image Processing, Applications and Systems (IPAS)},
  pages={291--297},
  year={2018},
  organization={IEEE}
}

@inproceedings{netsch2001towards,
  title={Towards real-time multi-modality 3-D medical image registration},
  author={Netsch, Thomas and Rosch, Peter and van Muiswinkel, Arianne and Weese, J{\"u}rgen},
  booktitle={Proceedings Eighth IEEE International Conference on Computer Vision. ICCV 2001},
  volume={1},
  pages={718--725},
  year={2001},
  organization={IEEE}
}

@article{han2002fuzzy,
  title={Fuzzy color histogram and its use in color image retrieval},
  author={Han, Ju and Ma, Kai-Kuang},
  journal={IEEE Transactions on image Processing},
  volume={11},
  number={8},
  pages={944--952},
  year={2002},
  publisher={IEEE}
}

@article{li2002tongue,
  title={Tongue image matching using color content},
  author={Li, Chun Hung and Yuen, Pong C},
  journal={Pattern Recognition},
  volume={35},
  number={2},
  pages={407--419},
  year={2002},
  publisher={Elsevier}
}

@article{wei1997real,
  title={Real-time visual servoing for laparoscopic surgery. Controlling robot motion with color image segmentation},
  author={Wei, Guo-Qing and Arbter, Klaus and Hirzinger, Gerd},
  journal={IEEE Engineering in Medicine and Biology Magazine},
  volume={16},
  number={1},
  pages={40--45},
  year={1997},
  publisher={IEEE}
}

@inproceedings{arjunan2009image,
  title={Image Classification in CBIR systems with color histogram features},
  author={Arjunan, R Vijaya and Kumar, V Vijaya},
  booktitle={2009 International Conference on Advances in Recent Technologies in Communication and Computing},
  pages={593--595},
  year={2009},
  organization={IEEE}
}

@article{tesavr2008medical,
  title={Medical image analysis of 3D CT images based on extension of Haralick texture features},
  author={Tesa{\v{r}}, Ludv{\'\i}k and Shimizu, Akinobu and Smutek, Daniel and Kobatake, Hidefumi and Nawano, Shigeru},
  journal={Computerized Medical Imaging and Graphics},
  volume={32},
  number={6},
  pages={513--520},
  year={2008},
  publisher={Elsevier}
}

@article{patil2011geometrical,
  title={Geometrical and texture features estimation of lung cancer and TB images using chest X-ray database},
  author={Patil, SA and Udupi, VR},
  journal={International Journal of Biomedical Engineering and Technology},
  volume={6},
  number={1},
  pages={58--75},
  year={2011},
  publisher={Inderscience Publishers}
}

@article{sertel2009histopathological,
  title={Histopathological image analysis using model-based intermediate representations and color texture: Follicular lymphoma grading},
  author={Sertel, Olcay and Kong, Jun and Catalyurek, Umit V and Lozanski, Gerard and Saltz, Joel H and Gurcan, Metin N},
  journal={Journal of Signal Processing Systems},
  volume={55},
  number={1-3},
  pages={169},
  year={2009},
  publisher={Springer}
}

@article{nanni2010local,
  title={Local binary patterns variants as texture descriptors for medical image analysis},
  author={Nanni, Loris and Lumini, Alessandra and Brahnam, Sheryl},
  journal={Artificial intelligence in medicine},
  volume={49},
  number={2},
  pages={117--125},
  year={2010},
  publisher={Elsevier}
}

@inproceedings{lowe1999object,
  title={Object recognition from local scale-invariant features},
  author={Lowe, David G},
  booktitle={Proceedings of the seventh IEEE international conference on computer vision},
  volume={2},
  pages={1150--1157},
  year={1999},
  organization={Ieee}
}

@inproceedings{ke2004pca,
  title={PCA-SIFT: A more distinctive representation for local image descriptors},
  author={Ke, Yan and Sukthankar, Rahul},
  booktitle={Proceedings of the 2004 IEEE Computer Society Conference on Computer Vision and Pattern Recognition, 2004. CVPR 2004.},
  volume={2},
  pages={II--II},
  year={2004},
  organization={IEEE}
}

@article{mikolajczyk2005performance,
  title={A performance evaluation of local descriptors},
  author={Mikolajczyk, Krystian and Schmid, Cordelia},
  journal={IEEE transactions on pattern analysis and machine intelligence},
  volume={27},
  number={10},
  pages={1615--1630},
  year={2005},
  publisher={IEEE}
}

@article{bay2008speeded,
  title={Speeded-up robust features (SURF)},
  author={Bay, Herbert and Ess, Andreas and Tuytelaars, Tinne and Van Gool, Luc},
  journal={Computer vision and image understanding},
  volume={110},
  number={3},
  pages={346--359},
  year={2008},
  publisher={Elsevier}
}

@inproceedings{detone2018superpoint,
  title={Superpoint: Self-supervised interest point detection and description},
  author={DeTone, Daniel and Malisiewicz, Tomasz and Rabinovich, Andrew},
  booktitle={Proceedings of the IEEE Conference on Computer Vision and Pattern Recognition Workshops},
  pages={224--236},
  year={2018}
}

@inproceedings{dalal2005histograms,
  title={Histograms of oriented gradients for human detection},
  author={Dalal, Navneet and Triggs, Bill},
  booktitle={2005 IEEE computer society conference on computer vision and pattern recognition (CVPR'05)},
  volume={1},
  pages={886--893},
  year={2005},
  organization={IEEE}
}

@inproceedings{bauer2012segmentation,
  title={Segmentation of brain tumor images based on integrated hierarchical classification and regularization},
  author={Bauer, Stefan and Fejes, Thomas and Slotboom, Johannes and Wiest, Roland and Nolte, Lutz-P and Reyes, Mauricio},
  booktitle={MICCAI BraTS Workshop. Nice: Miccai Society},
  year={2012}
}

@article{subbanna2012probabilistic,
  title={Probabilistic gabor and markov random fields segmentation of brain tumours in mri volumes},
  author={Subbanna, N and Arbel, T},
  journal={Proc MICCAI Brain Tumor Segmentation Challenge (BRATS)},
  pages={28--31},
  year={2012}
}

@inproceedings{shin2012hybrid,
  title={Hybrid clustering and logistic regression for multi-modal brain tumor segmentation},
  author={Shin, Hoo-Chang},
  booktitle={Proc. of Workshops and Challanges in Medical Image Computing and Computer-Assisted Intervention (MICCAI’12)},
  year={2012}
}

@article{festa2013automatic,
  title={Automatic brain tumor segmentation of multi-sequence MR images using random decision forests},
  author={Festa, Joana and Pereira, Sergio and Mariz, Jose Antonio and Sousa, Nuno and Silva, Carlos A},
  journal={Proceedings of NCI-MICCAI BRATS},
  volume={1},
  pages={23--26},
  year={2013}
}

@article{reza2013multi,
  title={Multi-class abnormal brain tissue segmentation using texture},
  author={Reza, S and Iftekharuddin, KM},
  journal={Multimodal Brain Tumor Segmentation},
  volume={38},
  year={2013}
}

@article{goetz2014extremely,
  title={Extremely randomized trees based brain tumor segmentation},
  author={Goetz, Michael and Weber, Christian and Bloecher, Josiah and Stieltjes, Bram and Meinzer, Hans-Peter and Maier-Hein, Klaus},
  journal={Proceeding of BRATS challenge-MICCAI},
  pages={006--011},
  year={2014}
}

@inproceedings{long2015fully,
  title={Fully convolutional networks for semantic segmentation},
  author={Long, Jonathan and Shelhamer, Evan and Darrell, Trevor},
  booktitle={Proceedings of the IEEE conference on computer vision and pattern recognition},
  pages={3431--3440},
  year={2015}
}

@article{corso2008efficient,
  title={Efficient multilevel brain tumor segmentation with integrated bayesian model classification},
  author={Corso, Jason J and Sharon, Eitan and Dube, Shishir and El-Saden, Suzie and Sinha, Usha and Yuille, Alan},
  journal={IEEE transactions on medical imaging},
  volume={27},
  number={5},
  pages={629--640},
  year={2008},
  publisher={IEEE}
}

@inproceedings{wels2008discriminative,
  title={A discriminative model-constrained graph cuts approach to fully automated pediatric brain tumor segmentation in 3-D MRI},
  author={Wels, Michael and Carneiro, Gustavo and Aplas, Alexander and Huber, Martin and Hornegger, Joachim and Comaniciu, Dorin},
  booktitle={International Conference on Medical Image Computing and Computer-Assisted Intervention},
  pages={67--75},
  year={2008},
  organization={Springer}
}

@article{zikic2014segmentation,
  title={Segmentation of brain tumor tissues with convolutional neural networks},
  author={Zikic, Darko and Ioannou, Yani and Brown, Matthew and Criminisi, Antonio},
  journal={Proceedings MICCAI-BRATS},
  pages={36--39},
  year={2014}
}

@article{zhao2018deep,
  title={A deep learning model integrating FCNNs and CRFs for brain tumor segmentation},
  author={Zhao, Xiaomei and Wu, Yihong and Song, Guidong and Li, Zhenye and Zhang, Yazhuo and Fan, Yong},
  journal={Medical image analysis},
  volume={43},
  pages={98--111},
  year={2018},
  publisher={Elsevier}
}

@inproceedings{dong2017automatic,
  title={Automatic brain tumor detection and segmentation using U-Net based fully convolutional networks},
  author={Dong, Hao and Yang, Guang and Liu, Fangde and Mo, Yuanhan and Guo, Yike},
  booktitle={annual conference on medical image understanding and analysis},
  pages={506--517},
  year={2017},
  organization={Springer}
}

@inproceedings{lyksborg2015ensemble,
  title={An ensemble of 2D convolutional neural networks for tumor segmentation},
  author={Lyksborg, Mark and Puonti, Oula and Agn, Mikael and Larsen, Rasmus},
  booktitle={Scandinavian Conference on Image Analysis},
  pages={201--211},
  year={2015},
  organization={Springer}
}

@inproceedings{kamnitsas2017ensembles,
  title={Ensembles of multiple models and architectures for robust brain tumour segmentation},
  author={Kamnitsas, Konstantinos and Bai, Wenjia and Ferrante, Enzo and McDonagh, Steven and Sinclair, Matthew and Pawlowski, Nick and Rajchl, Martin and Lee, Matthew and Kainz, Bernhard and Rueckert, Daniel and others},
  booktitle={International MICCAI Brainlesion Workshop},
  pages={450--462},
  year={2017},
  organization={Springer}
}

@inproceedings{chen2018focus,
  title={Focus, Segment and Erase: An Efficient Network for Multi-Label Brain Tumor Segmentation},
  author={Chen, Xuan and Hao Liew, Jun and Xiong, Wei and Chui, Chee-Kong and Ong, Sim-Heng},
  booktitle={Proceedings of the European Conference on Computer Vision (ECCV)},
  pages={654--669},
  year={2018}
}

@inproceedings{zhao20173d,
  title={3D brain tumor segmentation through integrating multiple 2D FCNNs},
  author={Zhao, Xiaomei and Wu, Yihong and Song, Guidong and Li, Zhenye and Zhang, Yazhuo and Fan, Yong},
  booktitle={International MICCAI Brainlesion Workshop},
  pages={191--203},
  year={2017},
  organization={Springer}
}

@inproceedings{krahenbuhl2011efficient,
  title={Efficient inference in fully connected crfs with gaussian edge potentials},
  author={Kr{\"a}henb{\"u}hl, Philipp and Koltun, Vladlen},
  booktitle={Advances in neural information processing systems},
  pages={109--117},
  year={2011}
}

@article{bakas2017advancing,
  title={Advancing the cancer genome atlas glioma MRI collections with expert segmentation labels and radiomic features},
  author={Bakas, Spyridon and Akbari, Hamed and Sotiras, Aristeidis and Bilello, Michel and Rozycki, Martin and Kirby, Justin S and Freymann, John B and Farahani, Keyvan and Davatzikos, Christos},
  journal={Scientific data},
  volume={4},
  pages={170117},
  year={2017},
  publisher={Nature Publishing Group}
}

@inproceedings{cciccek20163d,
  title={3D U-Net: learning dense volumetric segmentation from sparse annotation},
  author={{\c{C}}i{\c{c}}ek, {\"O}zg{\"u}n and Abdulkadir, Ahmed and Lienkamp, Soeren S and Brox, Thomas and Ronneberger, Olaf},
  booktitle={International conference on medical image computing and computer-assisted intervention},
  pages={424--432},
  year={2016},
  organization={Springer}
}

@article{oktay2018attention,
  title={Attention u-net: Learning where to look for the pancreas},
  author={Oktay, Ozan and Schlemper, Jo and Folgoc, Loic Le and Lee, Matthew and Heinrich, Mattias and Misawa, Kazunari and Mori, Kensaku and McDonagh, Steven and Hammerla, Nils Y and Kainz, Bernhard and others},
  journal={arXiv preprint arXiv:1804.03999},
  year={2018}
}

@article{brugger2019partially,
  title={A Partially Reversible U-Net for Memory-Efficient Volumetric Image Segmentation},
  author={Br{\"u}gger, Robin and Baumgartner, Christian F and Konukoglu, Ender},
  journal={arXiv preprint arXiv:1906.06148},
  year={2019}
}

@inproceedings{isensee2018no,
  title={No new-net},
  author={Isensee, Fabian and Kickingereder, Philipp and Wick, Wolfgang and Bendszus, Martin and Maier-Hein, Klaus H},
  booktitle={International MICCAI Brainlesion Workshop},
  pages={234--244},
  year={2018},
  organization={Springer}
}

@inproceedings{mehta20193despnet,
author="Nuechterlein, Nicholas and Mehta, Sachin",
title="3D-ESPNet with Pyramidal Refinement for Volumetric Brain Tumor Image Segmentation",
booktitle="Brainlesion: Glioma, Multiple Sclerosis, Stroke and Traumatic Brain Injuries",
year="2019",
publisher="Springer International Publishing",
}

@article{pereira2019adaptive,
  title={Adaptive feature recombination and recalibration for semantic segmentation with Fully Convolutional Networks},
  author={Pereira, S{\'e}rgio and Pinto, Adriano and Amorim, Joana and Ribeiro, Alexandrine and Alves, Victor and Silva, Carlos A},
  journal={IEEE transactions on medical imaging},
  year={2019},
  publisher={IEEE}
}

@article{roy2018recalibrating,
  title={Recalibrating fully convolutional networks with spatial and channel “squeeze and excitation” blocks},
  author={Roy, Abhijit Guha and Navab, Nassir and Wachinger, Christian},
  journal={IEEE transactions on medical imaging},
  volume={38},
  number={2},
  pages={540--549},
  year={2018},
  publisher={IEEE}
}

@inproceedings{isensee2017brain,
  title={Brain tumor segmentation and radiomics survival prediction: Contribution to the brats 2017 challenge},
  author={Isensee, Fabian and Kickingereder, Philipp and Wick, Wolfgang and Bendszus, Martin and Maier-Hein, Klaus H},
  booktitle={International MICCAI Brainlesion Workshop},
  pages={287--297},
  year={2017},
  organization={Springer}
}

@inproceedings{jungo2017towards,
  title={Towards uncertainty-assisted brain tumor segmentation and survival prediction},
  author={Jungo, Alain and McKinley, Richard and Meier, Raphael and Knecht, Urspeter and Vera, Luis and P{\'e}rez-Beteta, Juli{\'a}n and Molina-Garc{\'\i}a, David and P{\'e}rez-Garc{\'\i}a, V{\'\i}ctor M and Wiest, Roland and Reyes, Mauricio},
  booktitle={International MICCAI Brainlesion Workshop},
  pages={474--485},
  year={2017},
  organization={Springer}
}

@inproceedings{islam2017multi,
  title={Multi-modal pixelnet for brain tumor segmentation},
  author={Islam, Mobarakol and Ren, Hongliang},
  booktitle={International MICCAI Brainlesion Workshop},
  pages={298--308},
  year={2017},
  organization={Springer}
}

@inproceedings{jesson2017brain,
  title={Brain tumor segmentation using a 3D FCN with multi-scale loss},
  author={Jesson, Andrew and Arbel, Tal},
  booktitle={International MICCAI Brainlesion Workshop},
  pages={392--402},
  year={2017},
  organization={Springer}
}

@inproceedings{mckinley2018ensembles,
  title={Ensembles of densely-connected CNNs with label-uncertainty for brain tumor segmentation},
  author={McKinley, Richard and Meier, Raphael and Wiest, Roland},
  booktitle={International MICCAI Brainlesion Workshop},
  pages={456--465},
  year={2018},
  organization={Springer}
}

@inproceedings{zhou2018learning,
  title={Learning contextual and attentive information for brain tumor segmentation},
  author={Zhou, Chenhong and Chen, Shengcong and Ding, Changxing and Tao, Dacheng},
  booktitle={International MICCAI Brainlesion Workshop},
  pages={497--507},
  year={2018},
  organization={Springer}
}

@article{cabezas2018survival,
  title={Survival prediction using ensemble tumor segmentation and transfer learning},
  author={Cabezas, Mariano and Valverde, Sergi and Gonz{\'a}lez-Vill{\`a}, Sandra and Cl{\'e}rigues, Albert and Salem, Mostafa and Kushibar, Kaisar and Bernal, Jose and Oliver, Arnau and Llad{\'o}, Xavier},
  journal={arXiv preprint arXiv:1810.04274},
  year={2018}
}

@inproceedings{feng2018brain,
  title={Brain tumor segmentation using an ensemble of 3d u-nets and overall survival prediction using radiomic features},
  author={Feng, Xue and Tustison, Nicholas and Meyer, Craig},
  booktitle={International MICCAI Brainlesion Workshop},
  pages={279--288},
  year={2018},
  organization={Springer}
}

@inproceedings{sun2018tumor,
  title={Tumor segmentation and survival prediction in glioma with deep learning},
  author={Sun, Li and Zhang, Songtao and Luo, Lin},
  booktitle={International MICCAI Brainlesion Workshop},
  pages={83--93},
  year={2018},
  organization={Springer}
}

@inproceedings{weninger2018segmentation,
  title={Segmentation of brain tumors and patient survival prediction: Methods for the brats 2018 challenge},
  author={Weninger, Leon and Rippel, Oliver and Koppers, Simon and Merhof, Dorit},
  booktitle={International MICCAI Brainlesion Workshop},
  pages={3--12},
  year={2018},
  organization={Springer}
}

@inproceedings{gates2018glioma,
  title={Glioma Segmentation and a Simple Accurate Model for Overall Survival Prediction},
  author={Gates, Evan and Pauloski, J Gregory and Schellingerhout, Dawid and Fuentes, David},
  booktitle={International MICCAI Brainlesion Workshop},
  pages={476--484},
  year={2018},
  organization={Springer}
}

@article{kanazawa2004image,
  title={Image mosaicing by stratified matching},
  author={Kanazawa, Yasushi and Kanatani, Kenichi},
  journal={Image and Vision computing},
  volume={22},
  number={2},
  pages={93--103},
  year={2004},
  publisher={Elsevier}
}

@article{zagrouba2009efficient,
  title={An efficient image-mosaicing method based on multifeature matching},
  author={Zagrouba, Ezzeddine and Barhoumi, Walid and Amri, Slim},
  journal={Machine Vision and Applications},
  volume={20},
  number={3},
  pages={139--162},
  year={2009},
  publisher={Springer}
}

@inproceedings{kourogi1999real,
  title={Real-time image mosaicing from a video sequence},
  author={Kourogi, Masakatsu and Kurata, Takeshi and Hoshino, Junichi and Muraoka, Yoichi},
  booktitle={Proceedings 1999 International Conference on Image Processing (Cat. 99CH36348)},
  volume={4},
  pages={133--137},
  year={1999},
  organization={IEEE}
}

@inproceedings{heikkila2005image,
  title={An image mosaicing module for wide-area surveillance},
  author={Heikkil{\"a}, Marko and Pietik{\"a}inen, Matti},
  booktitle={Proceedings of the third ACM international workshop on Video surveillance \& sensor networks},
  pages={11--18},
  year={2005}
}

@inproceedings{botterill2010real,
  title={Real-time aerial image mosaicing},
  author={Botterill, Tom and Mills, Steven and Green, Richard},
  booktitle={2010 25th International Conference of Image and Vision Computing New Zealand},
  pages={1--8},
  year={2010},
  organization={IEEE}
}

@inproceedings{lv2019improved,
  title={An Improved SURF in Image Mosaic Based on Deep Learning},
  author={Lv, Hao and Zhang, Haiyang and Zhao, Changming and Liu, Chun and Qi, Faguo and Zhang, Zilong},
  booktitle={2019 IEEE 4th International Conference on Image, Vision and Computing (ICIVC)},
  pages={223--226},
  year={2019},
  organization={IEEE}
}

@article{zhang2019convolutional,
  title={Convolutional neural network-based registration for mosaicing of microscopic images},
  author={Zhang, Junhua and Huang, Yihua and Song, Yingchao and Jiang, Yi and Zhang, Lun and Zhang, Yufeng},
  journal={Journal of Electronic Imaging},
  volume={28},
  number={4},
  pages={043006},
  year={2019},
  publisher={International Society for Optics and Photonics}
}

@article{bano2020deep,
  title={Deep learning-based fetoscopic mosaicking for field-of-view expansion},
  author={Bano, Sophia and Vasconcelos, Francisco and Tella-Amo, Marcel and Dwyer, George and Gruijthuijsen, Caspar and Vander Poorten, Emmanuel and Vercauteren, Tom and Ourselin, Sebastien and Deprest, Jan and Stoyanov, Danail},
  journal={International Journal of Computer Assisted Radiology and Surgery},
  pages={1--10},
  year={2020},
  publisher={Springer}
}

@article{nguyen2018unsupervised,
  title={Unsupervised deep homography: A fast and robust homography estimation model},
  author={Nguyen, Ty and Chen, Steven W and Shivakumar, Shreyas S and Taylor, Camillo Jose and Kumar, Vijay},
  journal={IEEE Robotics and Automation Letters},
  volume={3},
  number={3},
  pages={2346--2353},
  year={2018},
  publisher={IEEE}
}

@inproceedings{wang2018non,
  title={Non-local neural networks},
  author={Wang, Xiaolong and Girshick, Ross and Gupta, Abhinav and He, Kaiming},
  booktitle={Proceedings of the IEEE conference on computer vision and pattern recognition},
  pages={7794--7803},
  year={2018}
}

@article{jiang2018modeling,
  title={Modeling multimodal clues in a hybrid deep learning framework for video classification},
  author={Jiang, Yu-Gang and Wu, Zuxuan and Tang, Jinhui and Li, Zechao and Xue, Xiangyang and Chang, Shih-Fu},
  journal={IEEE Transactions on Multimedia},
  volume={20},
  number={11},
  pages={3137--3147},
  year={2018},
  publisher={IEEE}
}

@inproceedings{anderson2018bottom,
  title={Bottom-up and top-down attention for image captioning and visual question answering},
  author={Anderson, Peter and He, Xiaodong and Buehler, Chris and Teney, Damien and Johnson, Mark and Gould, Stephen and Zhang, Lei},
  booktitle={Proceedings of the IEEE conference on computer vision and pattern recognition},
  pages={6077--6086},
  year={2018}
}

@inproceedings{bu2016map2dfusion,
  title={Map2DFusion: Real-time incremental UAV image mosaicing based on monocular slam},
  author={Bu, Shuhui and Zhao, Yong and Wan, Gang and Liu, Zhenbao},
  booktitle={2016 IEEE/RSJ International Conference on Intelligent Robots and Systems (IROS)},
  pages={4564--4571},
  year={2016},
  organization={IEEE}
}

@inproceedings{azzari2008markerless,
  title={Markerless augmented reality using image mosaics},
  author={Azzari, Pietro and Di Stefano, Luigi and Tombari, Federico and Mattoccia, Stefano},
  booktitle={International Conference on Image and Signal Processing},
  pages={413--420},
  year={2008},
  organization={Springer}
}

@article{li2018deep,
  title={Deep learning based module defect analysis for large-scale photovoltaic farms},
  author={Li, Xiaoxia and Yang, Qiang and Lou, Zhuo and Yan, Wenjun},
  journal={IEEE Transactions on Energy Conversion},
  volume={34},
  number={1},
  pages={520--529},
  year={2018},
  publisher={IEEE}
}

@inproceedings{bano2020deep2,
  title={Deep placental vessel segmentation for fetoscopic mosaicking},
  author={Bano, Sophia and Vasconcelos, Francisco and Shepherd, Luke M and Vander Poorten, Emmanuel and Vercauteren, Tom and Ourselin, Sebastien and David, Anna L and Deprest, Jan and Stoyanov, Danail},
  booktitle={International Conference on Medical Image Computing and Computer-Assisted Intervention},
  pages={763--773},
  year={2020},
  organization={Springer}
}

@article{detone2016deep,
  title={Deep image homography estimation},
  author={DeTone, Daniel and Malisiewicz, Tomasz and Rabinovich, Andrew},
  journal={arXiv preprint arXiv:1606.03798},
  year={2016}
}

@article{tella2018probabilistic,
  title={Probabilistic visual and electromagnetic data fusion for robust drift-free sequential mosaicking: application to fetoscopy},
  author={Tella-Amo, Marcel and Peter, Loic and Shakir, Dzhoshkun I and Deprest, Jan and Stoyanov, Danail and Iglesias, Juan E and Vercauteren, Tom and Ourselin, Sebastien},
  journal={Journal of Medical Imaging},
  volume={5},
  number={2},
  pages={021217},
  year={2018},
  publisher={International Society for Optics and Photonics}
}

@inproceedings{buades2005non,
  title={A non-local algorithm for image denoising},
  author={Buades, Antoni and Coll, Bartomeu and Morel, J-M},
  booktitle={2005 IEEE Computer Society Conference on Computer Vision and Pattern Recognition (CVPR'05)},
  volume={2},
  pages={60--65},
  year={2005},
  organization={IEEE}
}

@inproceedings{lin2014microsoft,
  title={Microsoft coco: Common objects in context},
  author={Lin, Tsung-Yi and Maire, Michael and Belongie, Serge and Hays, James and Perona, Pietro and Ramanan, Deva and Doll{\'a}r, Piotr and Zitnick, C Lawrence},
  booktitle={European conference on computer vision},
  pages={740--755},
  year={2014},
  organization={Springer}
}

@inproceedings{bano2020vessel,
title={Deep Placental Vessel Segmentation for Fetoscopic Mosaicking},
author={Bano, Sophia and Vasconcelos, Francisco and Shepherd, Luke M. and Vander Poorten, Emmanue and Vercauteren, Tom and Ourselin, Sebastien and David, Anna L. and Deprest, Jan and Stoyanov, Danail},
booktitle={International Conference on Medical Image Computing and Computer-Assisted Intervention},
year={2020},
organization={Springer}
}

@article{brown2007automatic,
  title={Automatic panoramic image stitching using invariant features},
  author={Brown, Matthew and Lowe, David G},
  journal={International journal of computer vision},
  volume={74},
  number={1},
  pages={59--73},
  year={2007},
  publisher={Springer}
}

@inproceedings{brasch2018semantic,
  title={Semantic monocular SLAM for highly dynamic environments},
  author={Brasch, Nikolas and Bozic, Aljaz and Lallemand, Joe and Tombari, Federico},
  booktitle={2018 IEEE/RSJ International Conference on Intelligent Robots and Systems (IROS)},
  pages={393--400},
  year={2018},
  organization={IEEE}
}

@inproceedings{godard2019digging,
  title={Digging into self-supervised monocular depth estimation},
  author={Godard, Cl{\'e}ment and Mac Aodha, Oisin and Firman, Michael and Brostow, Gabriel J},
  booktitle={Proceedings of the IEEE international conference on computer vision},
  pages={3828--3838},
  year={2019}
}

@inproceedings{frankle2018lottery,
  title={The Lottery Ticket Hypothesis: Finding Sparse, Trainable Neural Networks},
  author={Frankle, Jonathan and Carbin, Michael},
  booktitle={International Conference on Learning Representations},
  year={2018}
}

@inproceedings{xu2015show,
  title={Show, attend and tell: Neural image caption generation with visual attention},
  author={Xu, Kelvin and Ba, Jimmy and Kiros, Ryan and Cho, Kyunghyun and Courville, Aaron and Salakhudinov, Ruslan and Zemel, Rich and Bengio, Yoshua},
  booktitle={International conference on machine learning},
  pages={2048--2057},
  year={2015}
}

@inproceedings{zamir2018taskonomy,
  title={Taskonomy: Disentangling task transfer learning},
  author={Zamir, Amir R and Sax, Alexander and Shen, William and Guibas, Leonidas J and Malik, Jitendra and Savarese, Silvio},
  booktitle={Proceedings of the IEEE conference on computer vision and pattern recognition},
  pages={3712--3722},
  year={2018}
}

@article{finlayson2017color,
  title={Color homography: theory and applications},
  author={Finlayson, Graham and Gong, Han and Fisher, Robert B},
  journal={IEEE transactions on pattern analysis and machine intelligence},
  volume={41},
  number={1},
  pages={20--33},
  year={2017},
  publisher={IEEE}
}

@article{su2020multimodal,
  title={Multimodal Glioma Image Segmentation Using Dual Encoder Structure and Channel Spatial Attention Block},
  author={Su, Run and Liu, Jinhuai and Zhang, Deyun and Cheng, Chuandong and Ye, Mingquan},
  journal={Frontiers in Neuroscience},
  year={2020},
  publisher={Frontiers Research Foundation}
}

@inproceedings{jiang2019two,
  title={Two-Stage Cascaded U-Net: 1st Place Solution to BraTS Challenge 2019 Segmentation Task},
  author={Jiang, Zeyu and Ding, Changxing and Liu, Minfeng and Tao, Dacheng},
  booktitle={International MICCAI Brainlesion Workshop},
  pages={231--241},
  year={2019},
  organization={Springer}
}

@article{wang2020modality,
  title={Modality-Pairing Learning for Brain Tumor Segmentation},
  author={Wang, Yixin and Zhang, Yao and Hou, Feng and Liu, Yang and Tian, Jiang and Zhong, Cheng and Zhang, Yang and He, Zhiqiang},
  journal={arXiv preprint arXiv:2010.09277},
  year={2020}
}

@inproceedings{dorent2019hetero,
  title={Hetero-modal variational encoder-decoder for joint modality completion and segmentation},
  author={Dorent, Reuben and Joutard, Samuel and Modat, Marc and Ourselin, S{\'e}bastien and Vercauteren, Tom},
  booktitle={International Conference on Medical Image Computing and Computer-Assisted Intervention},
  pages={74--82},
  year={2019},
  organization={Springer}
}

@inproceedings{qi20173d,
  title={3d graph neural networks for rgbd semantic segmentation},
  author={Qi, Xiaojuan and Liao, Renjie and Jia, Jiaya and Fidler, Sanja and Urtasun, Raquel},
  booktitle={Proceedings of the IEEE International Conference on Computer Vision},
  pages={5199--5208},
  year={2017}
}

@inproceedings{landrieu2018large,
  title={Large-scale point cloud semantic segmentation with superpoint graphs},
  author={Landrieu, Loic and Simonovsky, Martin},
  booktitle={Proceedings of the IEEE Conference on Computer Vision and Pattern Recognition},
  pages={4558--4567},
  year={2018}
}

@inproceedings{lu2019graph,
  title={Graph-FCN for image semantic segmentation},
  author={Lu, Yi and Chen, Yaran and Zhao, Dongbin and Chen, Jianxin},
  booktitle={International Symposium on Neural Networks},
  pages={97--105},
  year={2019},
  organization={Springer}
}

@article{liu2020scg,
  title={SCG-Net: Self-Constructing Graph Neural Networks for Semantic Segmentation},
  author={Liu, Qinghui and Kampffmeyer, Michael and Jenssen, Robert and Salberg, Arnt-B{\o}rre},
  journal={arXiv preprint arXiv:2009.01599},
  year={2020}
}

@article{zhang2019dual,
  title={Dual graph convolutional network for semantic segmentation},
  author={Zhang, Li and Li, Xiangtai and Arnab, Anurag and Yang, Kuiyuan and Tong, Yunhai and Torr, Philip HS},
  journal={arXiv preprint arXiv:1909.06121},
  year={2019}
}

@article{zhang2020exploring,
  title={Exploring Task Structure for Brain Tumor Segmentation From Multi-Modality MR Images},
  author={Zhang, Dingwen and Huang, Guohai and Zhang, Qiang and Han, Jungong and Han, Junwei and Wang, Yizhou and Yu, Yizhou},
  journal={IEEE Transactions on Image Processing},
  volume={29},
  pages={9032--9043},
  year={2020},
  publisher={IEEE}
}


\end{document}